%% file: main.tex
\documentclass[10pt,twocolumn,letterpaper]{article}

\usepackage{iccv}
\usepackage{times}
\usepackage{epsfig}
\usepackage{graphicx}
\usepackage{amsmath}
\usepackage{amssymb}

\usepackage[accsupp]{axessibility}  

\usepackage{booktabs}
\usepackage{caption}
\usepackage{subfigure}
\usepackage[table]{xcolor}
\usepackage{setspace}
\usepackage{marvosym}
\usepackage[pagebackref=true,breaklinks=true,bookmarks=false,colorlinks=true]{hyperref}

\usepackage[capitalize]{cleveref}
\crefname{section}{Sec.}{Secs.}
\crefname{Section}{Sec.}{Secs.}
\crefname{table}{Table}{Tables}
\crefname{table}{Tab.}{Tabs.}

\definecolor{Dark}{rgb}{0.8902,0.9098,0.9412}
\definecolor{correct}{RGB}{173, 173, 173}
\definecolor{incorrect}{RGB}{192, 0, 0}

\iccvfinalcopy 

\def\iccvPaperID{1502} 

\begin{document}

\title{Robo3D: Towards Robust and Reliable 3D Perception against Corruptions}

\author{Lingdong Kong$^{1,2,3,*}$ \quad Youquan Liu$^{1,4,*}$ \quad Xin Li$^{1,5,*}$ \quad Runnan Chen$^{1,6}$ \quad Wenwei Zhang$^{1,7}$\\ Jiawei Ren$^{7}$ \quad Liang Pan$^{7}$ \quad Kai Chen$^{1}$ \quad Ziwei Liu$^{7,\textrm{\Letter}}$
\\[0.12ex]
\small{
$^{1}$Shanghai AI Laboratory \quad
$^{2}$National University of Singapore \quad
$^{3}$CNRS@CREATE \quad
$^{4}$Hochschule Bremerhaven}\\
\small{
$^{5}$East China Normal University \quad
$^{6}$The University of Hong Kong \quad
$^{7}$S-Lab, Nanyang Technological University
}
\\
{\scriptsize\texttt{\{konglingdong,liuyouquan,lixin,zhangwenwei,chenkai\}@pjlab.org.cn} \quad {\scriptsize\texttt{\{jiawei011,liang.pan,ziwei.liu\}@ntu.edu.sg}} ~~~~~~~
\vspace{-0.21cm}
}}

\twocolumn[{
    \renewcommand\twocolumn[1][]{#1}
    \maketitle
    \centering
    \vspace{-0.2cm}
    \includegraphics[width=\textwidth]{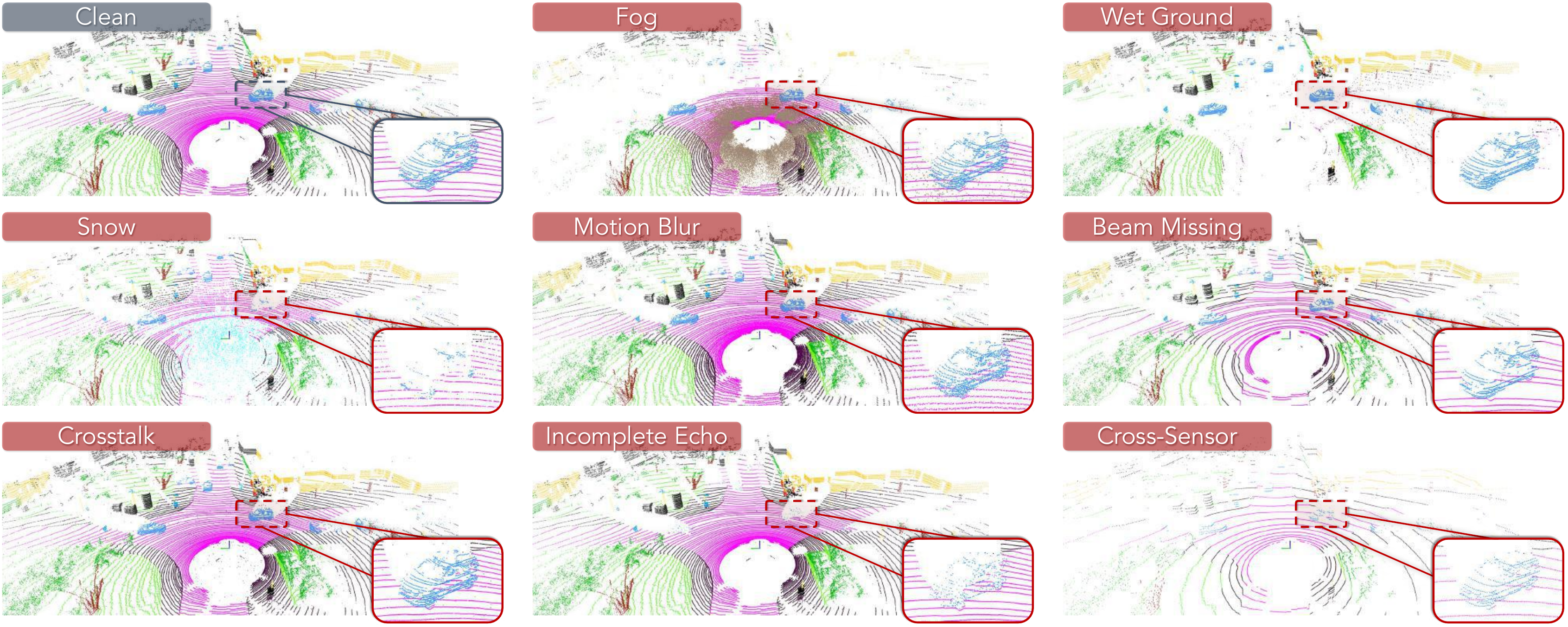}
    \vspace{-0.6cm}
    \captionof{figure}{Taxonomy of the \textbf{Robo3D} benchmark. We simulate \textbf{\textit{eight}} corruption types from \textbf{\textit{three}} categories: \textbf{1)} \textit{Severe weather conditions}, such as fog, rain, and snow; \textbf{2)} \textit{External disturbances} that are caused by motion blur or result in the missing of LiDAR beams; and \textbf{3)} \textit{Internal sensor failure}, including LiDAR crosstalk, possible incomplete echo, and cross-sensor scenarios. Each corruption is further split into \textbf{\textit{three}} levels (light, moderate, and heavy) based on its severity.}
    \label{fig:taxonomy}
    \vspace{0.65cm}
}]


\newcommand\blfootnote[1]{%
\begingroup
\renewcommand\thefootnote{}{}\footnote{#1}%
\addtocounter{footnote}{-1}%
\endgroup
}

\blfootnote{${(*)}$~The first three authors contributed equally to this work.}

\vspace{-0.3cm}
\begin{abstract}
   The robustness of 3D perception systems under natural corruptions from environments and sensors is pivotal for safety-critical applications. Existing large-scale 3D perception datasets often contain data that are meticulously cleaned. Such configurations, however, cannot reflect the reliability of perception models during the deployment stage. In this work, we present \textbf{Robo3D}, the first comprehensive benchmark heading toward probing the robustness of 3D detectors and segmentors under out-of-distribution scenarios against natural corruptions that occur in real-world environments. Specifically, we consider eight corruption types stemming from severe weather conditions, external disturbances, and internal sensor failure. We uncover that, although promising results have been progressively achieved on standard benchmarks, state-of-the-art 3D perception models are at risk of being vulnerable to corruptions. We draw key observations on the use of data representations, augmentation schemes, and training strategies, that could severely affect the model's performance. To pursue better robustness, we propose a density-insensitive training framework along with a simple flexible voxelization strategy to enhance the model resiliency. We hope our benchmark and approach could inspire future research in designing more robust and reliable 3D perception models. Our robustness benchmark suite is publicly available\footnote{\url{https://github.com/ldkong1205/Robo3D}.}.
   \vspace{-0.2cm}
\end{abstract}

\section{Introduction}
\label{sec:intro}
3D perception aims to detect and segment accurate position, orientation, semantics, and temporary relation of the objects and backgrounds around the ego-vehicle in the three-dimensional world \cite{3ddet_survey,guo2020deep,huang2202multi}. With the emergence of large-scale autonomous driving perception datasets, various approaches in the fields of LiDAR semantic segmentation and 3D object detection advent each year, with record-breaking performances on the mainstream benchmarks \cite{geiger2012we,SemanticKITTI,nuScenes,Panoptic-nuScenes,WaymoOpen}.

Despite the great success achieved on the ``clean" evaluation sets, the  model's robustness against out-of-distribution (OoD) scenarios remain obscure. Recent attempts mainly focus on probing the OoD robustness from two aspects. The first line focuses on the transfer of 3D perception models to unseen domains, \textit{e.g.}, sim2real \cite{SynLiDAR}, day2night \cite{xmuda}, and city2city \cite{kong2021conda} adaptations, to probe the model's generalizability. The second line aims to design adversarial examples which can cause the model to make incorrect predictions while keeping the attacked input close to its original format, \textit{i.e.}, to test the model's worst-case scenarios \cite{rossolini2022real,cao2021invisible,tu2020physically}.

In this work, different from the above two directions, we aim at understanding the cause of performance deterioration under real-world corruption and sensor failure. Current 3D perception models learn point features from LiDAR sensors or RGB-D cameras, where data corruptions are inevitable due to issues of data collection, processing, weather conditions, and scene complexity \cite{ren2022modelnet-c}. While recent works target creating corrupted point clouds from indoor scenes \cite{3dcc} or object-centric CAD models \cite{sun,PV-Ada,alliegro2022open}, we simulate corruptions on large-scale LiDAR point clouds from the complex outdoor driving scenes \cite{geiger2012we,SemanticKITTI,nuScenes,WaymoOpen}.

As shown in~\cref{fig:taxonomy}, we consider three distinct corruption sources that are with a high likelihood to occur in real-world deployment: \textit{\textbf{1) Severe weather conditions}} (\textit{fog}, \textit{rain}, and \textit{snow}) which cause back-scattering, attenuation, and reflection of the laser pulses \cite{Fog,Snow_Wet,Wet_Surface}; \textit{\textbf{2) External disturbances}}, \textit{e.g.}, bumpy surfaces, dust, insects, that often lead to nonnegligible \textit{motion blur} and LiDAR \textit{beam missing} issues~\cite{Dust}; and \textit{\textbf{3) Internal sensor failure}}, such as the  \textit{incomplete echo} or miss detection of instances with a dark color (\textit{e.g.}, black car) and \textit{crosstalk} among multiple sensors, which likely deteriorates the 3D perception accuracy \cite{lidar_camera_fusion,Crosstalk}. Besides the environmental factors, it is also important to understand the \textit{cross-sensor} discrepancy to avoid sudden failure caused by the sensor configuration change.

To properly fulfill such pursues, we simulate physically-principled corruptions on the \textit{val} sets of KITTI~\cite{geiger2012we}, SemanticKITTI~\cite{SemanticKITTI}, nuScenes~\cite{nuScenes}, and Waymo Oepn~\cite{WaymoOpen}, as our corruption suite dubbed \textbf{\textit{Robo3D}}. Analogous to the popular 2D corruption benchmarks~\cite{ImageNet-C,Kinetics-C,michaelis2019dragon}, we create three severity levels for each corruption and design suitable metrics as the main indicator for robustness comparisons. Finally, we conduct exhaustive experiments to understand the pros and cons of different designs from existing models. We observe that modern 3D perception models are at risk of being vulnerable even though their performance on standard benchmarks is improving. Through fine-grained analyses on a wide range of 3D perception datasets, we diagnose that:
\begin{itemize}
    \item \textit{Sensor setups have direct impacts on feature learning}. 3D perception models trained on data collected with different sensor configurations and protocols often yield inconsistent resilience.
    \item \textit{3D data representations often coupled with the model's robustness}. The voxel and point-voxel fusion approaches exhibit clear superiority over the projection-based methods, \textit{e.g.}, range view.
    \item \textit{3D detectors and segmentors show distinct sensitivities to different corruption types}. A sophisticated combination of both tasks is a viable way to achieve robust and reliable 3D perception.
    \item \textit{Out-of-context augmentation (OCA) and flexible rasterization strategies can improve model's robustness}. We thus propose a solution to enhance the robustness of existing 3D perception models, which consists of a density-insensitive training framework and a simple flexible voxelization strategy.
\end{itemize}

The key contributions of this work are summarized as:
\begin{itemize}
    \item We introduce {Robo3D}, the first systematically-designed robustness evaluation suite for LiDAR-based 3D perception under corruptions and sensor failure.
    \item We benchmark 34 perception models for LiDAR-based semantic segmentation and 3D object detection tasks, on their robustness against corruptions.
    \item Based on our observations, we draw in-depth discussions on the design receipt and propose novel techniques for building more robust 3D perception models.
\end{itemize}

\section{Related Work}
\label{sec:related}

\noindent\textbf{LiDAR-based Semantic Segmentation}. The design choice of 3D segmentors often correlates with the LiDAR representations, which can be categorized into raw point \cite{thomas2019kpconv}, range view \cite{wu2018squeezeseg,milioto2019rangenet++,kong2023rethinking}, bird's eye view~\cite{zhang2020polarnet}, voxel \cite{2019Minkowski}, and multi-view fusion \cite{2020AMVNet,xu2021rpvnet} methods. The projection-based approach rasterizes irregular point clouds into 2D grids, which avoids the need for 3D operators and is thus more hardware-friendly for deployment \cite{cortinhal2020salsanext,zhao2021fidnet,cheng2022cenet}. The voxel-based methods which retain the 3D structure are achieving better performance than other single modalities~\cite{zhu2021cylindrical,2022GASN}. Efficient operators like the sparse convolution are widely adopted to ease the memory footprint \cite{tang2020searching,tang2022torchsparse}. Most recently, some works start to explore possible complementary between two views \cite{2020AMVNet,2022_2DPASS,qiu2022GFNet} or even more views \cite{xu2021rpvnet}.
Although promising results have been achieved, the robustness of 3D segmentors against corruptions remains obscure. As we will discuss in the next sections, these methods have the tendency of being less robust, mainly due to the lack of a comprehensive robustness evaluation benchmark.

\noindent\textbf{LiDAR-based 3D Object Detection}.
Sharing similar basics with LiDAR segmentation, modern 3D object detectors also adopt various data representations. Point-based methods \cite{pointrcnn,pointgnn,std,3dssd} implicitly capture local structures and fine-grained patterns without any quantization to retain the original point cloud geometry. Voxel-based methods \cite{second,voxelnet,centerpoint,parta2,li2022homogeneous,lidarrcnn,li2023logonet,ma2023detzero} transform irregular point clouds to compact grids while only those non-empty voxels are stored and utilized for feature extraction through the sparse convolution \cite{second}. Recently, some works \cite{votr,centerformer} start to explore long-range contextual dependencies among voxels with self-attention~\cite{2017transformer}. The pillar-based methods \cite{lang2019pointpillars,pillarnet} better balance the accuracy and speed by controlling the resolution in the vertical axis. The point-voxel fusion methods \cite{shi2022pv,pvrcnn} can integrate the merits of both representations to learn more discriminative features. The above methods, however, mainly focused on obtaining better performance on clean point clouds, while paying much less attention to the model's robustness. As we will show in the following sections, these models are prone to degradation under data corruptions and sensor failure.

\noindent\textbf{Common Corruptions}. The corruption robustness often refers to the capability of a conventionally trained model for maintaining satisfactory performance under natural distribution shifts. ImageNet-C \cite{ImageNet-C} is the pioneering work in this line of research which benchmarks image classification models to common corruptions and perturbations. Follow-up studies extend on a similar aspect to other perception tasks, \textit{e.g.}, object detection \cite{michaelis2019dragon}, image segmentation \cite{Cityscapes-C}, navigation \cite{RobustNav}, video classification \cite{Kinetics-C}, and pose estimation~\cite{AdvMix}. The importance of evaluating the model's robustness against corruptions has been constantly proven. Since we are targeting a different sensor, \textit{i.e.}, LiDAR, most of the well-studied corruption types -- such as those designed for camera malfunctions -- become unrealistic or unsuitable for such a data format. This motivates us to explore new taxonomy for defining more proper corruption types for the 3D perception tasks in autonomous driving scenarios.

\noindent\textbf{3D Perception Robustness}. 
Several recent attempts proposed to investigate the vulnerability of point cloud classifiers and detectors in indoor scenes \cite{3dcc,ren2022modelnet-c,sun,PV-Ada,alliegro2022open}. Recently, there are works started to explore the robustness of 3D object detectors under adversarial attacks \cite{sensor_adversarial_traits,tu2020physically,xie2023adversarial}. In the context of corruption robustness, we notice several concurrent works \cite{lidar_camera_fusion,li2021common,albreiki2022robustness,dong2023benchmarking,yan2023benchmarking}. These works, however, all consider a single task alone and might be constrained by either a limited number of corruption types or candidate datasets. Our benchmark properly defines a more diverse range of corruption types for the general 3D perception task and includes significantly more models from both LiDAR-based semantic segmentation and 3D object detection tasks.

\section{The Robo3D Benchmark}
\label{sec:benchmark}

Tailored for LiDAR-based 3D perception tasks, we summarize eight corruption types commonly occurring in real-world deployment in our benchmark, as shown in~\cref{fig:taxonomy}. This section elaborates on the detailed definition of each corruption type (\cref{sec:corruption_types}), configurations of different robustness simulation sets (\cref{sec:corruption_sets}), and evaluation metrics for robustness measurements (\cref{sec:evaluation_metrics}).
 
\subsection{Corruption Types}
\label{sec:corruption_types}
Given a point $\mathbf{p}\in\mathbb{R}^4$ in a LiDAR point cloud with coordinates $(p^x, p^y, p^z)$ and intensity $p^i$, our goal is to simulate a corrupted point $\mathbf{\hat{p}}$ via a mapping $\mathbf{\hat{p}} = \mathcal{C}(\mathbf{p})$, with rules constrained by \textit{physical principles} or \textit{engineering experiences}. Due to space limits, We present more detailed definitions and implementation procedures of our corruption simulation algorithms in the Appendix.

\noindent\textit{\textbf{1) Fog}}.
The LiDAR sensor emits laser pulses for accurate range measurement. Back-scattering and attenuation of LiDAR points tend to happen in foggy weather since the water particles in the air will cause inevitable pulse reflection \cite{STF,Fog_benchmark}. In our benchmark, we adopt the physically valid fog simulation method \cite{Fog} to create fog-corrupted data. For each $\mathbf{p}$, we calculate its attenuated response $p^{i_{\text{hard}}}$ and the maximum fog response $p^{i_{\text{soft}}}$ as follows:
\begin{equation}
p^{i_{\text{hard}}} = 
p^ie^{-2\alpha\sqrt{(p^x)^2 +(p^y)^2 + (p^z)^2}}~,
\end{equation}
\begin{equation}
p^{i_{\text{soft}}} =
p^i\frac{(p^x)^2 +(p^y)^2 + (p^z)^2}{\beta_0}\beta_{\text{bs}}\times{p}^i_{\text{tmp}}~,
\end{equation}
\begin{equation}    
\mathbf{\hat{p}} = \mathcal{C}_{\text{fog}}(\mathbf{p}) = 
\begin{cases}
(\hat{p}^x, \hat{p}^y, \hat{p}^z, p^{i_{\text{soft}}}), & \text{if~~$p^{i_{\text{soft}}} > p^{i_{\text{hard}}}$}~, \\
(p^x, p^y, p^z, p^{i_{\text{hard}}}), & \text{else}~. 
\end{cases}
\end{equation}
where $\alpha$ is the attenuation coefficient, $\beta_{\text{bs}}$ denotes the back-scattering coefficient, $\beta_0$ describes the differential reflectivity of the target objects, and the ${p}^i_{\text{tmp}}$ symbol is the received response for the soft target term.

\noindent\textit{\textbf{2) Wet Ground}}.
The emitted laser pulses will likely lose certain amounts of energy when hitting wet surfaces, which causes significantly attenuated laser echoes depending on the water height $d_w$ and mirror refraction rate \cite{Wet_Surface}. We follow \cite{Snow_Wet} to model the attenuation caused by ground wetness. 
A pre-processing step is taken to estimate the ground plane with existing semantic labels or RANSAC \cite{fischler1981random}. Next, a ground plane point of its measured intensity $\hat{p}^i$ is obtained based on the modified reflectivity, and the point is only kept if its intensity is greater than the noise floor $i_n$ via the following mapping:
\begin{equation}    
\mathcal{C}_{\text{wet}}(\mathbf{p}) =
\begin{cases}
(p^x, p^y, p^z, \hat{p}^i), & \text{if~~~~$\hat{p}^{i} > i_n$ \text{\&~~$\mathbf{p}\in$ ground}}~,\\
\text{None}, & \text{elif~$\hat{p}^{i} < i_n$ \text{\&~~$\mathbf{p}\in$ ground}}~,\\ 
(p^x, p^y, p^z, p^i), & \text{elif~$\mathbf{p}\notin$ \text{ground}}~.
\end{cases}
\end{equation}

\noindent\textit{\textbf{3) Snow}}.
For each laser beam in snowy weather, the set of particles in the air will intersect with it and derive the angle of the beam cross-section
that is reflected by each particle, taking potential occlusions into account~\cite{4DenoiseNet}. We follow~\cite{Snow_Wet} to simulate snow-corrupted data $\mathcal{C}_{\text{snow}}(\mathbf{p})$ which is similar to the fog simulation. This physically-based method samples snow particles in the 2D space and modify the measurement for each LiDAR beam in accordance with the induced geometry, where the number of sampling snow particles is set according to a given snowfall rate $r_s$.

\noindent\textit{\textbf{4) Motion Blur}}. Since the LiDAR sensor is often mounted on the roof-top or side of the vehicle, it inevitably suffers from the blur caused by vehicle movement, especially on bumpy surfaces or during U-turning. To simulate blur-corrupted data  $\mathcal{C}_{\text{motion}}(\mathbf{p})$, we add a jittering noise to each coordinate $(p^x, p^y, p^z)$ with a translation value sampled from the Gaussian distribution with standard deviation $\sigma_{t}$. This simulation process is shown as follows:
\begin{equation}    
\mathcal{C}_{\text{motion}}
(\mathbf{p}) = (p^x+o_1, p^y+o_2, p^z+o_3, p^i)~,
\end{equation}
where $o_1,o_2, o_3$ are the random offsets sampled from Gaussian distribution $N \in \{0, {\sigma_{t}}^2 \}$ and $\{o_1,o_2, o_3\} \in\mathbb{R}^{1\times1 }$.

\noindent\textit{\textbf{5) Beam Missing}}. The dust and insect tend to form agglomerates in front of the LiDAR surface and will not likely disappear without human intervention, such as drying and cleaning~\cite{Dust}. This type of occlusion causes zero readings on masked areas and results in the loss of certain light impulses. To mimic such a behavior, we randomly sample a total number of $m$ beams and drop points on these beams from the original point cloud to generate $\mathcal{C}_{\text{beam}}(\mathbf{p})$:
\begin{equation}    
\mathcal{C}_{\text{beam}}(\mathbf{p}) =
\begin{cases}
(p^x, p^y, p^z, p^i), & \text{if~~~~$\mathbf{p} \notin m $ },\\
\text{None}, & \text{else}~.
\end{cases}
\end{equation}

\noindent\textit{\textbf{6) Crosstalk}}. Considering that the road is often shared by multiple vehicles, the time-of-flight of light impulses from one sensor might interfere with impulses from other sensors within  a similar frequency range~\cite{Crosstalk}. Such a crosstalk phenomenon often creates noisy points within the mid-range areas in between two (or multiple) sensors. To simulate this corruption $\mathcal{C}_{\text{cross}}(\mathbf{p})$, we randomly sample a subset of $k_t$ percent points from the original point cloud and add large jittering noise with a translation value sampled from the Gaussian distribution with standard deviation $\sigma_{c}$. This simulation process is shown as follows:
\begin{equation}    
\mathcal{C}_{\text{cross}}(\mathbf{p}) =
\begin{cases}
(p^x, p^y, p^z, p^i), & \text{if~~$\mathbf{p} \notin$ \text{set of \{$k_t$\} }},\\
(p^x, p^y, p^z, p^i) + \xi_c, & \text{else}~,
\end{cases}
\end{equation}
where $\xi_c$ is the random offset sampled from Gaussian distribution $N \in \{0, {\sigma_{c}}^2 \}$ and $\xi_c \in\mathbb{R}^{1\times4 }$. 

\noindent\textit{\textbf{7) Incomplete Echo}}. The near-infrared spectrum of the laser pulse emitted from the LiDAR sensor is vulnerable to vehicles or other instances with dark colors~\cite{lidar_camera_fusion}. The LiDAR readings are thus incomplete in such scan echoes, resulting in significant point miss detection. We simulate this corruption which denotes $\mathcal{C}_{\text{echo}}(\mathbf{p})$ by randomly querying $k_e$ percent points for \textit{vehicle}, \textit{bicycle}, and \textit{motorcycle} classes, via either semantic masks or 3D bounding boxes. Next, we drop the queried points from the original point cloud, along with their point-level semantic labels. Note that we do not alter the ground-truth bounding boxes since they should remain at their original positions in the real world. The overall operation can be summarized as follows:
\begin{equation}    
\mathcal{C}_{\text{echo}}(\mathbf{p}) =
\begin{cases}
(p^x, p^y, p^z, p^i), & \text{if~~$\mathbf{p} \notin$ \text{set of \{$k_e$\} }},\\
\text{None}, & \text{else}~.
\end{cases}
\end{equation}

\noindent\textit{\textbf{8) Cross-Sensor}}. Due to the large variety of LiDAR sensor configurations (\textit{e.g.}, beam number, FOV, and sampling frequency), it is important to design robust 3D perception  models that are capable of maintaining satisfactory performance under cross-device cases~\cite{std}. While previous works directly form such settings with two different datasets, the domain idiosyncrasy in between (\textit{e.g.} different label mappings and data collection protocols) further hinders the direct robustness comparison. In our benchmark, we follow \cite{wei2022distillation} and generate cross-sensor data $\mathcal{C}_{\text{sensor}}(\mathbf{p})$ by first dropping points of certain beams from the point cloud and then sub-sample $k_c$ percent points from each beam. This simulation process is shown as follows:
\begin{equation}    
\mathcal{C}_{\text{sensor}}(\mathbf{p}) =
\begin{cases}
\text{None},  & \text{if~~$ \text{$\mathbf{p}\in$ set of \{$k_c$\}} $
}, \\
(p^x, p^y, p^z, p^i), & \text{else}~.
\end{cases}
\end{equation}

\subsection{Corruption Sets}
\label{sec:corruption_sets}
Following the above taxonomy, we create new robustness evaluation sets upon the \textit{val} sets of existing large-scale 3D perception datasets \cite{geiger2012we,SemanticKITTI,nuScenes, Panoptic-nuScenes,WaymoOpen} to fulfill \textit{SemanticKITTI-C}, \textit{KITTI-C}, \textit{nuScenes-C}, and \textit{WOD-C}. They are constructed with eight corruption types under three severity levels, resulting in a total number of 97704, 90456, 144456, and 143424 annotated LiDAR point clouds, respectively. Kindly refer to the Appendix for more details in terms of these robustness evaluation collections.

\begin{figure*}
    \centering
    \subfigure[KITTI-C]{
        \includegraphics[width=0.15255\linewidth]{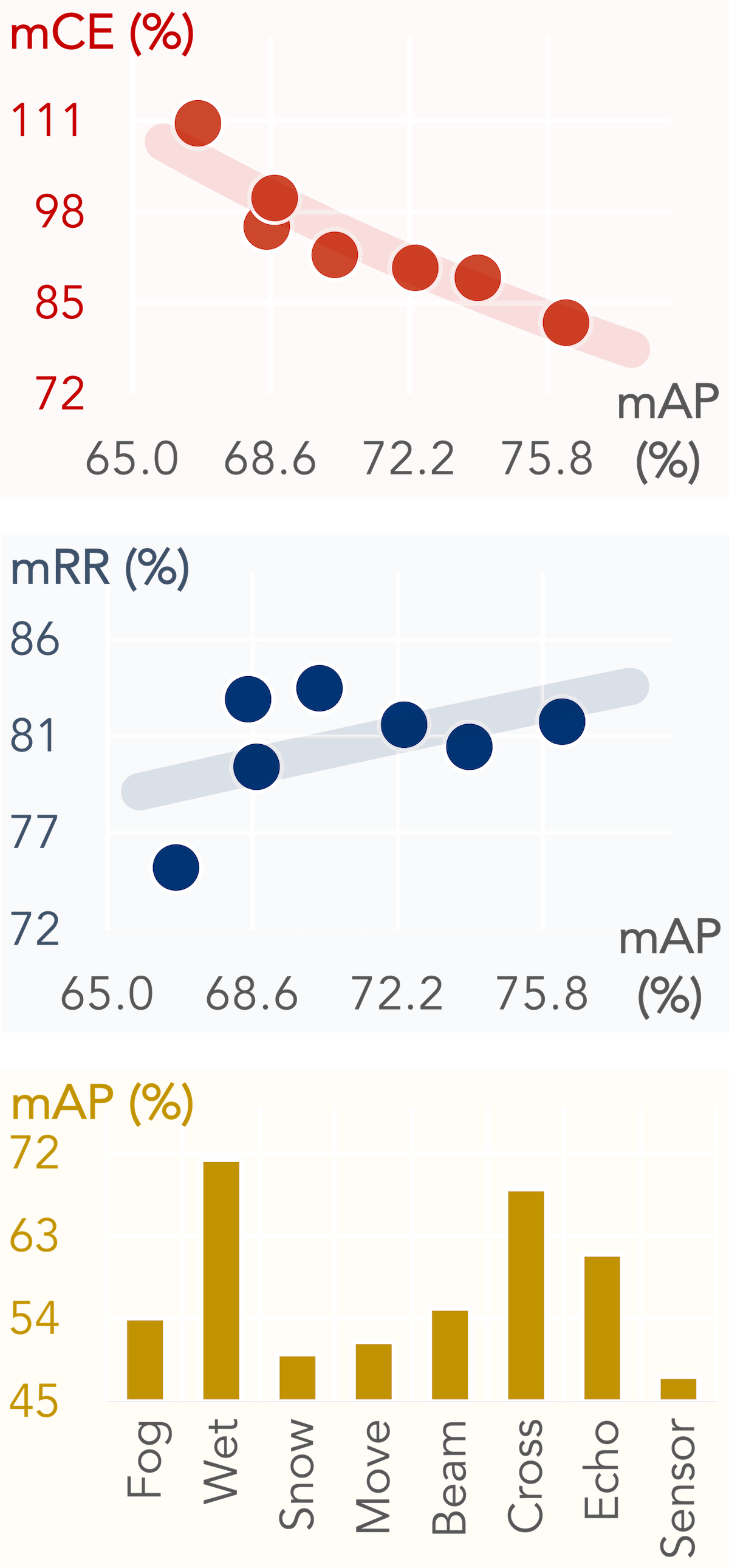}
    }
    \subfigure[SemanticKITTI-C]{
        \includegraphics[width=0.15255\linewidth]{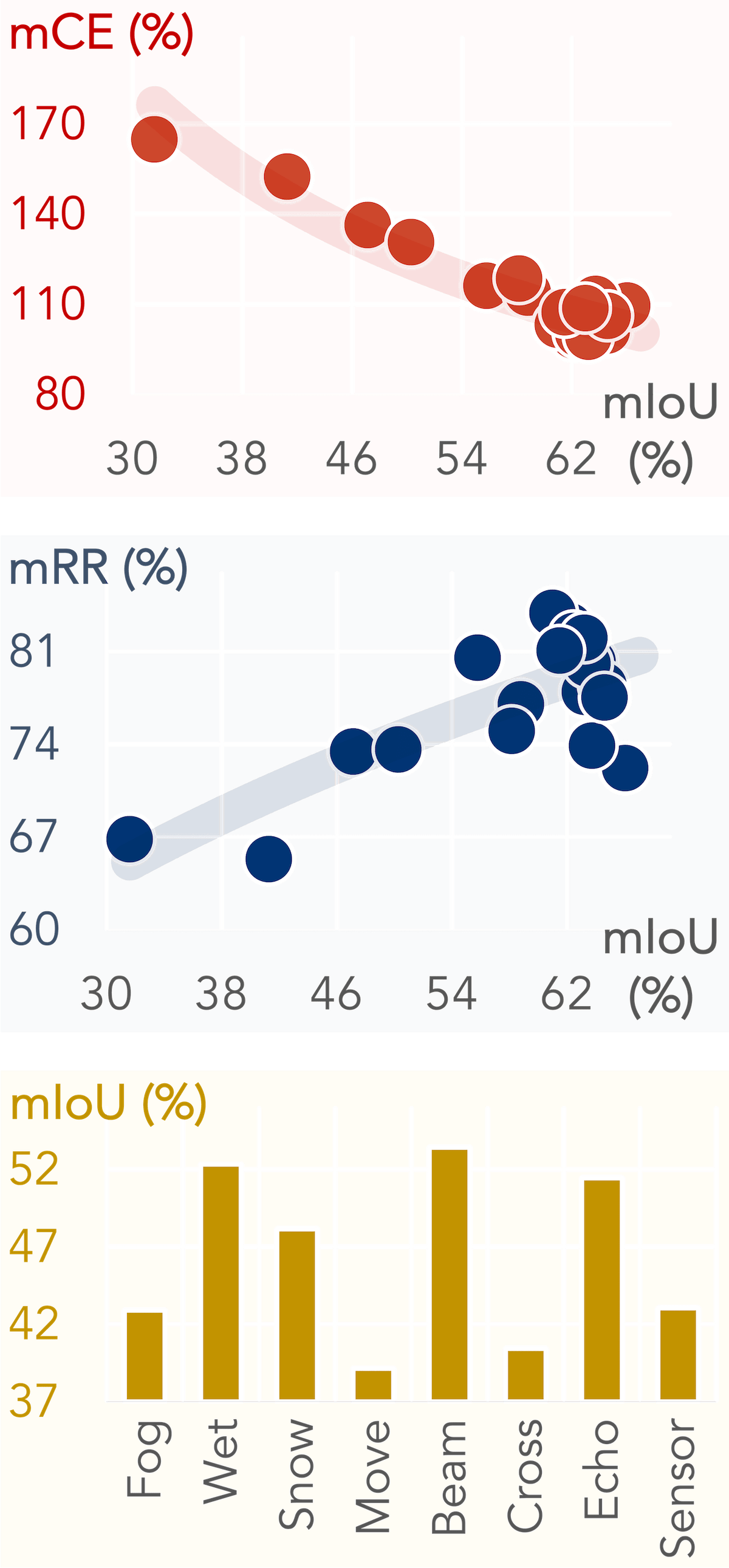}
    }
    \subfigure[nuScenes-C~(Det3D)]{
        \includegraphics[width=0.15255\linewidth]{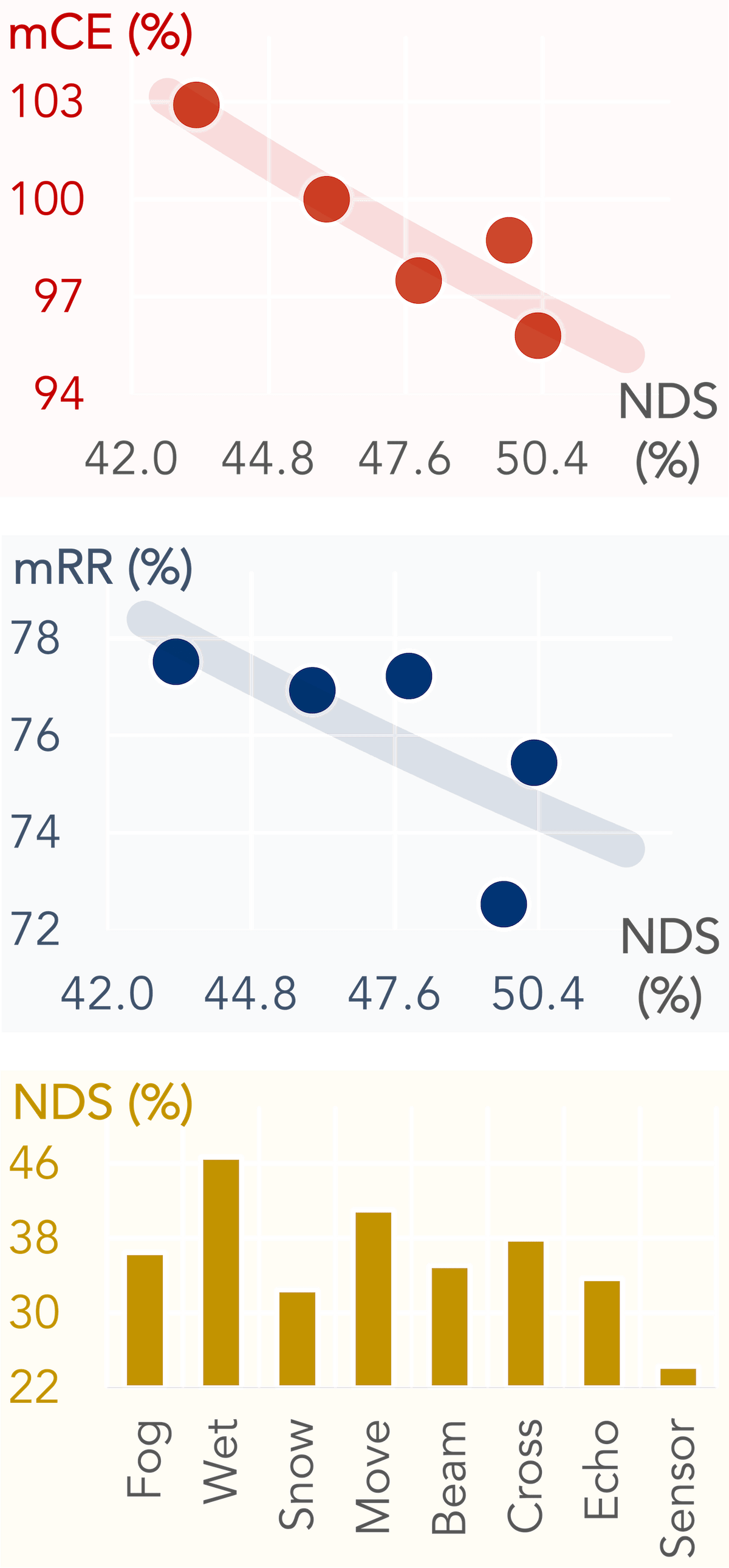}
    }
    \subfigure[nuScenes-C~(Seg3D)]{
        \includegraphics[width=0.15255\linewidth]{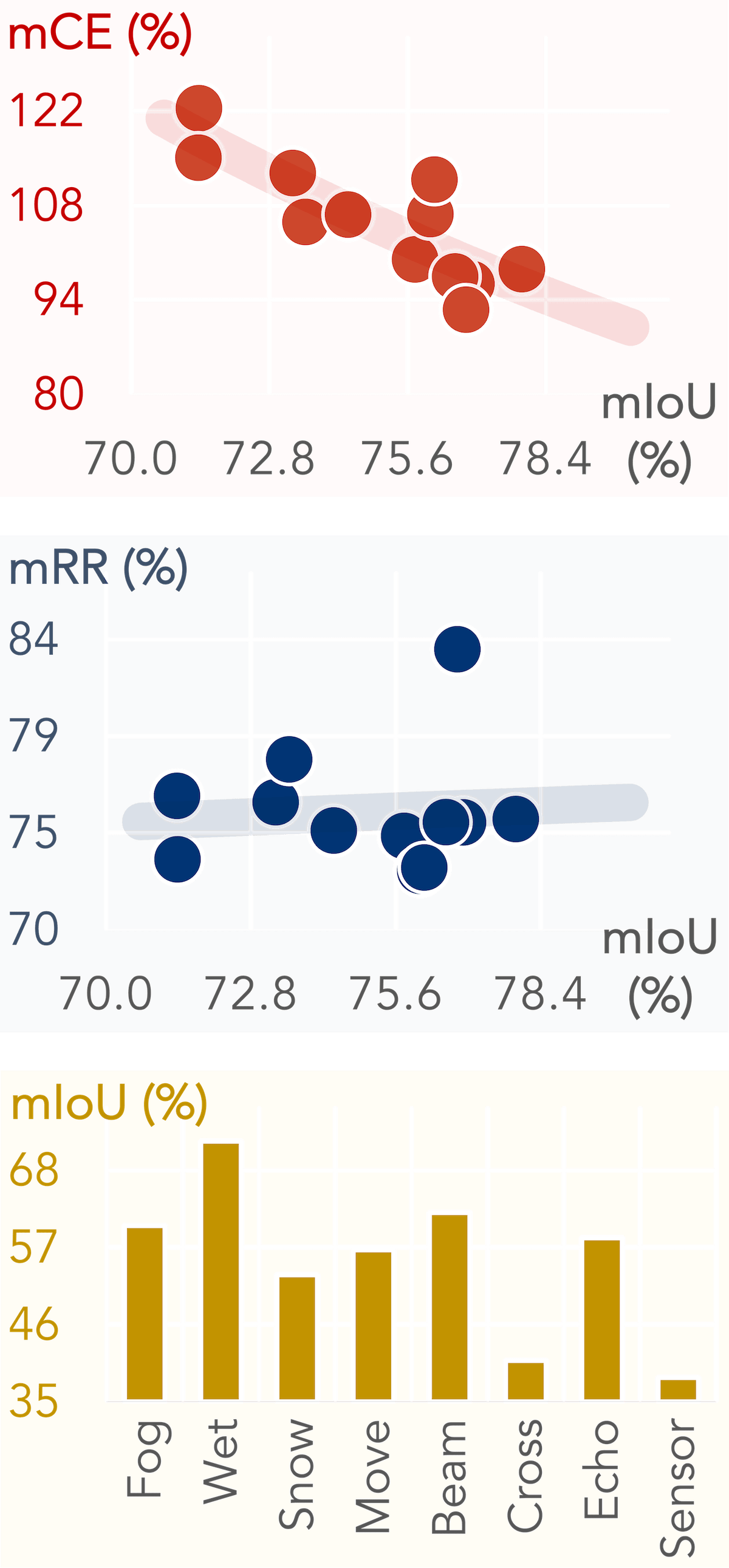}
    }
    \subfigure[WOD-C~(Det3D)]{
        \includegraphics[width=0.15255\linewidth]{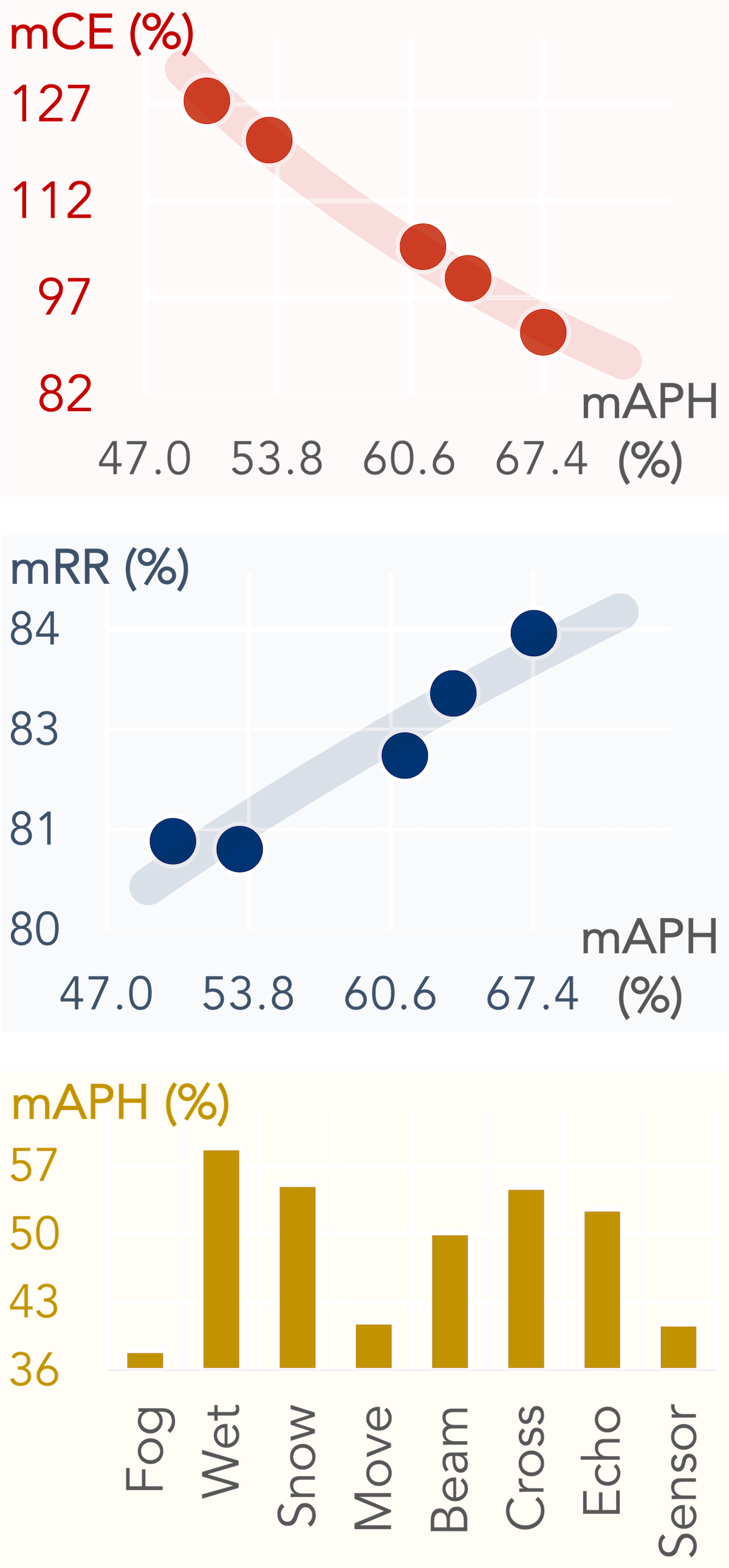}
    }
    \subfigure[WOD-C~(Seg3D)]{
        \includegraphics[width=0.15255\linewidth]{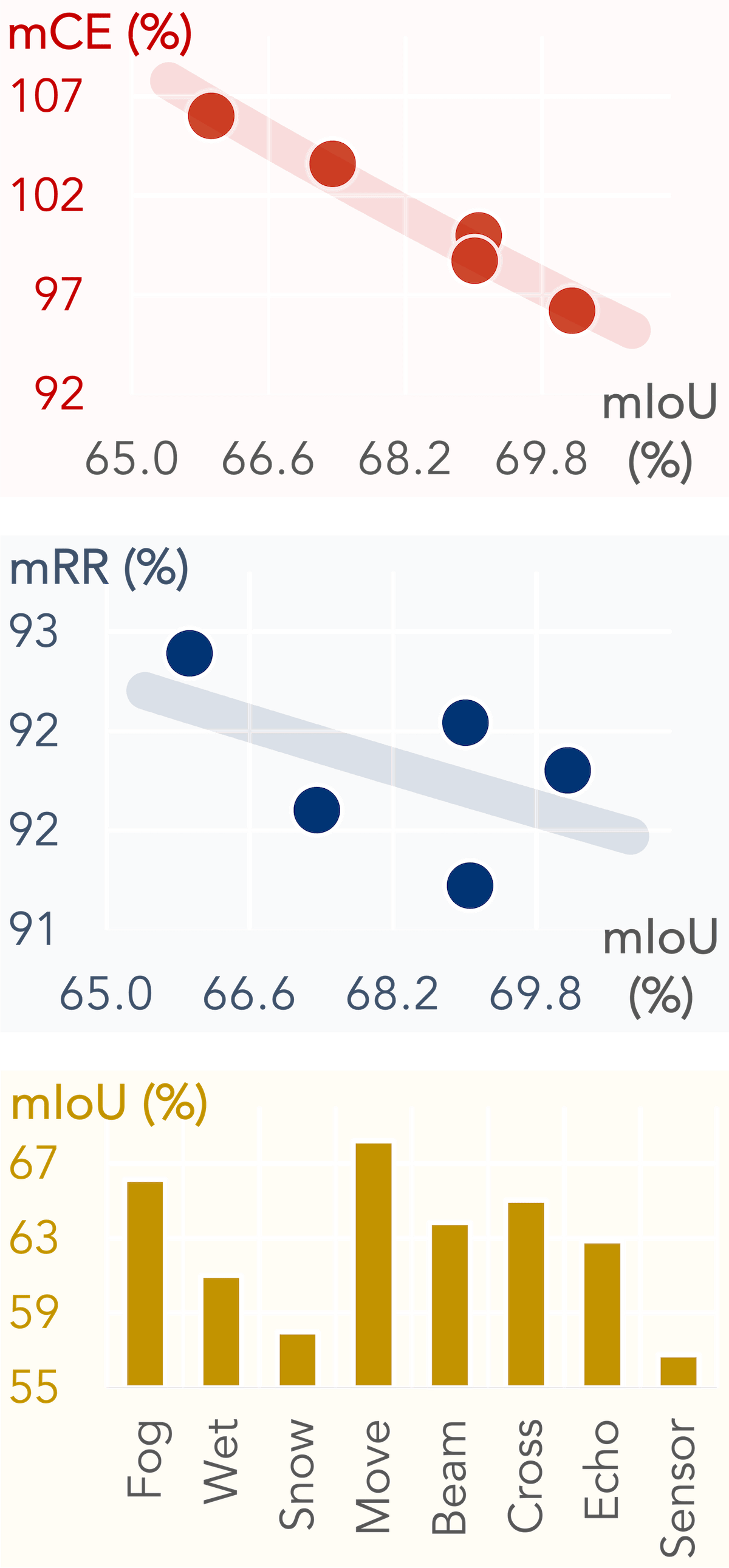}
    }
    \vspace{-0.2cm}
    \caption{Benchmarking results of \textbf{\textit{34}} LiDAR-based detection and segmentation models on the \textbf{\textit{six}} robustness sets in Robo3D. Figures from top to bottom: the task-specific accuracy (mAP, mIoU, NDS, mAPH) \textit{vs.} \textbf{[first row]} mean corruption error (mCE), \textbf{[second row]} mean resilience rate (mRR), and \textbf{[third row]} sensitivity analysis among different corruption types.}
    \label{fig:robo3d_benchmark}
    \vspace{0.1cm}
\end{figure*}

\subsection{Evaluation Metrics}
\label{sec:evaluation_metrics}
\noindent\textbf{Corruption Error (CE)}. We follow \cite{ImageNet-C} and use the mean CE (mCE) as the primary metric in comparing models' robustness. To normalize the severity effects, we choose CenterPoint \cite{centerpoint} and MinkUNet \cite{tang2020searching} as the baseline models for the 3D detectors and segmentors, respectively. The CE and mCE scores are calculated as follows:
\begin{equation}
    \text{CE}_i=\frac{\sum^{3}_{l=1}(1 - \text{Acc}_{i,l})}{\sum^{3}_{l=1}(1 - \text{Acc}_{i,l}^{\text{baseline}})}~,~~~
    \text{mCE}=\frac{1}{N}\sum^N_{i=1}\text{CE}_i~,
\end{equation}
where $\text{Acc}_{i,l}$ denotes the task-specific accuracy scores, \textit{i.e.}, mIoU for LiDAR semantic segmentation, and AP, NDS, or APH(L2) for 3D object detection, on corruption type $i$ at severity level $l$. $N=8$ is the total number of corruption types.

\noindent\textbf{Resilience Rate (RR)}. We define mean RR (mRR) as the relative robustness indicator for measuring how much accuracy can a model retain when evaluated on the corruption sets. The RR and mRR scores are calculated as follows.
\begin{equation}
    \text{RR}_i=\frac{\sum^{3}_{l=1}\text{Acc}_{i,l}}{3\times \text{Acc}_{\text{clean}}}~,~~~~~
    \text{mRR}=\frac{1}{N}\sum^N_{i=1}\text{RR}_i~,
\end{equation}
where $\text{Acc}_{\text{clean}}$ denotes the task-specific accuracy score on the ``clean" evaluation set.

\begin{figure}[t]
    \begin{center}
    \includegraphics[width=0.479\textwidth]{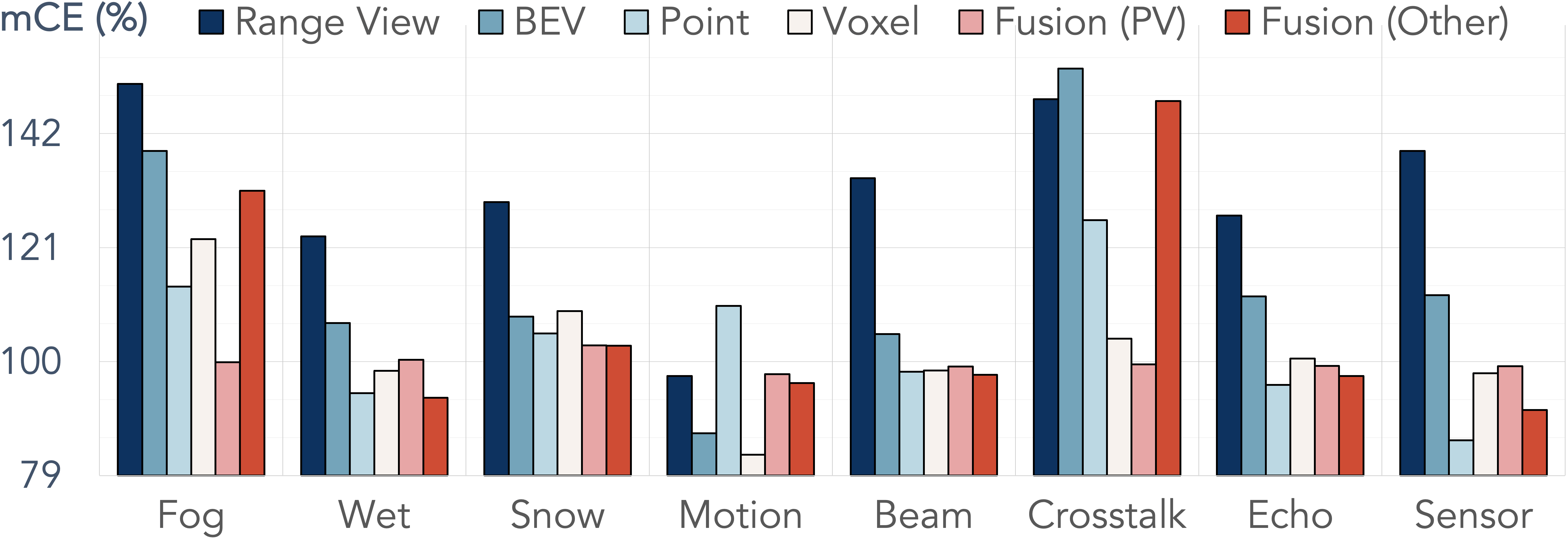}
    \end{center}
    \vspace{-0.4cm}
    \caption{The robustness comparisons among different LiDAR representations (modalities) on \textit{SemanticKITTI-C}.}
    \label{fig:ablation_modality}
\end{figure}

\section{Experimental Analysis}
\label{sec:experiment}

\subsection{Benchmark Configuration}
\noindent{\textbf{3D Perception Models}}. We benchmark 34 LiDAR-based detection and segmentation models and variants. \textit{\textbf{Detectors:}} SECOND \cite{second}, PointPillars \cite{lang2019pointpillars} PointRCNN \cite{pointrcnn}, Part-A$^2$ \cite{parta2}, PV-RCNN \cite{pvrcnn}, CenterPoint \cite{centerpoint}, and PV-RCNN++ \cite{pvrcnn++}. \textit{\textbf{Segmentors:}} SqueezeSeg \cite{wu2018squeezeseg}, SqueezeSegV2 \cite{wu2019squeezesegv2}, RangeNet++ \cite{milioto2019rangenet++}, SalsaNext \cite{cortinhal2020salsanext}, FIDNet \cite{zhao2021fidnet}, CENet \cite{cheng2022cenet}, PolarNet \cite{zhang2020polarnet}, KPConv \cite{thomas2019kpconv}, PIDS \cite{zhang2023pids}, WaffleIron \cite{puy23waffleiron}, MinkUNet \cite{2019Minkowski}, Cylinder3D~\cite{zhu2021cylindrical}, SPVCNN~\cite{tang2020searching}, RPVNet~\cite{xu2021rpvnet}, CPGNet~\cite{li2022cpgnet}, 2DPASS~\cite{2022_2DPASS}, and GFNet~\cite{qiu2022GFNet}. We also include three recent 3D augmentation methods, \textit{i.e.}, Mix3D~\cite{nekrasov2021mix3d}, LaserMix~\cite{kong2023lasermix}, and PolarMix~\cite{xiao2022polarmix}.

\noindent\textbf{Evaluation Protocol}.
Most models benchmarked follow similar data augmentation, pre-training, and validation configurations. We thus directly use public checkpoints for evaluation whenever applicable, or re-train the model following default settings. We notice that some models use extra tricks on original validation sets, \textit{e.g.}, test-time augmentation, model ensemble, \textit{etc}. For such cases, we re-train their models with conventional settings and report the reproduced results. This is to ensure that the robustness comparisons across different models on our corruption sets are fair and convincing. Kindly refer to the Appendix for more details on training and evaluation protocols and to access the pre-trained model weights for reproduction purposes.

\subsection{Benchmark Analysis}
In this section, we draw the following key observations based on the benchmarking results and analyze the potential causes behind them.

\noindent\textbf{O-1: 3D Perception Robustness} - \textit{existing 3D detectors and segmentors are vulnerable to real-world corruptions}. 
As shown in \cref{fig:robo3d_benchmark}, although the models' corruption errors often correlate with the task-specific accuracy (first row), their resilience scores are rather flattened or even descending towards vulnerabilities (second row). The per-corruption errors shown in \cref{tab:semkitti_c_ce} to \cref{tab:wod_c_det3d_ce} further verify such crux. Taking 3D segmentors as an example: although the very recent state-of-the-art methods \cite{2022_2DPASS,qiu2022GFNet,puy23waffleiron} have achieved promising results on the standard benchmark, they are actually less robust than the baseline, \textit{i.e.}, their mCE scores are higher than MinkUNet~\cite{2019Minkowski}. A similar trend appears for the 3D detectors, \textit{e.g.} \cref{fig:robo3d_benchmark}(c), where models with higher NDS are becoming less resilient. Due to the lack of a suitable robustness evaluation benchmark, the existing 3D perception models tend to overfit the ``clean" data distributions rather than realistic scenarios.

\noindent\textbf{O-2: Sensor Configurations} - \textit{models trained with LiDAR data from different sources exhibit inconsistent sensitivities to each corruption type}. As shown in the third row of \cref{fig:robo3d_benchmark}, the same corruption applied on different datasets shows diverse behaviors on model's robustness. Different data collection protocols and sensor setups cause a direct impact on model representation learning. For example, 3D detectors trained on 64-beam datasets (KITTI, WOD) are less robust to \textit{motion blur} and \textit{snow}, compared to their counterparts trained on the sparser dataset (nuScenes). We conjecture that the low-density inputs have incorporated certain resilience for models against noises that occur locally but might become fragile for scenarios that lose points in a global manner, \textit{i.e.}, the \textit{cross-sensor} corruption.

\input{tables/semkitti_ce}
\input{tables/nuscenes_seg_ce}

\noindent\textbf{O-3: Data Representations} - \textit{representing the LiDAR data as raw points, sparse voxels, or the fusion of them tend to yield better robustness}. It can be easily seen from \cref{fig:ablation_modality} that the corruption errors of projection-based methods, \textit{i.e.} range view and BEV, are much higher than other modalities, for almost every corruption type in the benchmark. Such disadvantages also hold for fusion-based models that use a 2D branch, \textit{e.g.}, RPVNet~\cite{xu2021rpvnet} and GFNet~\cite{qiu2022GFNet}. In general, the point-based methods \cite{thomas2019kpconv,puy23waffleiron,zhang2023pids} are more robust to situations where a significant amount of points are missing while suffering from translation, jittering, and outliers. We conjecture that the sub-sampling and local aggregation widely used in existing point-based architectures are natural rescues for point drops and occlusions. Among all representations, voxel/pillar and point-voxel fusion exhibit a clear superiority under various corruption types, as verified in \cref{tab:semkitti_c_ce}, \cref{tab:nuscenes_c_seg_ce}, and \cref{tab:wod_c_seg3d_ce}, respectively. The voxelization processes that quantize the irregular points are conducive to mitigating the local variations and often yield a more steady representation for feature learning.

\input{tables/wod_seg_ce}
\input{tables/kitti_ce}

\begin{figure*}
    \centering
    \subfigure[Voxel Size on \textit{SemanticKITTI-C} (Seg3D)]{
        \includegraphics[width=0.4875\linewidth]{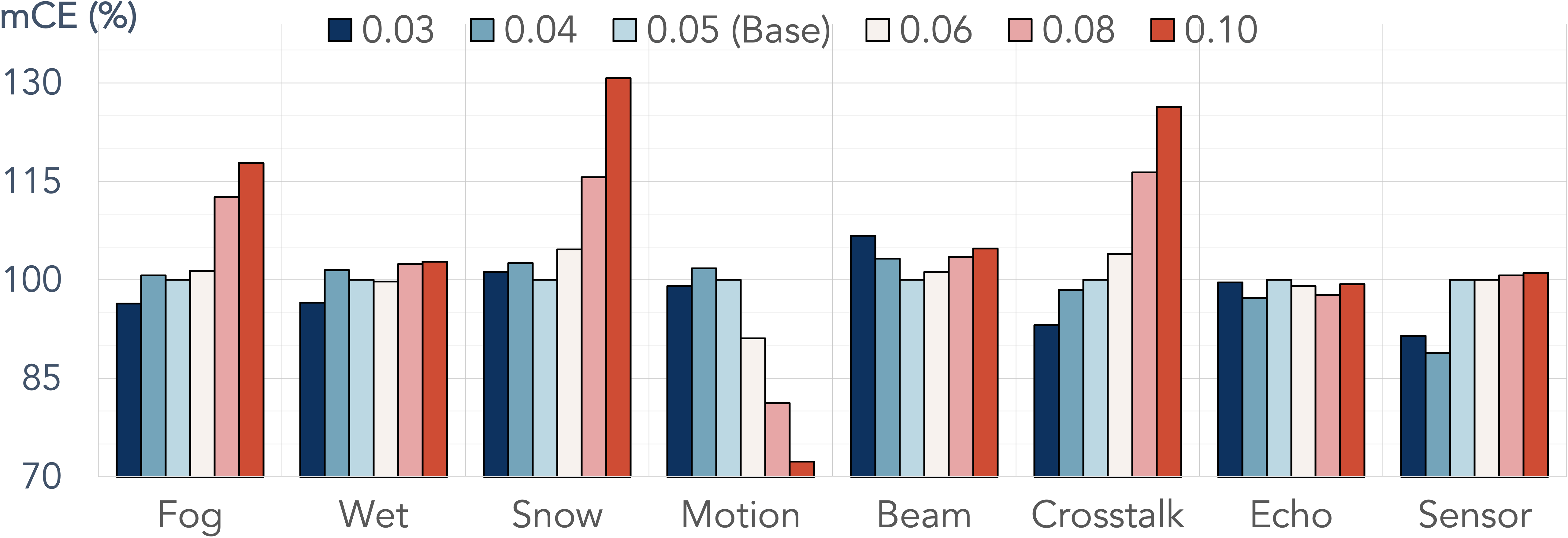}
    }
    \subfigure[Augmentation on \textit{SemanticKITTI-C} (Seg3D)]{
        \includegraphics[width=0.4875\linewidth]{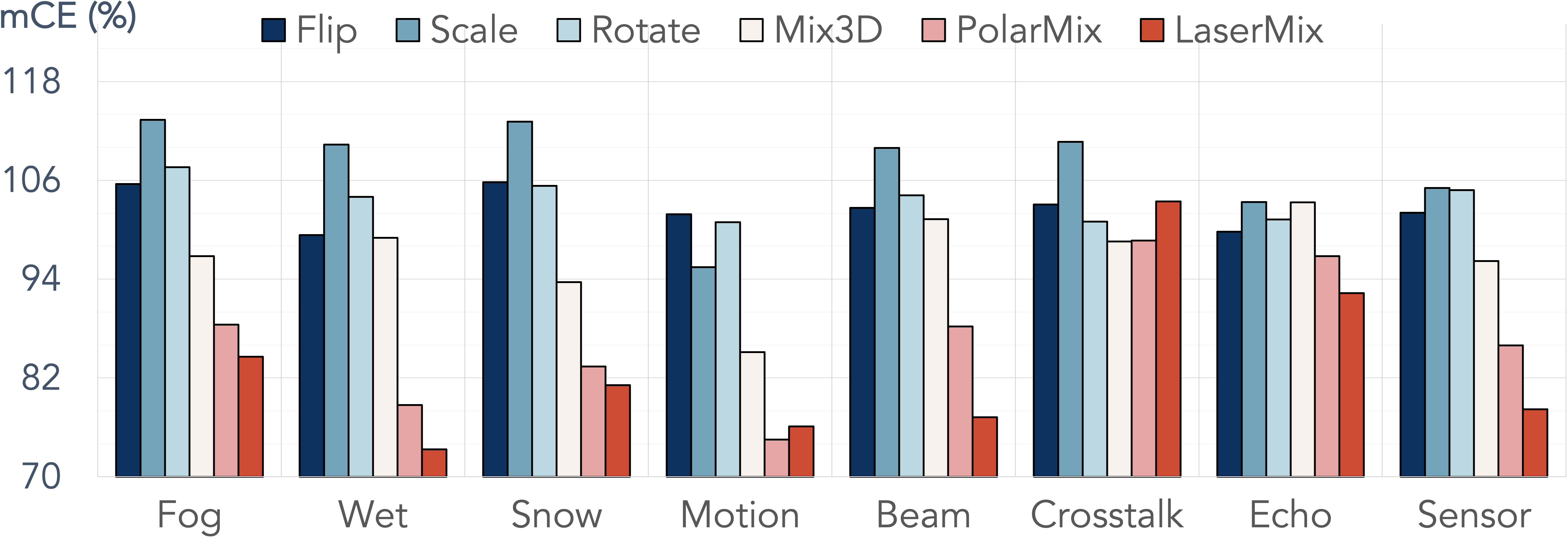}
    }
    \subfigure[Voxel Size on \textit{WOD-C} (Det3D)]{
        \includegraphics[width=0.4875\linewidth]{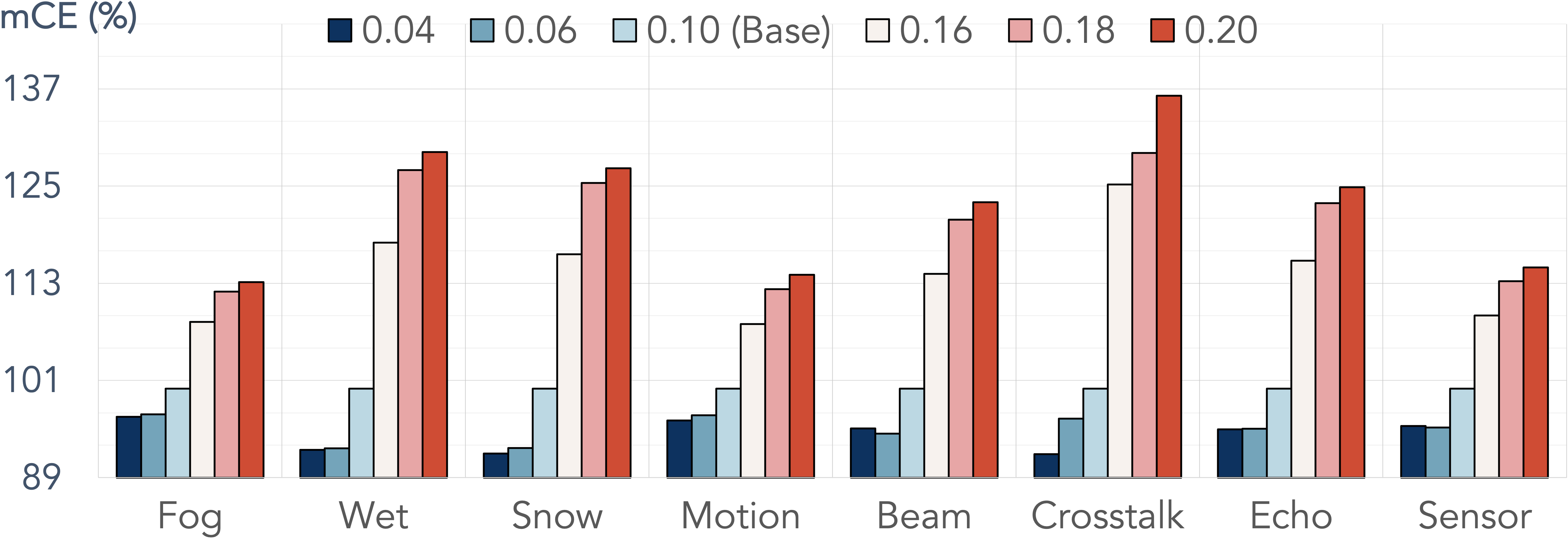}
    }
    \subfigure[Augmentation on \textit{WOD-C} (Det3D)]{
        \includegraphics[width=0.4875\linewidth]{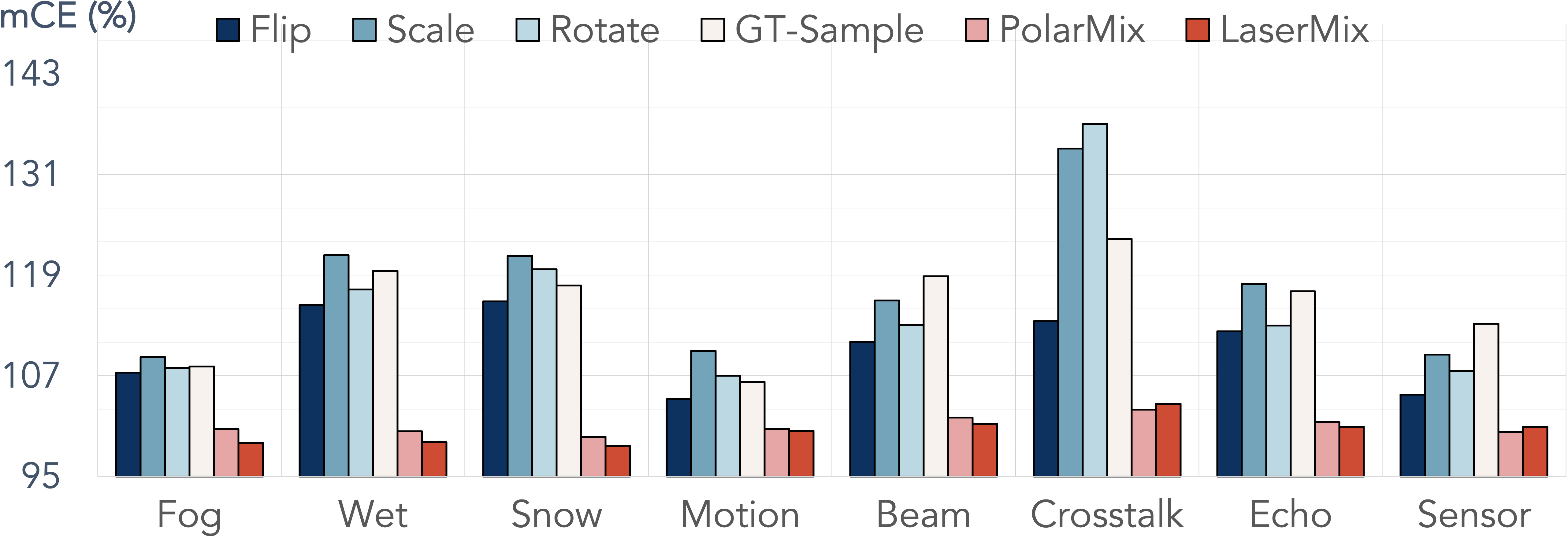}
    }
    \vspace{-0.2cm}
    \caption{Corruption sensitivity analysis on \textit{voxel size} (a \& c) and \textit{augmentation} (b \& d) for the baseline LiDAR semantic segmentation and 3D object detection models~\cite{2019Minkowski,centerpoint}. Different corruptions exhibit variances under certain configurations.}
    \label{fig:pilot_study_vs_aug}
\end{figure*}

\noindent\textbf{O-4: Task Particularity} - \textit{The 3D detectors and segmentors show different sensitivities to corruption scenarios}. 
The detection task only targets classification and localization at the object level; corruptions that occur at points inside an instance range have less impact on detecting the object. However, the segmentation task is to identify the semantic meaning of each point in the point cloud. Such a task discrepancy is affecting the model's robustness across different corruptions. From ~\cref{fig:robo3d_benchmark} we find that 3D detectors tend to be more robust to point-level variations, such as \textit{motion blur} and \textit{crosstalk}. These two corruptions likely yield noise offsets that are out of the grid size; while these point translations could easily be misclassified by the 3D segmentation models. On the contrary, the 3D segmentors are more steady to environmental changes like \textit{fog}, \textit{wet ground}, and \textit{snow}. From hindsight, we believe that a sophisticated combination of the detection and segmentation tasks would be a viable solution for building more robust and reliable 3D perception systems against different corruptions.

\input{tables/nuscenes_det_ce}
\input{tables/wod_det_ce}

\begin{figure*}[t]
    \begin{center}
    \includegraphics[width=1.0\textwidth]{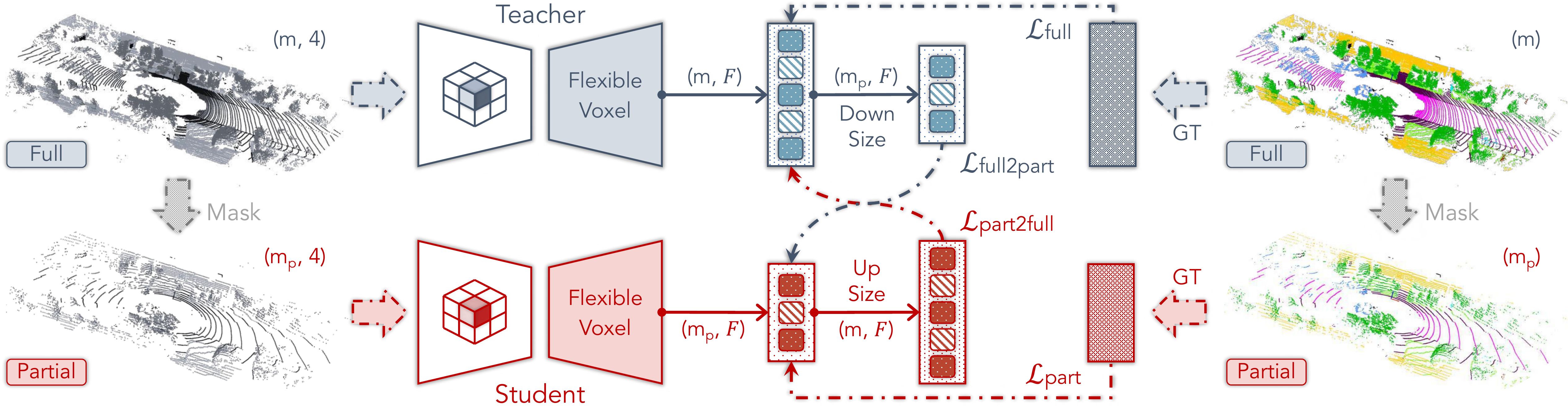}
    \end{center}
    \vspace{-0.4cm}
    \caption{The proposed density-insensitive training framework. The ``full" and ``partial" point clouds are fed into the teacher branch and student branch, respectively, for feature learning, while the latter is generated by randomly masking the original point cloud. To encourage cross-density consistency, we calculate the \textit{completion} and \textit{confirmation} losses which measure the distances of sub-sampled teacher's prediction and interpolated student's prediction between the other branch's outputs.}
    \label{fig:framework}
    \vspace{0.1cm}
\end{figure*}

\noindent\textbf{O-5: Augmentation \& Regularization Effects} - \textit{The recent out-of-context augmentation (OCA) techniques improve 3D robustness by large margins; the flexible rasterization strategies help learn more robust features}. The in-context augmentations (ICAs), \textit{i.e.}, flip, scale, and rotation, are commonly used in 3D detectors and segmentors. Although these techniques help boost perception accuracy, they are less effective in improving robustness. Recent works \cite{nekrasov2021mix3d,kong2023lasermix,xiao2022polarmix} proposed OCAs with the goal of further enhancing model performance on the ``clean" sets. We implement these augmentations on baseline models and test their effectiveness on our robustness evaluation sets, as shown in \cref{fig:pilot_study_vs_aug}~(b) \& (d). Since corrupted data often deviate from the training distribution, the model will inevitably degrade under OoD scenarios. OCAs that mix and swap regions without maintaining the consistency of scene layouts are yielding much lower CE scores across all corruptions, except for \textit{wet ground}, where the loss of ground points restricts the effectiveness of scene mixing. Another key factor that influences the robustness (especially for voxel- and point-voxel fusion-based methods) is representation capacity, \textit{i.e.}, voxel size. As shown in \cref{fig:pilot_study_vs_aug}~(a) \& (c), the 3D segmentors under translations within small regions (\textit{motion blur}) favor a larger voxel size to suppress the global translations; conversely, they are more robust against outliers (\textit{fog}, \textit{snow}, and \textit{crosstalk}) given more fine-grained voxelizations to eliminate the local variations. For 3D detectors, a consensus is formed toward using a higher voxelization resolution, and improvements are constantly achieved across all corruption types in the benchmark.

\section{Boosting Corruption Robustness}
\label{sec:approach}
Motivated by the above observations, we propose two novel techniques to enhance the robustness against corruptions. We conduct experiments on the \textit{SemanticKITTI-C} dataset without loss of generality and include more results on other datasets in the Appendix.

\noindent\textbf{Flexible Voxelization}. The widely used sparse convolution \cite{tang2022torchsparse} requires the formal transformation of the point coordinates $\mathbf{p}_k=(p^x_k, p^y_k, p^z_k)$ into a sparse voxel. This process is often formulated as follows:
\begin{equation}
    \mathbf{v}_k = (v^x_k, v^y_k, v^z_k) = \texttt{floor}((\frac{p^x_k}{l^x}), (\frac{p^y_k}{l^y}), (\frac{p^z_k}{l^z}))~,~
\end{equation}
where $l^x$, $l^y$, and $l^z$ denote the voxel size along each axis and are often set as fixed values. As discussed in \cref{fig:pilot_study_vs_aug}~(a) \& (c), the model tends to show an erratic resilience under different corruptions, \textit{e.g.}, favor a larger voxel size for \textit{motion blur} while is more robust against \textit{fog}, \textit{snow}, and \textit{crosstalk} with a smaller voxel size. To pursue better generalizability among all corruptions, we switch the naive constant into a dynamic alternative $l_{\text{dv}} = (l^x \pm \text{dv}^x, l^y \pm \text{dv}^y, l^z \pm \text{dv}^z)$, where $\text{dv}^x$, $\text{dv}^y$, $\text{dv}^z$ are the offsets sampled from the continuous uniform distribution with an interval $\gamma$.

\begin{figure}[t]
    \begin{center}
    \includegraphics[width=0.48\textwidth]{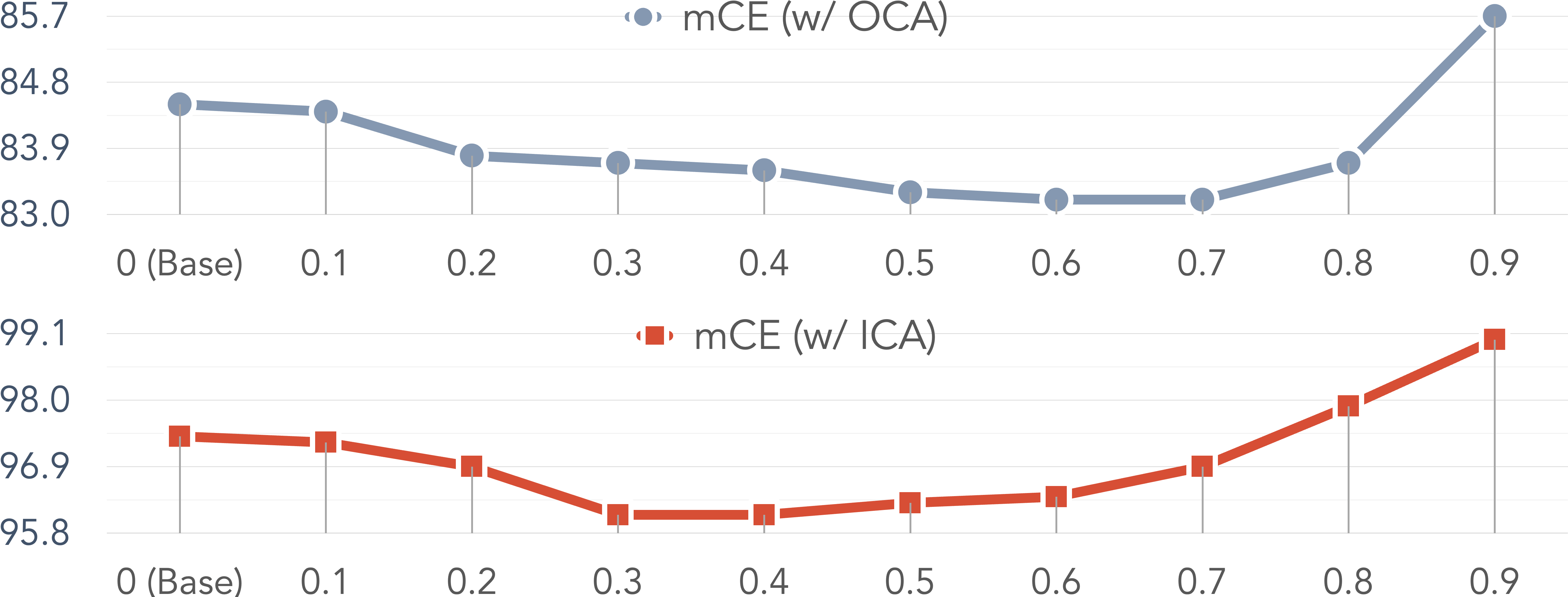}
    \end{center}
    \vspace{-0.3cm}
    \caption{Ablation study on the masking ratio $\beta$ for models: \textbf{[top]} trained \textit{w/} OCA and \textbf{[bottom]} trained \textit{w/} ICA.}
    \label{fig:mask_ratio}
\end{figure}

\noindent\textbf{Density-Insensitive Training}. The natural corruptions often cause severe occlusion, attenuation, and reflection of light impulses, resulting in the unavoidable loss of LiDAR points in certain regions around the ego-vehicle~\cite{Wet_Surface}. For example, the \textit{wet ground} absorbs energy and loses points on the surfaces~\cite{Snow_Wet}; the potential \textit{incomplete echo} and \textit{beam missing} caused by reflection or dust and insects occlusion may lead to serious object failure~\cite{lidar_camera_fusion}. The 3D perception models that suffer from such OoD scenarios bear the risk of being involved in safety-critical issues. It is worth noting that such degradation is not compensable via either adjusting the voxel size or applying OCA (see~\cref{fig:pilot_study_vs_aug}). Inspired by recent masking-based representation methods~\cite{MaskedAutoencoders2021,hess2022maskclip,min2022voxelmae,hess2022voxelmae}, we propose a robust finetuning framework (see \cref{fig:framework}) that tends to be less sensitive to density variations. Specifically, we design a two-branch structure -- a teacher net $\mathcal{G}_{\theta}^{\text{tea}}$ and a student net $\mathcal{G}_{\theta}^{\text{stu}}$ -- that takes a pair of high- and low-density point clouds ($x$ and $\Tilde{x}$) as the input, where the sparser one is generated by randomly masking the points from the original point cloud with a ratio $\beta$. Note that here we use the random mask to sub-sample the given point clouds rather than simulating a specific corruption type defined in our benchmark, since the corruption ``pattern" in the actual scenario is often hard to predict. The loss functions of the $k$-th sample from the ``full" view and the ``partial" view are calculated as follows:
\begin{equation}
    \mathcal{L}_{\text{full}} = \mathcal{L}_{\text{task}}(y_k, \mathcal{G}_{\theta}^{\text{tea}}(x_k))~,~~
    \mathcal{L}_{\text{part}} = \mathcal{L}_{\text{task}}(\Tilde{y}_k, \mathcal{G}_{\theta}^{\text{stu}}(\Tilde{x}_k)),
\end{equation}
where $y_k$ and $\Tilde{y}_k$ are original and masked ground-truths, respectively. $\mathcal{L}_{\text{task}}$ denotes the task-specific loss, \textit{e.g.}, RPN loss for detection and cross-entropy loss for segmentation.

To encourage cross-consistency between the high- and low-density branches, we calculate $\mathcal{L}_{\text{part2full}}$ and $\mathcal{L}_{\text{full2part}}$, where the former is to mimic dense representations from sparse inputs (completion) and the latter is to pursue local agreements (confirmation). The completion loss is calculated as the distance between the teacher net's prediction of the ``full" input and the interpolated student net's prediction of the ``partial" input, which can be calculated as follows:
\begin{equation}
    \mathcal{L}_{\text{part2full}} = ||~ \mathcal{G}_{\theta}^{\text{tea}}(x), ~\texttt{interp}(\mathcal{G}_{\theta}^{\text{stu}}(\Tilde{x})) ~||^2_2 ~.~
\end{equation}
Similarly, the confirmation loss for pursuing local agreements can be calculated as follows:
\begin{equation}
    \mathcal{L}_{\text{full2part}} = ||~ \texttt{subsample}(\mathcal{G}_{\theta}^{\text{tea}}(x)), ~\mathcal{G}_{\theta}^{\text{stu}}(\Tilde{x}) ~||^2_2 ~.~
\end{equation}

The final objective is to optimize the summation of the above loss functions, \textit{i.e.}, $\mathcal{L}=\mathcal{L}_{\text{full}}+\mathcal{L}_{\text{part}}+\alpha_{1}\mathcal{L}_{\text{part2full}}+\alpha_{2}\mathcal{L}_{\text{full2part}}$, where $\alpha_{1}$ and $\alpha_{2}$ are the weight coefficients. 

\noindent\textbf{Implementation Details}. We ablate each component and show the results in \cref{tab:dv_aug_mask}. Specifically, $\gamma$ is set as $0.02$ in our experiments, along with a mask ratio $\beta=0.4$ for models \textit{w/} ICA and $\beta=0.6$ for models \textit{w/} OCA. We initialize both teacher and student networks with the same baseline model and finetune our framework for $6$ epochs in total. The weight coefficients are set as $50$ and $100$, respectively.


\noindent\textbf{Experimental Analysis}. Despite its simplicity, we found this framework is conducive to mitigating robustness degradation from corruptions. The simple modification on voxel partition can boost the corruption robustness by large margins; it reduces $2.6\%$ mCE and $1.5\%$ mCE upon the two baselines, respectively. Then, we incorporate the cross-consistency learning between ``full" and ``partial" views. Among all variants, the one with both completion ($\mathcal{L}_{\text{part2full}}$) and confirmation ($\mathcal{L}_{\text{full2part}}$) objectives achieves the best possible results in terms of mCE and mRR.

We also show an ablation study of the masking ratio $\beta$ in \cref{fig:mask_ratio}. We observe that there is often a trade-off between the model's robustness and the proportion of information occlusion; a ratio between $0.3$ to $0.6$ tends to yield lower mCE (better robustness).
It is worth noting that both flexible voxelization and density-insensitive training will slightly lower the task-specific accuracy on the ``clean" sets, as shown in the last column of \cref{tab:dv_aug_mask}. We conjecture that such an out-of-context consistency regularization will likely relieve the 3D perception model from overfitting the training distribution and in return, become more robust against unseen scenarios from the OoD distribution.

\input{tables/ablation_optim}

\section{Discussion and Conclusion}
In this work, we established a comprehensive evaluation benchmark dubbed \textit{Robo3D} for probing and analyzing the robustness of LiDAR-based 3D perception models. We defined eight distinct corruption types with three severity levels on four large-scale autonomous driving datasets. We systematically benchmarked and analyzed representative 3D detectors and segmentors to understand their resilience under real-world corruptions and sensor failure. Several key insights are drawn from aspects including sensor setups, data representations, task particularity, and augmentation effects. To pursue better robustness, we proposed a cross-density consistency training framework and a simple yet effective flexible voxelization strategy. We hope this work could lay a solid foundation for future research on building more robust and reliable 3D perception models.

\noindent\textbf{Potential Limitation}. Although we benchmarked a wide range of corruptions that occur in the real world, we do not consider cases that are coupled with multiple corruptions at the same time. Besides, we do not include models that take multi-modal inputs, which could form future directions.

\noindent\textbf{Acknowledgements}. This research is part of the programme DesCartes and is supported by the National Research Foundation, Prime Minister’s Office, Singapore under its Campus for Research Excellence and Technological Enterprise (CREATE) programme. This study is supported by the Ministry of Education, Singapore, under its MOE AcRF Tier 2 (MOE-T2EP20221-0012), NTU NAP, and under the RIE2020 Industry Alignment Fund – Industry Collaboration Projects (IAF-ICP) Funding Initiative, as well as cash and in-kind contribution from the industry partner(s). This study is also supported by the National Key R\&D Program of China (No. 2022ZD0161600).

\section*{Appendix}
In this appendix, we supplement more materials to support the findings and conclusions in the main body of this paper. Specifically, this appendix is organized as follows.
\begin{itemize}
    \item \cref{add-sec:study} provides a comprehensive case study for analyzing each of the eight corruption types defined in the Robo3D benchmark.
    \item \cref{add-sec:implementation} elaborates on additional implementation details
for the generation of each corruption type.
    \item \cref{add-sec:experiment} includes additional (complete) experimental results and discussions for the 3D detectors and segmentors benchmarked in Robo3D.
    \item \cref{add-sec:qualitative} attaches qualitative results for the benchmarked methods under each corruption type.
    \item \cref{sec:public-resources-used} acknowledges the public resources used during the course of this work.
\end{itemize}

\section{Case Study: 3D Natural Corruption}
\label{add-sec:study}

The deployment environment of an autonomous driving system is diverse and complicated; any disturbances that occur in the sensing, transmission, or processing stages will cause severe corruptions. In this section, we provide concrete examples of the \textbf{\textit{formation}} and \textbf{\textit{effect}} of the eight corruption types defined in the main body of this paper, \textit{i.e.}, \textit{fog}, \textit{wet ground}, \textit{snow}, \textit{motion blur}, \textit{beam missing}, \textit{crosstalk}, \textit{incomplete echo}, and \textit{cross-sensor}.

Similar to the main body, we denote a point in a LiDAR point cloud as $\mathbf{p}\in\mathbb{R}^4$, which is defined by the point coordinates $(p^x, p^y, p^z)$ and point intensity $p^i$. We aim to simulate a corrupted point $\mathbf{\hat{p}}$ via a mapping $\mathbf{\hat{p}} = \mathcal{C}(\mathbf{p})$, with rules constrained by \textit{physical principles} or \textit{engineering experiences}. The detailed case study for each corruption type defined in the Robo3D benchmark is illustrated as follows.

\subsection{Fog}
\begin{figure}[t]
    \begin{center}
    \includegraphics[width=0.49\textwidth]{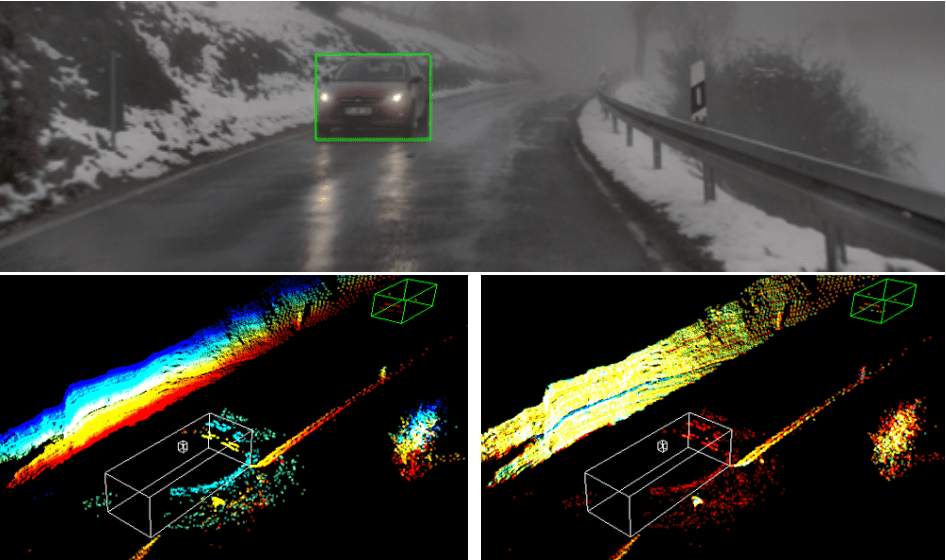}
    \end{center}
    \vspace{-0.4cm}
    \caption{Examples of the data corruptions introduced by \textbf{\textit{fog}}, where the range (bottom left) and intensity (bottom right) distributions are shifted from the uniform distribution of the ego-vehicle. \textbf{\textit{Image credit:}} Hahner \textit{et al.} \cite{Fog}.}
    \label{add-fig:fog}
\end{figure}

The weather phenomena are inevitable in driving scenarios \cite{rasshofer2011weather}. Among them, foggy weather mainly causes back-scattering and attenuation of LiDAR pulse transmissions and results in severe shifts of both range and intensity for the points in a LiDAR point cloud, as shown in \cref{add-fig:fog}.

In this work, we follow Hahner \textit{et al.} \cite{Fog} to generate physically accurate fog-corrupted data using ``clean" datasets. This approach uses a standard linear system \cite{rasshofer2011weather} to model the light pulse transmission under foggy weather. For each $\mathbf{p}$, we calculate its attenuated response $p^{i_{\text{hard}}}$ and the maximum fog response $p^{i_{\text{soft}}}$ as follows:
\begin{equation}
p^{i_{\text{hard}}} = 
p^ie^{-2\alpha\sqrt{(p^x)^2 +(p^y)^2 + (p^z)^2}},
\end{equation}
\begin{equation}
p^{i_{\text{soft}}} =
p^i\frac{(p^x)^2 +(p^y)^2 + (p^z)^2}{\beta_0}\beta\times{p^i}_{tmp}
\end{equation}
\begin{equation}    
\mathbf{\hat{p}} = \mathcal{C}_{\text{fog}}(\mathbf{p}) = 
\begin{cases}
(\hat{p}^x, \hat{p}^y, \hat{p}^z, p^{i_{\text{soft}}}), & \text{if~~$p^{i_{\text{soft}}} > p^{i_{\text{hard}}}$}\text{,} \\
(p^x, p^y, p^z, p^{i_{\text{hard}}}), & \text{else}. 
\end{cases}
\end{equation}
where $\alpha$ is the attenuation coefficient, $\beta$ denotes the back-scattering coefficient, $\beta_0$ describes the differential reflectivity of the target, and the ${p^i}_{tmp}$ is the received response for the soft target term.

\subsection{Wet Ground}
\begin{figure}[t]
    \begin{center}
    \includegraphics[width=0.49\textwidth]{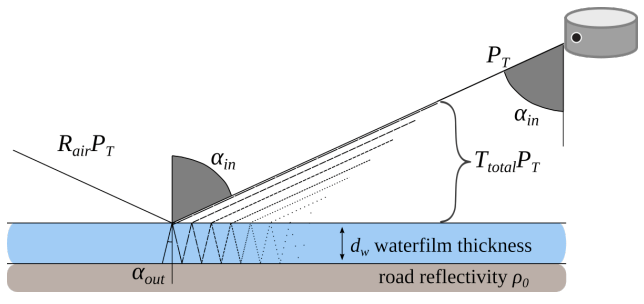}
    \end{center}
    \vspace{-0.5cm}
    \caption{An example of the geometrical optical model of the light pulse reflection in the \textbf{\textit{wet ground}} corruption. Depending on the water height and mirror refraction rate, the pulses emitted by the LiDAR sensor will lose certain amounts of energy when hitting wet surfaces. \textbf{\textit{Image credit:}} Hahner \textit{et al.} \cite{Snow_Wet}.}
    \label{add-fig:wet_ground}
    \vspace{0.2cm}
\end{figure}

As introduced in the main body, the emitted laser pulses from the LiDAR sensor tend to lose certain amounts of energy when hitting wet surfaces, which will cause significantly attenuated laser echoes depending on the water height $d_w$ and mirror refraction rate \cite{Snow_Wet}, as shown in \cref{add-fig:wet_ground}. 

In this work, we follow \cite{Snow_Wet} to model the attenuation caused by ground wetness. 
A pre-processing step is taken to estimate the ground plane with existing semantic labels or RANSAC \cite{fischler1981random}. Next, a ground plane point of its measured intensity $\hat{p}^i$ is obtained based on the modified reflectivity, and the point is only kept if its intensity is greater than the noise floor $i_n$ via mapping:
\begin{equation}    
\mathcal{C}_{\text{wet}}(\mathbf{p}) =
\begin{cases}
(p^x, p^y, p^z, \hat{p}^i), & \text{if~~~~$\hat{p}^{i} > i_n$ \text{\&~~$\mathbf{p}\in$ ground}}~,\\
\text{None}, & \text{elif~$\hat{p}^{i} < i_n$ \text{\&~~$\mathbf{p}\in$ ground}}~,\\ 
(p^x, p^y, p^z, p^i), & \text{elif~$\mathbf{p}\notin$ \text{ground~. }}
\end{cases}
\end{equation}

\begin{figure}[t]
    \begin{center}
    \includegraphics[width=0.49\textwidth]{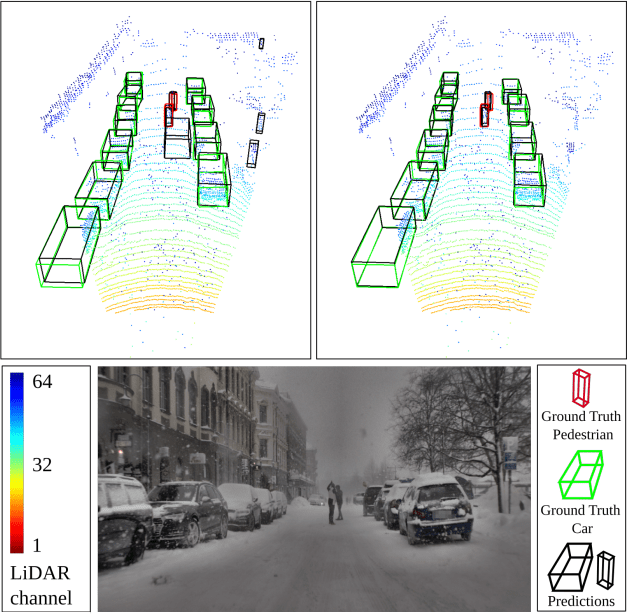}
    \end{center}
    \vspace{-0.4cm}
    \caption{Examples of the data corruptions introduced by \textbf{\textit{snow}}. As shown in the top-left, the particles brought by snowfall will likely cause false predictions for the objects in the 3D scene. \textbf{\textit{Image credit:}} Hahner \textit{et al.} \cite{Snow_Wet}.}
    \label{add-fig:snow}
\end{figure}

\begin{figure*}[!t]
    \begin{center}
    \includegraphics[width=1.0\textwidth]{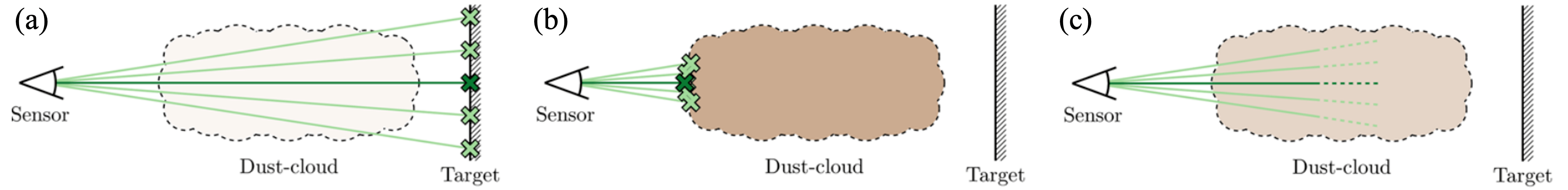}
    \end{center}
    \vspace{-0.4cm}
    \caption{Typical range measurement behaviors that will likely cause \textbf{\textit{beam missing}}. (a) Echoes return from the target (``clean" scenarios). (b) Echoes return from a dusty cloud between the sensor and the target (partial beam missing). (c) No echo returns from either the dusty cloud or the target (complete beam missing). \textbf{\textit{Image credit:}} Phillips \textit{et al.} \cite{Dust}.}
    \label{add-fig:beam_missing}
    \vspace{0.2cm}
\end{figure*}

\begin{figure*}[t]
    \begin{center}
    \includegraphics[width=1.0\textwidth]{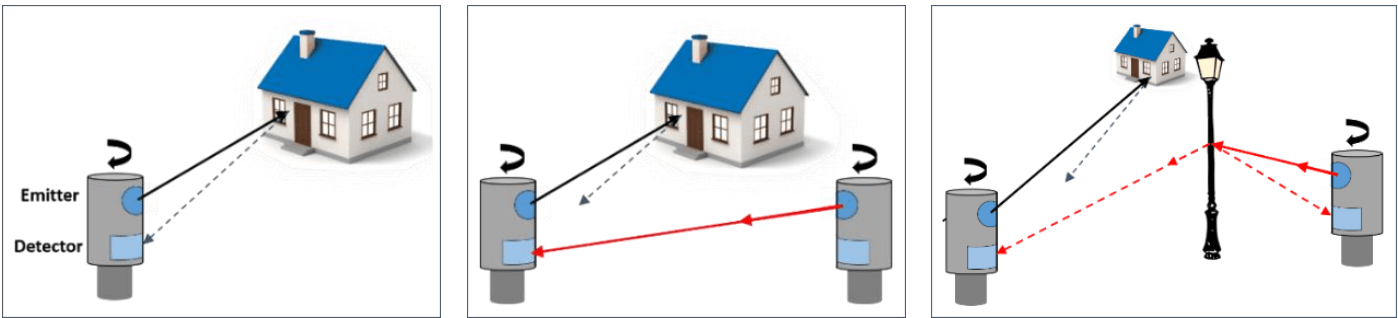}
    \end{center}
    \vspace{-0.4cm}
    \caption{An Illustration of potential \textbf{\textit{crosstalk}} scenarios in a multi-LiDAR system. [Left] The basic principle of range detection in the LiDAR sensing cycle. [Middle] The direct crosstalk scenario in a dual-LiDAR system. [Right] The indirect crosstalk scenario caused by reflection in a dual-LiDAR system. \textbf{\textit{Image credit:}} Diehm \textit{et al.} \cite{Crosstalk2}.}
    \label{add-fig:crosstalk1}
    \vspace{0.2cm}
\end{figure*}

\subsection{Snow}
Snow weather is another adverse weather condition that tends to happen in the real-world environment. For each laser beam in snowy weather, the set of particles in the air will intersect with it and derive the angle of the beam cross-section
that is reflected by each particle, taking potential occlusions into account~\cite{4DenoiseNet}. Some typical examples of the snow-corrupted data are shown in \cref{add-fig:snow}.

In this work, we follow~\cite{Snow_Wet} to simulate these snow-corrupted data $\mathcal{C}_{\text{snow}}(\mathbf{p})$, which is similar to the fog simulation. This physically-based method samples snow particles in the 2D space and modify the measurement for each LiDAR beam in accordance with the induced geometry, where the number of sampling snow particles is set according to a given snowfall rate $r_s$.

\begin{figure}[t]
    \begin{center}
    \includegraphics[width=0.49\textwidth]{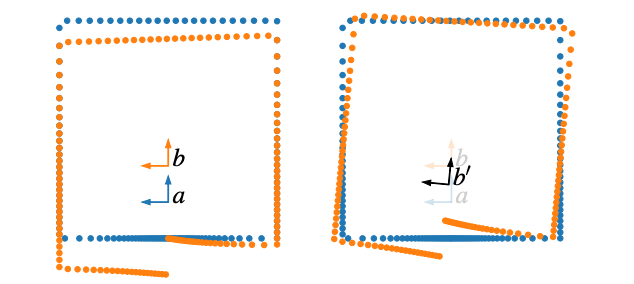}
    \end{center}
    \vspace{-0.4cm}
    \caption{Examples of the effect brought by \textbf{\textit{motion blur}} on the registration of a square room. The blue trajectory denotes the globally consistent map of the environment; the yellow points were acquired while the LiDAR sensor was moving from pose $a$ to pose $b$, resulting in a heavily skewed point cloud. \textbf{\textit{Image credit:}} Deschênes \textit{et al.} \cite{Motion_Blur}.}
    \label{add-fig:motion_blur}
    \vspace{0.1cm}
\end{figure}

\subsection{Motion Blur}
As one of the common in-vehicle sensors, LiDAR is often mounted on the rooftop or side of the vehicle and inevitably suffers from the blur caused by vehicle movement, especially on bumpy surfaces or during U-turning. A typical example of the effect brought by \textit{motion blur} is shown in \cref{add-fig:motion_blur}.

In this work, to simulate blur-corrupted data  $\mathcal{C}_{\text{motion}}(\mathbf{p})$, we add a jittering noise to each coordinate $(p^x, p^y, p^z)$ with a translation value sampled from the Gaussian distribution with standard deviation $\sigma_{t}$. The $\mathcal{C}_{\text{motion}}(\mathbf{p})$ is shown as:
\begin{equation}    
\mathcal{C}_{\text{motion}}
(\mathbf{p}) = (p^x+o_1, p^y+o_2, p^z+o_3, p^i)~,
\end{equation}
where $o_1,o_2, o_3$ are the random offsets sampled from Gaussian distribution $N \in \{0, {\sigma_{t}}^2 \}$ and $\{o_1,o_2, o_3\} \in\mathbb{R}^{1\times1 }$.

\subsection{Beam Missing}
As shown in \cref{add-fig:beam_missing}, the dust and insect tend to form agglomerates in front of the LiDAR surface and will not likely disappear without human intervention, such as drying and cleaning~\cite{Dust}. This type of occlusion causes zero readings on masked areas and results in the loss of certain light impulses.

In this work, to mimic such a behavior, we randomly sample a total number of $m$ beams and drop points on these beams from the original point cloud to generate $\mathcal{C}_{\text{beam}}(\mathbf{p})$:
\begin{equation}    
\mathcal{C}_{\text{beam}}(\mathbf{p}) =
\begin{cases}
(p^x, p^y, p^z, p^i), & \text{if~~~~$\mathbf{p} \notin m $ },\\
\text{None}, & \text{else}~.
\end{cases}
\end{equation}

\subsection{Crosstalk}
\begin{figure*}[t]
    \begin{center}
    \includegraphics[width=1.0\textwidth]{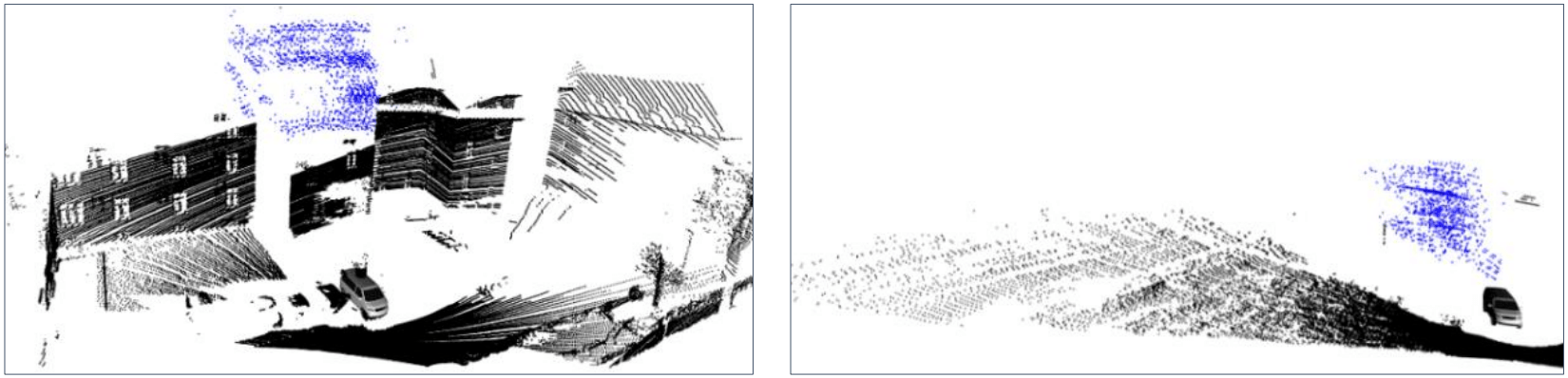}
    \end{center}
    \vspace{-0.4cm}
    \caption{Examples of the data corruptions introduced by \textbf{\textit{crosstalk}}. The point clouds are acquired by a Velodyne HDL-64 with interference from another sensor of the same type in close vicinity. The crosstalk points are shown in \textcolor{blue}{blue}. \textbf{\textit{Image credit:}} Diehm \textit{et al.} \cite{Crosstalk2}.}
    \label{add-fig:crosstalk2}
    \vspace{0.2cm}
\end{figure*}

Considering that the road is often shared by multiple vehicles (see \cref{add-fig:crosstalk1}), the time-of-flight of light impulses from one sensor might interfere with impulses from other sensors within  a similar frequency range~\cite{Crosstalk}. Such a crosstalk phenomenon often creates noisy points within the mid-range areas in between two (or multiple) sensors. \cref{add-fig:crosstalk2} shows two real-world examples of crosstalk-corrupted point clouds.

In this work, to simulate $\mathcal{C}_{\text{cross}}(\mathbf{p})$, we randomly sample a subset of $k_t$ percent points from the original point cloud and add large jittering noise with a translation value sampled from the Gaussian distribution with standard deviation $\sigma_{c}$.
\begin{equation}    
\mathcal{C}_{\text{cross}}(\mathbf{p}) =
\begin{cases}
(p^x, p^y, p^z, p^i), & \text{if~~$\mathbf{p} \notin$ \text{set of \{$k_t$\} }},\\
(p^x, p^y, p^z, p^i) + \xi_c, & \text{else}~,
\end{cases}
\end{equation}
where $\xi_c$ is the random offset sampled from Gaussian distribution $N \in \{0, {\sigma_{c}}^2 \}$ and $\xi_c \in\mathbb{R}^{1\times4 }$.

\subsection{Incomplete Echo}
\begin{figure}[t]
    \begin{center}
    \includegraphics[width=0.49\textwidth]{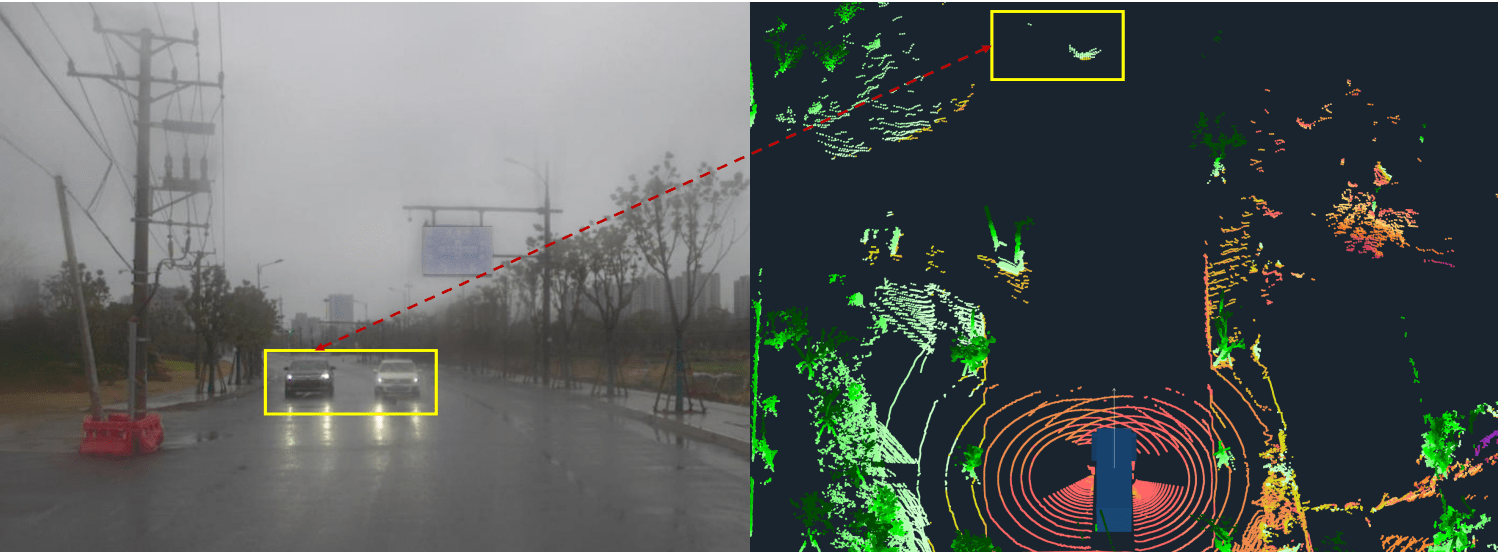}
    \end{center}
    \vspace{-0.4cm}
    \caption{Examples of the data corruptions introduced by \textbf{\textit{imcomplete echo}}. The black car on the left has nearly zero pulse return, due to the destroyed echo cycle. \textbf{\textit{Image credit:}} Yu \textit{et al.} \cite{lidar_camera_fusion}.}
    \label{add-fig:imcomplete_echo}
    \vspace{0.1cm}
\end{figure}

The near-infrared spectrum of the laser pulse emitted from the LiDAR sensor is vulnerable to vehicles or other instances with dark colors~\cite{lidar_camera_fusion}. The LiDAR readings are thus incomplete in such scan echoes, resulting in significant point miss detection (see \cref{add-fig:imcomplete_echo} for a real-world example).

In this work, we simulate this corruption which denotes $\mathcal{C}_{\text{echo}}(\mathbf{p})$ by randomly querying $k_e$ percent points for \textit{vehicle}, \textit{bicycle}, and \textit{motorcycle} classes, via either semantic masks or 3D bounding boxes. Next, we drop the queried points from the original point cloud, along with their point-level semantic labels. Note that we do not alter the ground-truth bounding boxes since they should remain at their original positions in the real world. This can be formed as:
\begin{equation}    
\mathcal{C}_{\text{echo}}(\mathbf{p}) =
\begin{cases}
(p^x, p^y, p^z, p^i), & \text{if~~$\mathbf{p} \notin$ \text{set of \{$k_e$\} }},\\
\text{None}, & \text{else}~.
\end{cases}
\end{equation}

\subsection{Cross-Sensor}
\begin{figure}[t]
    \begin{center}
    \includegraphics[width=0.49\textwidth]{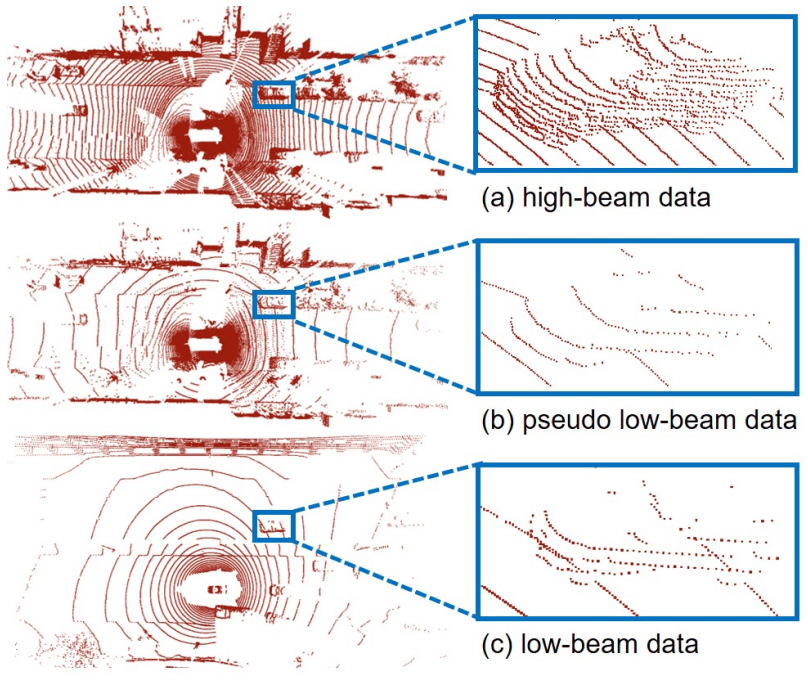}
    \end{center}
    \vspace{-0.5cm}
    \caption{Examples of the data distribution discrepancy brought by \textbf{\textit{cross-sensor}} effect. (a) A typical point cloud acquired by a 64-beam LiDAR sensor. (b) A simulated point cloud from 64 beams to 32 beams. (c) A typical point cloud acquired by a 32-beam LiDAR sensor. \textbf{\textit{Image credit:}} Wei \textit{et al.} \cite{wei2022distillation}.}
    \label{add-fig:cross_sensor}
    \vspace{0.1cm}
\end{figure}

A typical \textit{cross-sensor} example is shown in \cref{add-fig:cross_sensor}. Due to the large variety of LiDAR sensor configurations (\textit{e.g.}, beam number, FOV, and sampling frequency), it is important to design robust 3D perception  models that are capable of maintaining satisfactory performance under cross-device cases~\cite{std}. While previous works directly form such settings with two different datasets, the domain idiosyncrasy in between (\textit{e.g.}, different label mappings and data collection protocols) further hinders the direct robustness comparison.

In our benchmark, we follow \cite{wei2022distillation} and generate cross-sensor data $\mathcal{C}_{\text{sensor}}(\mathbf{p})$ by first dropping points of certain beams from the point cloud and then sub-sample $k_c$ percent points from each beam:
\begin{equation}    
\mathcal{C}_{\text{sensor}}(\mathbf{p}) =
\begin{cases}
\text{None},  & \text{if~~$ \text{$\mathbf{p}\in$ set of \{$k_c$\}} $
}, \\
(p^x, p^y, p^z, p^i), & \text{else}~.
\end{cases}
\end{equation}

\section{Additional Implementation Detail}
\label{add-sec:implementation}

In this section, we provide additional implementation details to enable the reproduction of the corruption generations in the Robo3D benchmark. Note that our physically principled corruption creation procedures can also be used on other LiDAR-based point cloud datasets with minimal modifications.

\subsection{Fog Simulation}
Following~\cite{Fog}, we uniformly sample the attenuation coefficient $\alpha$ from $[0, 0.005, 0.01, 0.02, 0.03, 0.06]$. For the \textit{SemanticKITTI-C}, \textit{KITTI-C}, \textit{nuScenes-C}, and \textit{WOD-C} datasets, we set the back-scattering coefficient $\beta$ to $\{0.008, 0.05, 0.2\}$ to split severity levels into light, moderate, and heavy levels. The semantic classes of \textit{fog} are $21$, $41$, and $23$ for \textit{SemanticKITTI-C}, \textit{nuScenes-C}, and \textit{WOD-C}, respectively. And $\mathbf{p}  \text{~belongs to fog class}$ will be mapped to class 0 or 255 (\textit{i.e.}, the \textit{ignored} label).

\subsection{Wet Ground Simulation}
We follow~\cite{Snow_Wet} and set the parameter of water height $d_w$ to $\{0.2~mm, 1.0~mm, 1.2~mm\}$ for different severity levels of \textit{wet ground}. Note that the ground plane estimation method is different across four benchmarks.  We estimate the ground plane via RANDSAC~\cite{fischler1981random} for the \textit{KITTI-C} since it only provides detection labels. For \textit{SemanticKITTI-C}, we use semantic classes of \textit{road}, \textit{parking}, \textit{sidewalk}, and \textit{other ground} to build the ground plane. The \textit{driveable surface}, \textit{other flat}, and \textit{sidewalk} classes are used to construct the ground plane in \textit{nuScenes-C}. For \textit{WOD-C}, the ground plane is estimated by \textit{curb}, \textit{road}, \textit{other ground}, \textit{walkable}, and \textit{sidewalk} classes.

\subsection{Snow Simulation}
We use the method proposed in~\cite{Snow_Wet} to construct \textit{snow} corruptions. The value of snowfall rate parameter $r_s$ is set to $\{0.5, 1.0, 2.5\}$ to simulate light, moderate, and heavy snowfall for the \textit{SemanticKITTI-C}, \textit{KITTI-C}, \textit{nuScenes-C}, and \textit{WOD-C} datasets, and the ground plane estimation is the same as the \textit{wet ground} simulation. The semantic class of snow is $22$, $42$, and $24$ for the \textit{SemanticKITTI-C}, \textit{nuScenes-C}, and \textit{WOD-C} datasets, respectively. And $\mathbf{p}  \text{~belongs to snow class}$ will also be mapped to class 0 or 255 (\textit{i.e.}, the \textit{ignored} label).

\subsection{Motion Blur Simulation}
We add jittering noise from Gaussian distribution with standard deviation $\sigma_t$ to simulate motion blur. The $\sigma_t$ is set to $\{0.20, 0.25, 0.30\}$, $\{0.04, 0.08, 0.10\}$, $\{0.20, 0.30, 0.40\}$ and $\{0.06, 0.10, 0.13\}$ for the \textit{SemanticKITTI-C}, \textit{KITTI-C}, \textit{nuScenes-C}, and \textit{WOD-C} datasets, respectively.

\subsection{Beam Missing Simulation}
The value of parameter $m$ (number of beams to be dropped) is set to $\{48, 32, 16 \}$ for the benchmark of \textit{SemanticKITTI-C}, \textit{KITTI-C} and \textit{WOD-C}, respectively, while set as $\{24, 16, 8 \}$ for the \textit{nuScenes-C} dataset.

\subsection{Crosstalk Simulation}
We set the parameter of $k_t$ to $\{0.006, 0.008, 0.01\}$ for the \textit{SemanticKITTI-C}, \textit{KITTI-C}, and \textit{WOD-C} datasets, respectively, and $\{0.03, 0.07, 0.12\}$ for \textit{nuScenes-C} dataset. The semantic class of crosstalk is assigned to $23$, $43$, and $25$ for \textit{SemanticKITTI-C}, \textit{nuScenes-C}, and \textit{WOD-C} datasets, respectively. Meanwhile, the $\mathbf{p}  \text{~belongs to crosstalk class}$ will also be mapped to class 0 or 255 (\textit{i.e.}, the \textit{ignored} label).

\subsection{Incomplete Echo Simulation}
For \textit{SemanticKITTI-C}, the point labels of classes \textit{car}, \textit{bicycle}, \textit{motorcycle}, \textit{truck}, \textit{other-vehicle} are used as the semantic mask. For \textit{nuScenes-C}, we include \textit{bicycle}, \textit{bus}, \textit{car}, \textit{construction vehicle}, \textit{motorcycle}, \textit{truck} and \textit{trailer} class label to build semantic mask.  For \textit{WOD-C}, we adopt the point labels of classes \textit{car}, \textit{truck}, \textit{bus}, \textit{other-vehicle}, \textit{bicycle}, \textit{motorcycle} as the semantic mask. For \textit{KITTI-C}, we use 3D bounding box labels to create the semantic mask. The value of parameter $k_e$ is set to $\{0.75,0.85, 0.95\}$ for the four corruption sets during the \textit{incomplete echo} simulation.

\subsection{Cross-Sensor Simulation}
The value of parameter $m$ is set to $\{48, 32, 16 \}$ for the \textit{SemanticKITTI-C}, \textit{KITTI-C}, and \textit{WOD-C} datasets, respectively, and $\{24, 16, 12 \}$ for the \textit{nuScenes-C} dataset. Based on~\cite{wei2022distillation}, we then sub-sample 50$\%$ points from the remaining point clouds with an equal interval.

\begin{figure*}[t]
    \begin{center}
    \includegraphics[width=1.0\textwidth]{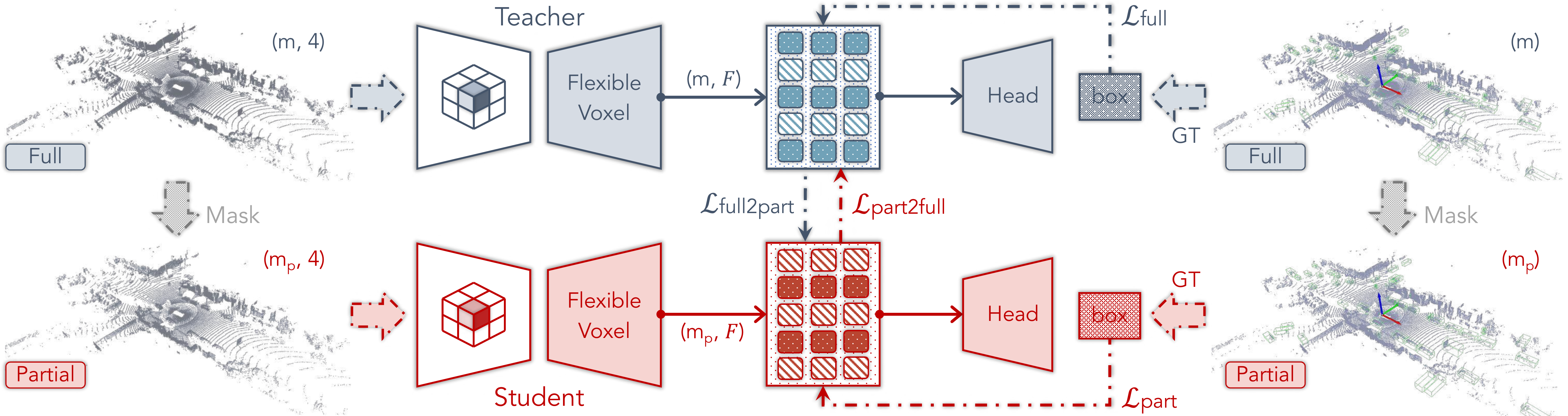}
    \end{center}
    \vspace{-0.4cm}
    \caption{The \textbf{3D object detection realization} of the proposed density-insensitive training framework. The ``full" and ``partial" point clouds are fed into the teacher branch and student branch, respectively, for feature learning, while the latter is generated by randomly masking the original point cloud. To encourage cross-density consistency, we calculate the \textit{completion} and \textit{confirmation} losses which measure the distances of the teacher's prediction (BEV feature map) and the student's prediction (BEV feature map) between the other branch's outputs.}
    \label{fig:framework_det3d}
    \vspace{0.2cm}
\end{figure*}

\begin{figure*}[t]
    \begin{center}
    \includegraphics[width=1.0\textwidth]{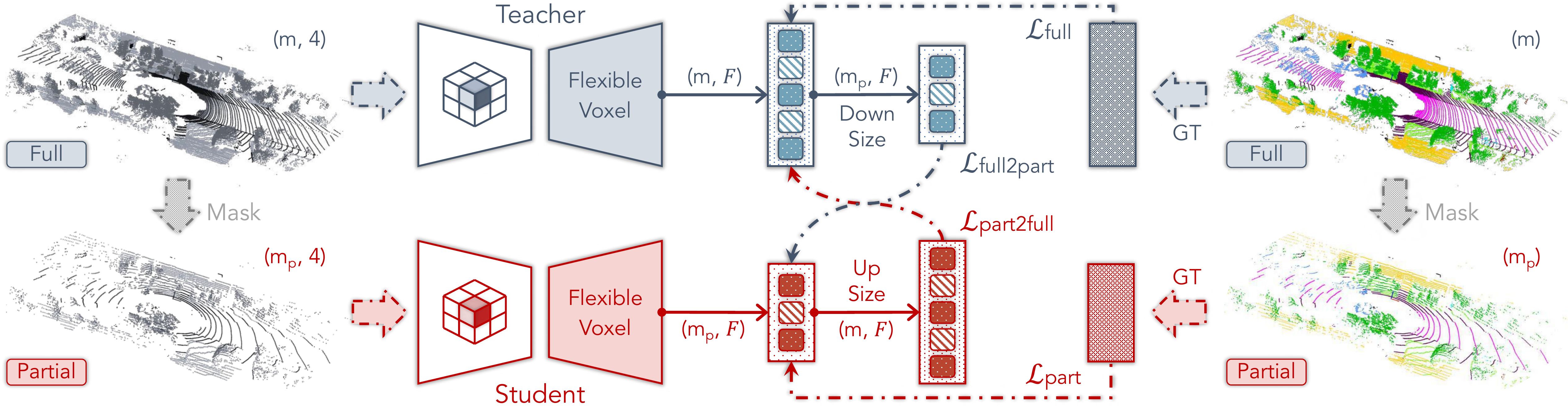}
    \end{center}
    \vspace{-0.4cm}
    \caption{The \textbf{3D semantic segmentation realization} of the proposed density-insensitive training framework. The ``full" and ``partial" point clouds are fed into the teacher branch and student branch, respectively, for feature learning, while the latter is generated by randomly masking the original point cloud. To encourage cross-density consistency, we calculate the \textit{completion} and \textit{confirmation} losses which measure the distances of the sub-sampled teacher's prediction and interpolated student's prediction between the other branch's outputs.}
    \label{fig:framework_seg3d}
    \vspace{0.2cm}
\end{figure*}

\section{Additional Experimental Result}
\label{add-sec:experiment}

In this section, we provide the complete experimental results for each of the 3D detectors and segmentors benchmarked in Robo3D. 

\subsection{SemanticKITTI-C}
The complete results in terms of corruption error (CE), resilience rate (RR), and task-specific accuracy (IoU) on the \textit{SemanticKITTI-C} dataset are shown in \cref{tab:semkitti_c_ce_supp}, \cref{tab:semkitti_c_rr}, and \cref{tab:semkitti_c_iou}, respectively.

\subsection{KITTI-C}
The complete results in terms of corruption error (CE), resilience rate (RR), and task-specific accuracy (AP) on the \textit{KITTI-C} dataset are shown in \cref{tab:kitti_c_ce_supp}, \cref{tab:kitti_c_rr}, and \cref{tab:kitti_c_ap}, respectively.

\subsection{nuScenes-C (Seg3D)}
The complete results in terms of corruption error (CE), resilience rate (RR), and task-specific accuracy (IoU) on the \textit{nuScenes-C (Seg3D)} dataset are shown in \cref{tab:nuscenes_c_seg_ce_supp}, \cref{tab:nuscenes_c_seg_rr}, and \cref{tab:nuscenes_c_seg_iou}, respectively. In addition to the benchmark results, we also show the voxel size analysis results of \textit{nuScenes-C (Seg3D)} in \cref{fig:vs_nusc_wod}~(a).

\subsection{nuScenes-C (Det3D)}
The complete results in terms of corruption error (CE), resilience rate (RR), and task-specific accuracy (NDS) on the \textit{nuScenes-C (Det3D)} dataset are shown in \cref{tab:nuscenes_c_det_ce_supp}, \cref{tab:nuscenes_c_det_rr}, and \cref{tab:nuscenes_c_det_nds}, respectively.

\subsection{WOD-C (Seg3D)}
\begin{figure*}
    \centering
    \subfigure[Voxel Size on \textit{nuScenes-C} (Seg3D)]{
        \includegraphics[width=0.65\linewidth]{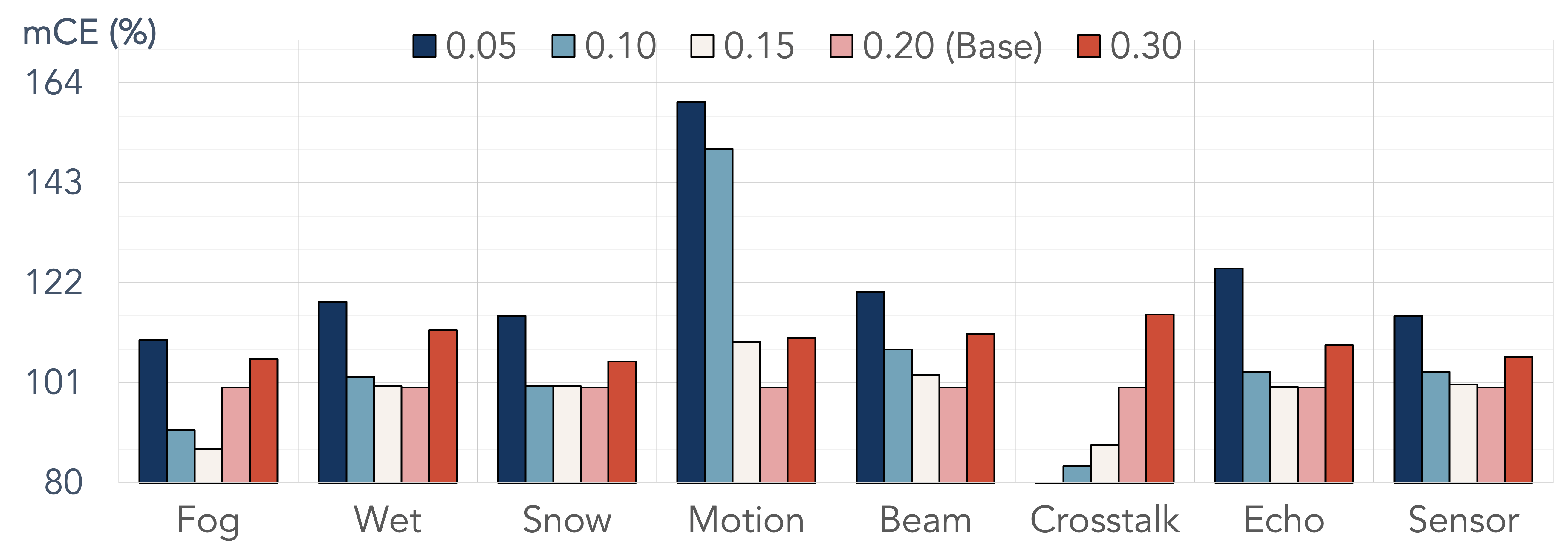}
    }
    \subfigure[Voxel Size on \textit{WOD-C} (Seg3D)]{
        \includegraphics[width=0.647\linewidth]{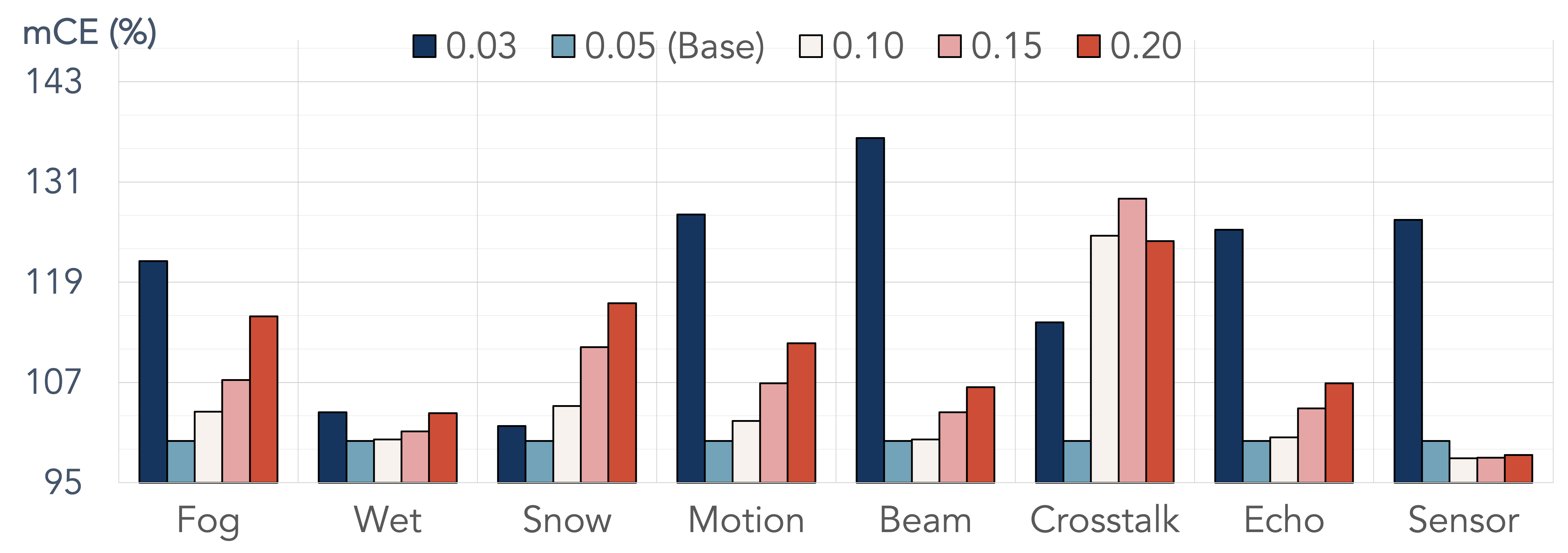}
    }
    \vspace{-0.1cm}
    \caption{Corruption sensitivity analysis of the \textit{voxel size} for the baseline LiDAR semantic segmentation model \cite{2019Minkowski}. The experiments are conducted on: a) the \textit{nuScenes-C (Seg3D)} dataset; and b) the \textit{WOD-C (Seg3D)} dataset. Different corruptions exhibit variances under certain configurations.}
    \label{fig:vs_nusc_wod}
\end{figure*}

The complete results in terms of corruption error (CE), resilience rate (RR), and task-specific accuracy (IoU) on the \textit{WOD-C (Seg3D)} dataset are shown in \cref{tab:wod_c_seg3d_ce_supp}, \cref{tab:wod_c_seg3d_rr}, and \cref{tab:wod_c_seg3d_iou}, respectively. In addition to the benchmark results, we also show the voxel size analysis results of \textit{WOD-C (Seg3D)} in \cref{fig:vs_nusc_wod}~(b).

\subsection{WOD-C (Det3D)}
The complete results in terms of corruption error (CE), resilience rate (RR), and task-specific accuracy (APH) on the \textit{WOD-C (Det3D)} dataset are shown in \cref{tab:wod_c_det3d_ce_supp}, \cref{tab:wod_c_det3d_rr}, and \cref{tab:wod_c_det3d_aph}, respectively.

\subsection{Density-Insensitive Training}
As stated in the main body, the corruptions in the real-world environment often cause severe occlusion, attenuation, and reflection of
LiDAR impulses, resulting in the unavoidable loss of points in certain regions around the ego-vehicle. To better handle such scenarios, we design a density-insensitive training framework, with realizations on both detection (see \cref{fig:framework_det3d}) and segmentation (see \cref{fig:framework_seg3d}).

Since detection and segmentation have different optimization objectives, we design different loss computation strategies within these two frameworks. Specifically, the \textit{completion} and \textit{confirmation} losses for the detection framework are calculated at the BEV feature maps; while for the segmentation framework, these two losses are computed at the logits level. Our experimental results in \cref{tab:framework} verify the effectiveness of this approach on both tasks. Although we use a random masking strategy to avoid information leaks, we observe overt improvements in a wide range of corruption types that contain point loss scenarios, such as \textit{beam missing}, \textit{incomplete echo}, and \textit{cross-sensor}. We believe more sophisticated designs based on our framework could further boost the corruption robustness of 3D perception models.

\section{Qualitative Experiment}
\label{add-sec:qualitative}

In this section, we provide extensive qualitative examples for illustrating the proposed corruption types and for comparing representative models benchmarked in Robo3D. 

\subsection{Corruption Types}
\begin{figure*}[t]
    \begin{center}
    \includegraphics[width=0.837\linewidth]{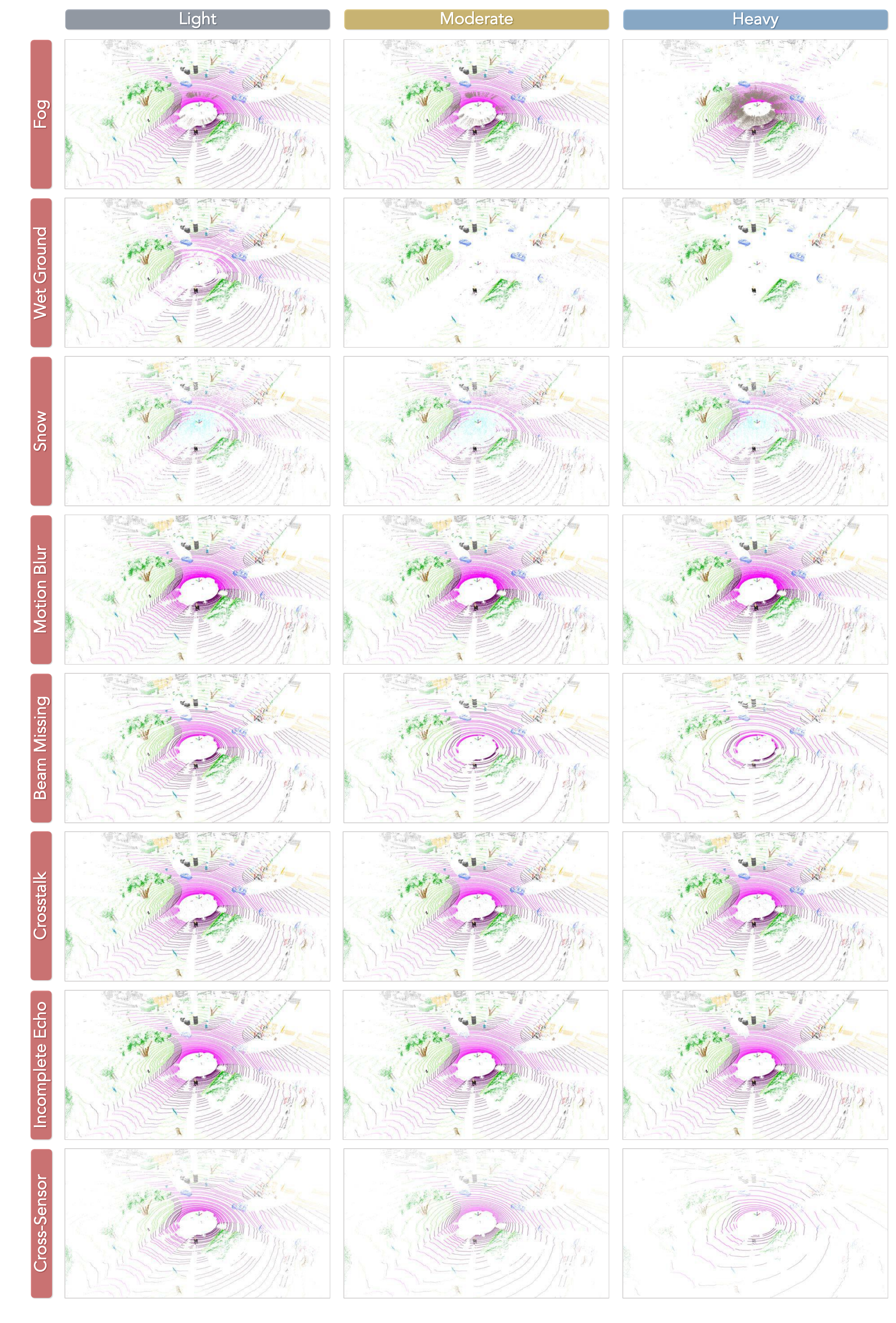}
    \end{center}
    \vspace{-0.4cm}
    \caption{Visual examples of each corruption type under three severity levels in our \textit{SemanticKITTI-C} dataset.}
    \label{figure:example_semkitti_c}
\end{figure*}

\begin{figure*}[t]
    \begin{center}
    \includegraphics[width=0.835\linewidth]{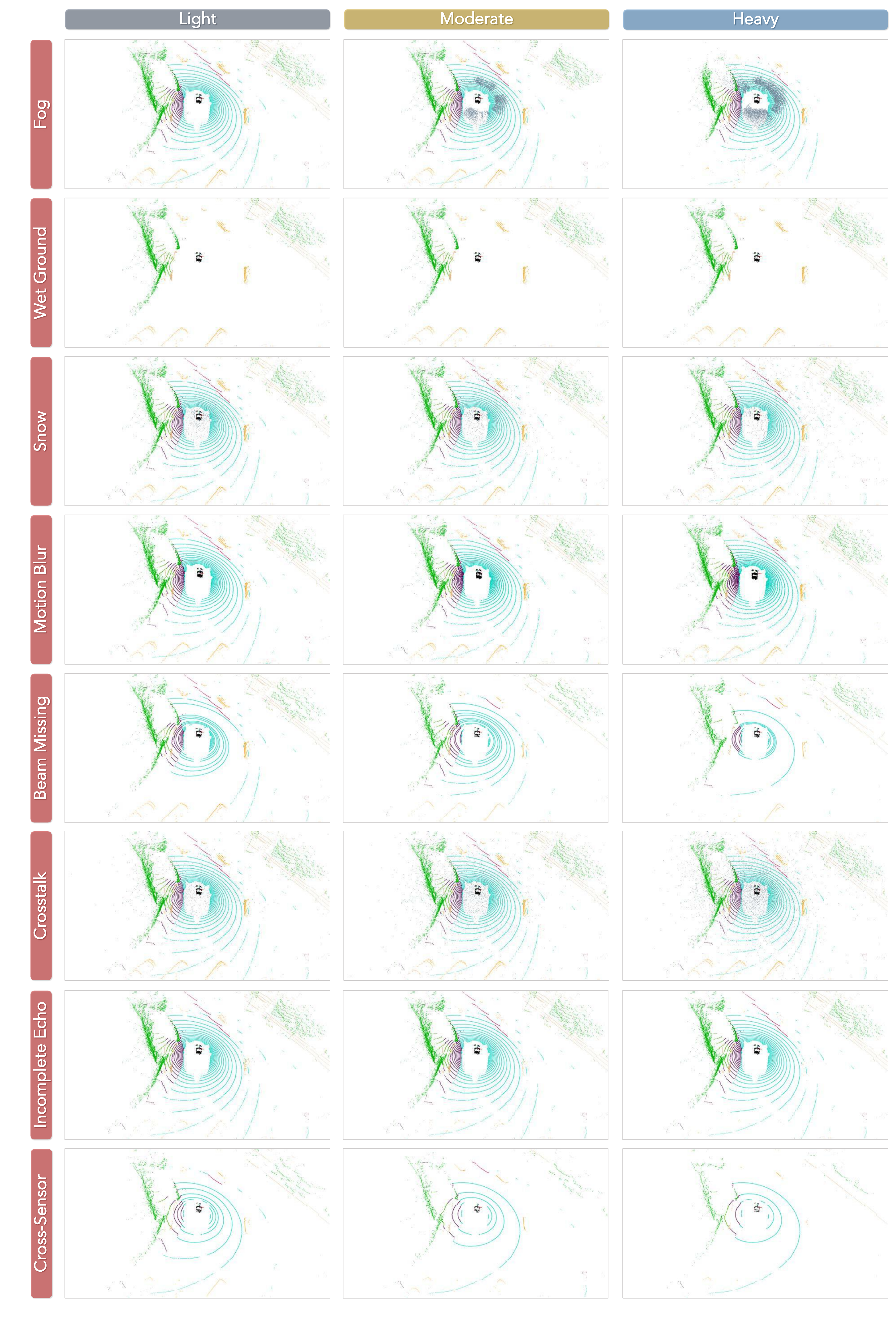}
    \end{center}
    \vspace{-0.4cm}
    \caption{Visual examples of each corruption type under three severity levels in our \textit{nuScenes-C} dataset.}
    \label{figure:example_nuscenes_c}
\end{figure*}

We show visualizations of the eight corruption types under three severity levels (light, moderate, and heavy) in \cref{figure:example_semkitti_c} and \cref{figure:example_nuscenes_c}.

\subsection{Visual Comparisons}
\begin{figure*}[t]
    \begin{center}
    \includegraphics[width=1.0\linewidth]{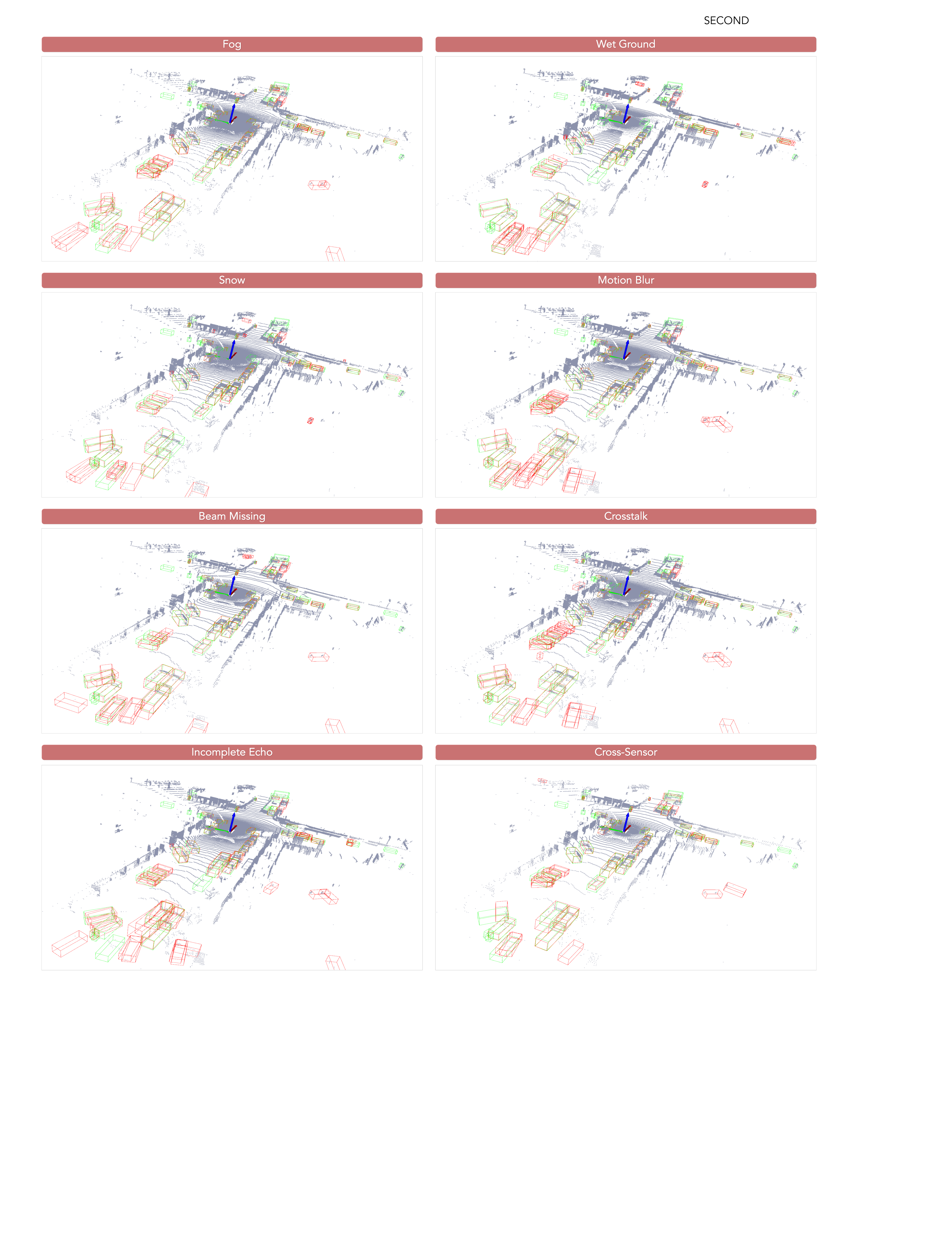}
    \end{center}
    \vspace{-0.4cm}
    \caption{\textbf{Qualitative results} of SECOND \cite{second} under each of the eight corruptions in \textit{WOD-C (Det3D)}. The green boxes represent the groundtruth, while the red boxes are the predictions. Best viewed in colors.}
    \label{figure:qualitative_det3d_supp_01}
\end{figure*}

\begin{figure*}[t]
    \begin{center}
    \includegraphics[width=1.0\linewidth]{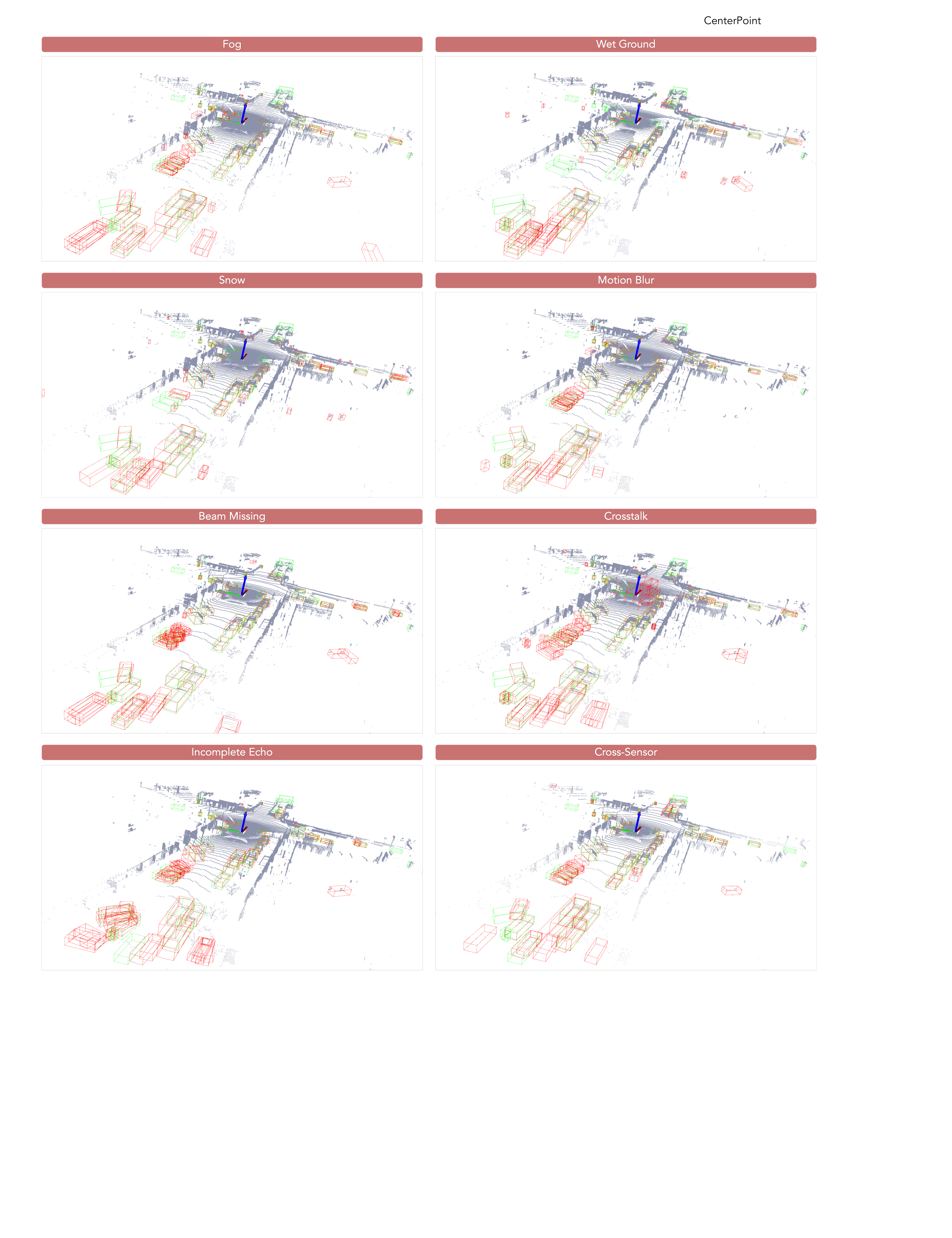}
    \end{center}
    \vspace{-0.4cm}
    \caption{\textbf{Qualitative results} of CenterPoint \cite{centerpoint} under each of the eight corruptions in \textit{WOD-C (Det3D)}. The green boxes represent the groundtruth, while the red boxes are the predictions. Best viewed in colors.}
    \label{figure:qualitative_det3d_supp_02}
\end{figure*}

For 3D object detection, we attach the qualitative results of SECOND \cite{second} and CenterPoint \cite{centerpoint} under each of the eight corruption types in the \textit{WOD-C (Det3D)} dataset. The results are shown in \cref{figure:qualitative_det3d_supp_01} and \cref{figure:qualitative_det3d_supp_02}.

\begin{figure*}[t]
    \begin{center}
    \includegraphics[width=1.0\linewidth]{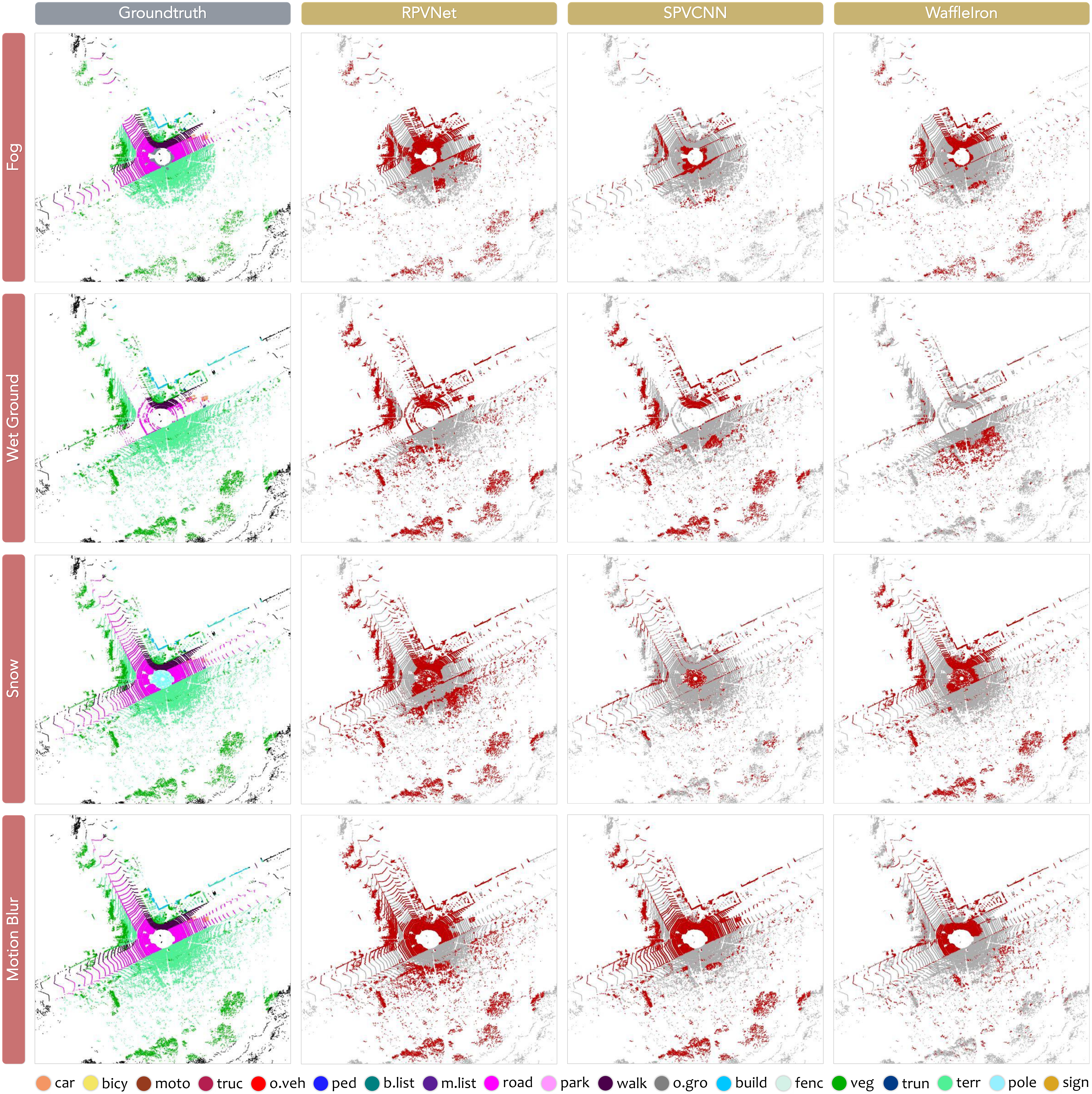}
    \end{center}
    \vspace{-0.3cm}
    \caption{\textbf{Qualitative comparisons (error maps)} of three LiDAR segmentation models (RPVNet \cite{xu2021rpvnet}, SPVCNN \cite{tang2020searching}, WaffleIron \cite{puy23waffleiron}) under the \textbf{\textit{fog}}, \textbf{\textit{wet ground}}, \textbf{\textit{snow}}, and \textbf{\textit{motion blur}} corruptions, in \textit{SemanticKITTI-C}. To highlight the differences, the \textbf{\textcolor{correct}{correct}} / \textbf{\textcolor{incorrect}{incorrect}} predictions are painted in \textbf{\textcolor{correct}{gray}} / \textbf{\textcolor{incorrect}{red}}, respectively. Each scene is visualized from the LiDAR bird's eye view and covers a $50$m by $50$m region, centered around the ego-vehicle. Best viewed in colors.}
    \label{figure:qualitative_supp_03}
\end{figure*}

\begin{figure*}[t]
    \begin{center}
    \includegraphics[width=1.0\linewidth]{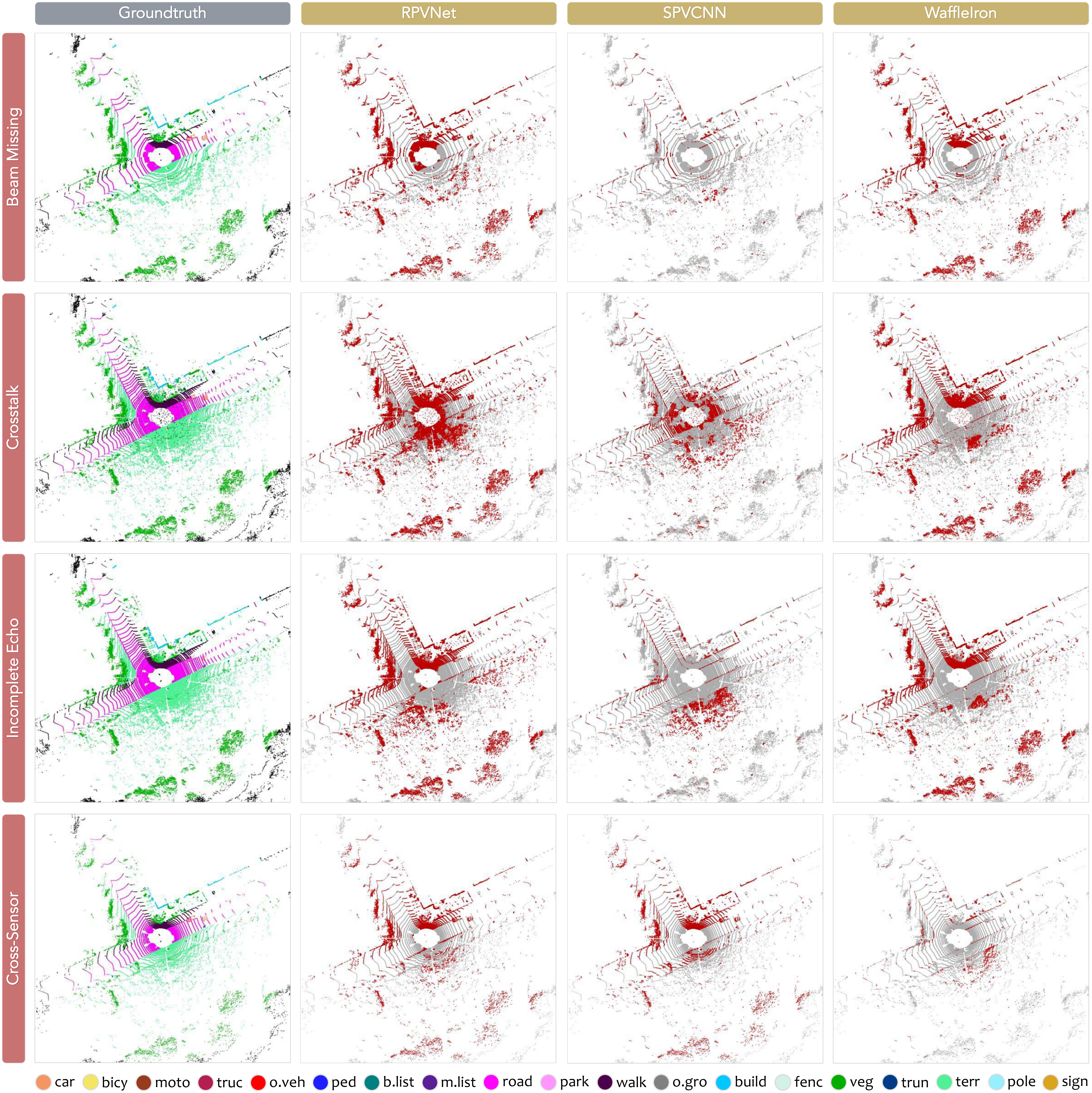}
    \end{center}
    \vspace{-0.3cm}
    \caption{\textbf{Qualitative comparisons (error maps)} of three LiDAR segmentation models (RPVNet \cite{xu2021rpvnet}, SPVCNN \cite{tang2020searching}, WaffleIron \cite{puy23waffleiron}) under the \textbf{\textit{beam missing}}, \textbf{\textit{crosstalk}}, \textbf{\textit{incomplete echo}}, and \textbf{\textit{cross-sensor}} corruptions, in \textit{SemanticKITTI-C}. To highlight the differences, the \textbf{\textcolor{correct}{correct}} / \textbf{\textcolor{incorrect}{incorrect}} predictions are painted in \textbf{\textcolor{correct}{gray}} / \textbf{\textcolor{incorrect}{red}}, respectively. Each scene is visualized from the LiDAR bird's eye view and covers a $50$m by $50$m region, centered around the ego-vehicle. Best viewed in colors.}
    \label{figure:qualitative_supp_04}
\end{figure*}

\begin{figure*}[t]
    \begin{center}
    \includegraphics[width=1.0\linewidth]{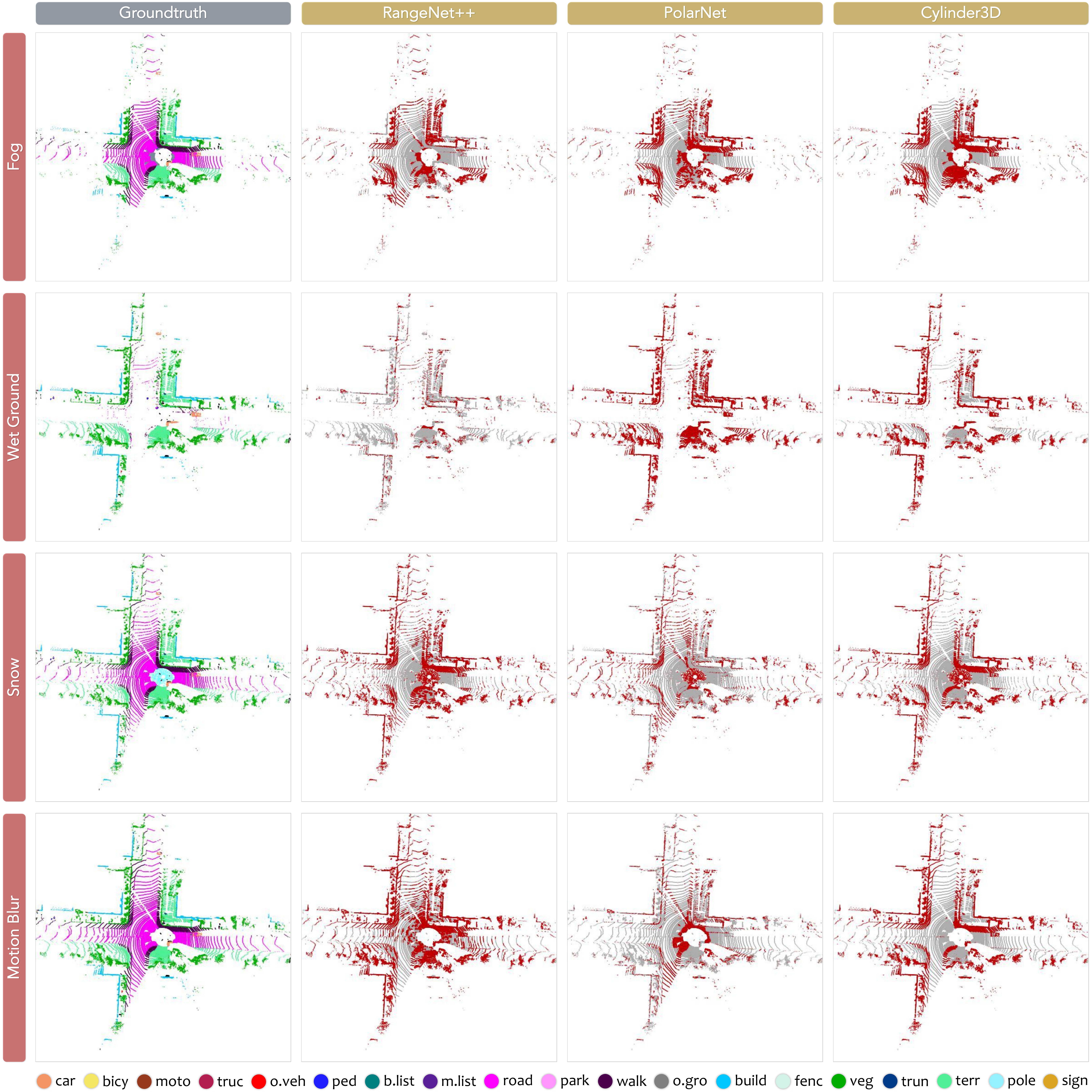}
    \end{center}
    \vspace{-0.3cm}
    \caption{\textbf{Qualitative comparisons (error maps)} of three LiDAR segmentation models (RangeNet++ \cite{milioto2019rangenet++}, PolarNet \cite{zhang2020polarnet}, Cylinder3D \cite{zhu2021cylindrical}) under the \textbf{\textit{fog}}, \textbf{\textit{wet ground}}, \textbf{\textit{snow}}, and \textbf{\textit{motion blur}} corruptions, in \textit{SemanticKITTI-C}. To highlight the differences, the \textbf{\textcolor{correct}{correct}} / \textbf{\textcolor{incorrect}{incorrect}} predictions are painted in \textbf{\textcolor{correct}{gray}} / \textbf{\textcolor{incorrect}{red}}, respectively. Each scene is visualized from the LiDAR bird's eye view and covers a $50$m by $50$m region, centered around the ego-vehicle. Best viewed in colors.}
    \label{figure:qualitative_supp_01}
\end{figure*}

\begin{figure*}[t]
    \begin{center}
    \includegraphics[width=1.0\linewidth]{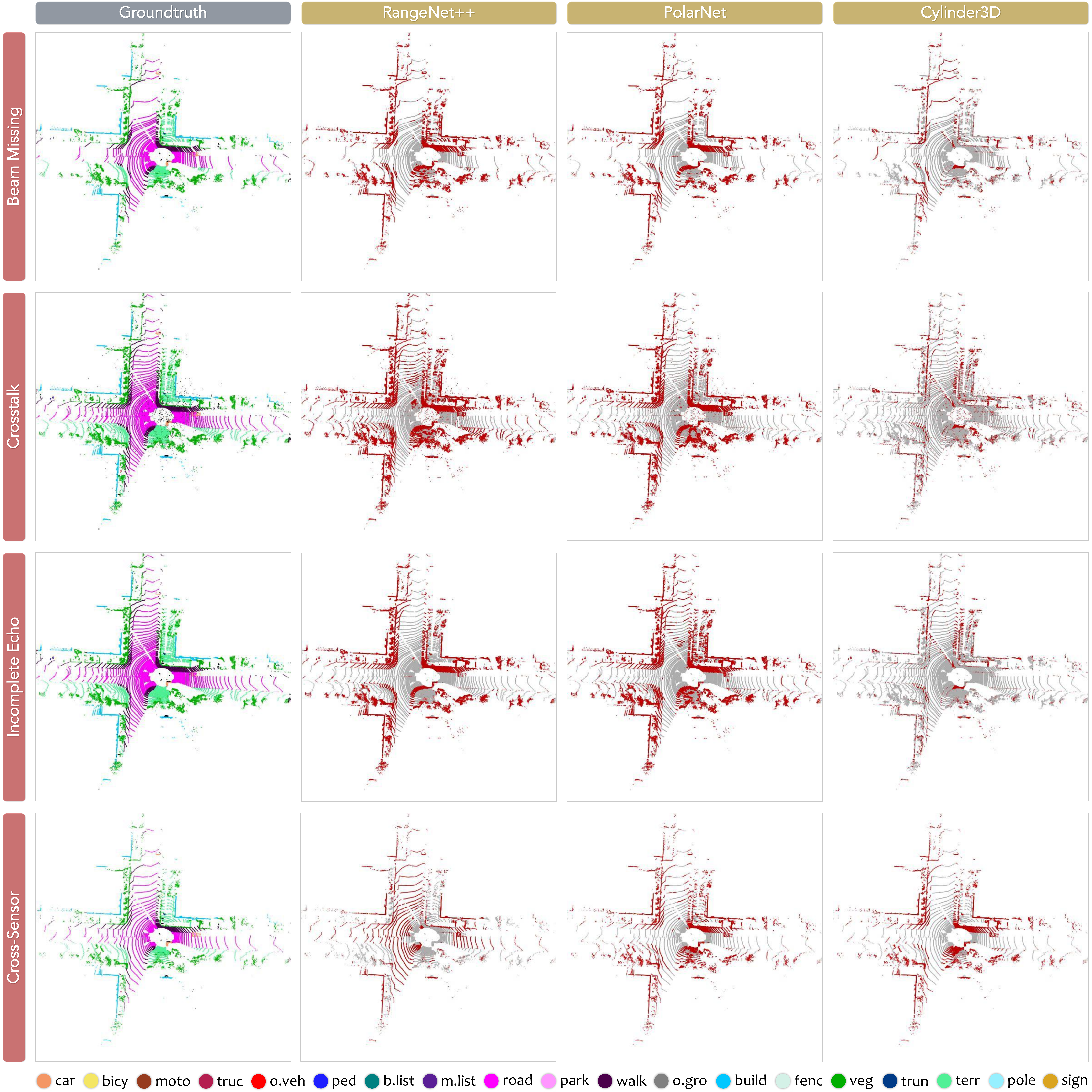}
    \end{center}
    \vspace{-0.3cm}
    \caption{\textbf{Qualitative comparisons (error maps)} of three LiDAR segmentation models (RangeNet++ \cite{milioto2019rangenet++}, PolarNet \cite{zhang2020polarnet}, Cylinder3D \cite{zhu2021cylindrical}) under the \textbf{\textit{beam missing}}, \textbf{\textit{crosstalk}}, \textbf{\textit{incomplete echo}}, and \textbf{\textit{cross-sensor}} corruptions, in \textit{SemanticKITTI-C}. To highlight the differences, the \textbf{\textcolor{correct}{correct}} / \textbf{\textcolor{incorrect}{incorrect}} predictions are painted in \textbf{\textcolor{correct}{gray}} / \textbf{\textcolor{incorrect}{red}}, respectively. Each scene is visualized from the LiDAR bird's eye view and covers a $50$m by $50$m region, centered around the ego-vehicle. Best viewed in colors.}
    \label{figure:qualitative_supp_02}
\end{figure*}

For 3D semantic segmentation, we attach qualitative results of six segmentors, \textit{i.e.}, RangeNet++ \cite{milioto2019rangenet++}, PolarNet \cite{zhang2020polarnet}, Cylinder3D \cite{zhu2021cylindrical}, RPVNet \cite{xu2021rpvnet}, SPVCNN \cite{tang2020searching}, and WaffleIron \cite{puy23waffleiron}, under each of the eight corruption types in the \textit{SemanticKITTI-C} dataset. The results are shown in \cref{figure:qualitative_supp_03}, \cref{figure:qualitative_supp_04}, \cref{figure:qualitative_supp_01}, and \cref{figure:qualitative_supp_02}.

\subsection{Video Demos}
In addition to the figures shown in this file, we have included four video demos on our project page. Each of these demos consists of hundred of frames that provide a more comprehensive evaluation of our proposed benchmark.

\section{Public Resources Used}
\label{sec:public-resources-used}

In this section, we acknowledge the use of the following public resources, during the course of this work:

\begin{itemize}
    \item SemanticKITTI\footnote{\url{http://semantic-kitti.org}.} \dotfill CC BY-NC-SA 4.0
    \item SemanticKITTI-API\footnote{\url{https://github.com/PRBonn/semantic-kitti-api}.} \dotfill MIT License
    \item nuScenes\footnote{\url{https://www.nuscenes.org/nuscenes}.} \dotfill CC BY-NC-SA 4.0
    \item nuScenes-devkit\footnote{\url{https://github.com/nutonomy/nuscenes-devkit}.} \dotfill Apache License 2.0
    \item Waymo Open Dataset\footnote{\url{https://waymo.com/open}.} \dotfill Waymo Dataset License
    \item RangeNet++\footnote{\url{https://github.com/PRBonn/lidar-bonnetal}.} \dotfill MIT License
    \item SalsaNext\footnote{\url{https://github.com/TiagoCortinhal/SalsaNext}.} \dotfill MIT License
    \item FIDNet\footnote{\url{https://github.com/placeforyiming/IROS21-FIDNet-SemanticKITTI}.} \dotfill Unknown
    \item CENet\footnote{\url{https://github.com/huixiancheng/CENet}.} \dotfill MIT License
    \item KPConv-PyTorch\footnote{\url{https://github.com/HuguesTHOMAS/KPConv-PyTorch}.} \dotfill MIT License
    \item PIDS\footnote{\url{https://github.com/lordzth666/WACV23_PIDS-Joint}.} \dotfill MIT License
    \item WaffleIron\footnote{\url{https://github.com/valeoai/WaffleIron}.} \dotfill Apache License 2.0
    \item PolarSeg\footnote{\url{https://github.com/edwardzhou130/PolarSeg}.} \dotfill BSD 3-Clause License
    \item MinkowskiEngine\footnote{\url{https://github.com/NVIDIA/MinkowskiEngine}.} \dotfill MIT License
    \item Cylinder3D\footnote{\url{https://github.com/xinge008/Cylinder3D}.} \dotfill Apache License 2.0
    \item PyTorch-Scatter\footnote{\url{https://github.com/rusty1s/pytorch_scatter}.} \dotfill MIT License
    \item SpConv\footnote{\url{https://github.com/traveller59/spconv}.} \dotfill Apache License 2.0
    \item TorchSparse\footnote{\url{https://github.com/mit-han-lab/torchsparse}.} \dotfill MIT License
    \item SPVCNN\footnote{\url{https://github.com/mit-han-lab/spvnas}.} \dotfill MIT License
    \item CPGNet\footnote{\url{https://github.com/huixiancheng/No-CPGNet}.} \dotfill Unknown
    \item 2DPASS\footnote{\url{https://github.com/yanx27/2DPASS}.} \dotfill MIT License
    \item GFNet\footnote{\url{https://github.com/haibo-qiu/GFNet}.} \dotfill Unknown
    \item PointPillars\footnote{\url{https://github.com/zhulf0804/PointPillars}.} \dotfill Unknown
    \item second.pytorch\footnote{\url{https://github.com/traveller59/second.pytorch}.} \dotfill MIT License
    \item OpenPCDet\footnote{\url{https://github.com/open-mmlab/OpenPCDet}.} \dotfill Apache License 2.0
    \item PointRCNN\footnote{\url{https://github.com/sshaoshuai/PointRCNN}.} \dotfill MIT License
    \item PartA2-Net\footnote{\url{https://github.com/sshaoshuai/PartA2-Net}.} \dotfill Apache License 2.0
    \item PV-RCNN\footnote{\url{https://github.com/sshaoshuai/PV-RCNN}.} \dotfill Unknown
    \item CenterPoint\footnote{\url{https://github.com/tianweiy/CenterPoint}.} \dotfill MIT License
    \item lidar-camera-robust-benchmark\footnote{\url{https://github.com/kcyu2014/lidar-camera-robust-benchmark}.} \dotfill Unknown
    \item LiDAR-fog-sim\footnote{\url{https://github.com/MartinHahner/LiDAR_fog_sim}.} \dotfill NonCommercial 4.0
    \item LiDAR-snow-sim\footnote{\url{https://github.com/SysCV/LiDAR_snow_sim}.} \dotfill NonCommercial 4.0
    \item mmdetection3d\footnote{\url{https://github.com/open-mmlab/mmdetection3d}.} \dotfill Apache License 2.0
    \item LaserMix\footnote{\url{https://github.com/ldkong1205/LaserMix}.} \dotfill CC BY-NC-SA 4.0
\end{itemize}

\clearpage
\input{tables_supp/semkitti_ce}
\input{tables_supp/semkitti_rr}
\input{tables_supp/semkitti_iou}

\input{tables_supp/kitti_ce}
\input{tables_supp/kitti_rr}
\input{tables_supp/kitti_ap}

\input{tables_supp/nuscenes_seg_ce}
\input{tables_supp/nuscenes_seg_rr}
\input{tables_supp/nuscenes_seg_iou}

\input{tables_supp/nuscenes_det_ce}
\input{tables_supp/nuscenes_det_rr}
\input{tables_supp/nuscenes_det_nds}

\input{tables_supp/wod_seg_ce}
\input{tables_supp/wod_seg_rr}
\input{tables_supp/wod_seg_iou}

\input{tables_supp/wod_det_ce}
\input{tables_supp/wod_det_rr}
\input{tables_supp/wod_det_aph}

\begin{table*}[t]
\caption{The \textbf{Corruption Error (CE)} comparisons between the proposed density-insensitive training framework and the baseline models \cite{2019Minkowski,centerpoint}, on \textit{SemanticKITTI-C} and \textit{WOD-C (Det3D)}, respectively. The task-specific accuracy is mean Intersection-over-Union (mIoU) for 3D semantic segmentation and mean Average Precision (mAPH) for 3D object detection.}
\vspace{0.cm}
\label{tab:framework}
\centering\scalebox{0.85}{
\begin{tabular}{r|c|cccccccc|c}
    \toprule
    \textbf{Method} & \textbf{mCE~$\downarrow$} & \textbf{Fog} & \textbf{Wet} & \textbf{Snow} & \textbf{Motion} & \textbf{Beam} & \textbf{Cross} & \textbf{Echo} & \textbf{Sensor} & \textbf{Acc}
    \\\midrule\midrule
    MinkUNet$_{18}$ \cite{2019Minkowski}, ICA & $100.00$ & $100.00$ & $100.00$ &$100.00$ & $100.00$ & $100.00$ & $100.00$ & $100.00$ & $100.00$ & $62.76$
    \\\midrule
    \textbf{Ours}, ICA & $96.10$ & $104.44$ & $97.09$ & $106.31$ & $85.42$ & $97.34$ & $97.30$ & $95.33$ & $86.08$ & $62.70$
    \\\midrule\midrule
    MinkUNet$_{18}$ \cite{2019Minkowski}, OCA & $86.09$ & $87.67$ & $77.90$ & $82.62$ & $73.82$ & $87.66$ & $97.88$ & $95.99$ & $85.15$ & $69.21$
    \\
    \midrule
    \textbf{Ours}, OCA & $83.23$ & $77.01$ & $101.23$ & $72.53$ & $75.73$ & $79.11$ & $97.78$ & $87.56$ & $66.85$ & $68.13$
    \\\midrule\midrule
    CenterPoint~\cite{centerpoint} & $100.00$ & $100.00$ & $100.00$ &$100.00$ & $100.00$ & $100.00$ & $100.00$ & $100.00$ & $100.00$ & $63.59$
    \\
    \midrule
    \textbf{Ours} & $99.05$ & $99.77$ & $99.52$ & $100.00$ & $100.00$ & $97.26$ & $99.14$ & $99.70$ & $97.02$ & $63.56$
    \\\bottomrule
    \end{tabular}
}
\vspace{0.2cm}
\end{table*}

\clearpage

\clearpage

\clearpage

\clearpage

\clearpage

\clearpage

\clearpage

\clearpage

\clearpage
{\small
\bibliographystyle{ieee_fullname}
\bibliography{egbib}
}

\end{document}

%% file: tables/semkitti_ce.tex
\begin{table}[t]
\caption{The \textbf{Corruption Error (CE)} of \textit{22 segmentors} on \textit{SemanticKITTI-C}. \textbf{Bold}: Best in col. \underline{Underline}: Second best in col. {\small\colorbox{Dark}{Dark}}: Best in row. {\small\colorbox{red!8}{Red}}: Worst in row.}
\vspace{-0.2cm}
\label{tab:semkitti_c_ce}
\centering\scalebox{0.626}{
\begin{tabular}{r|p{26pt}<{\centering}|p{21pt}<{\centering}p{21pt}<{\centering}p{21pt}<{\centering}p{21pt}<{\centering}p{21pt}<{\centering}p{21pt}<{\centering}p{21pt}<{\centering}p{24.5pt}<{\centering}}
    \toprule
    \textbf{Method} & \textbf{mCE~$\downarrow$} & \textbf{Fog} & \textbf{Wet} & \textbf{Snow} & \textbf{Move} & \textbf{Beam} & \textbf{Cross} & \textbf{Echo} & \textbf{Sensor}
    \\\midrule\midrule
    MinkU$_{18}$~\cite{2019Minkowski} & $100.0$ & $100.0$ & $100.0$ &$100.0$ & $100.0$ & $100.0$ & $100.0$ & $100.0$ & $100.0$
    \\\midrule
    SqSeg~\cite{wu2018squeezeseg} & $164.9$ & $183.9$ & $158.0$ & $165.5$ & \cellcolor{Dark}$122.4$ & $171.7$ & \cellcolor{red!9}$188.1$ & $158.7$ & $170.8$
    \\
    SqSegV2~\cite{wu2019squeezesegv2} & $152.5$ & $168.5$ & $141.2$ & $154.6$ & \cellcolor{Dark}$115.2$ & $155.2$ & \cellcolor{red!9}$176.0$ & $145.3$ & $163.5$
    \\
    RGNet$_{21}$~\cite{milioto2019rangenet++} & $136.3$ & \cellcolor{red!9}$156.3$ & $128.5$ & $133.9$ & \cellcolor{Dark}$102.6$ & $141.6$ & $148.9$ & $128.3$ & $150.6$
    \\
    RGNet$_{53}$~\cite{milioto2019rangenet++} & $130.7$ & $144.3$ & $123.7$ & $128.4$ & \cellcolor{Dark}$104.2$ & $135.5$ & $129.4$ & $125.8$ & \cellcolor{red!9}$153.9$
    \\
    SalsaNext~\cite{cortinhal2020salsanext} & $116.1$ & \cellcolor{red!9}$147.5$ & $112.1$ & $116.6$ & \cellcolor{Dark}$77.6$ & $115.3$ & $143.5$ & $114.0$ & $102.5$
    \\
    FIDNet~\cite{zhao2021fidnet} & $113.8$ & $127.7$ & $105.1$ & $107.7$ &  \cellcolor{Dark}$88.9$ & $116.0$ & $121.3$ & $113.7$ & \cellcolor{red!9}$130.0$
    \\
    CENet~\cite{cheng2022cenet} & $103.4$ & $129.8$ & $92.7$ & $99.2$ & \cellcolor{Dark}$70.5$ & $101.2$ & \cellcolor{red!9}$131.1$ & $102.3$ & $100.4$
    \\\midrule
    PolarNet~\cite{zhang2020polarnet} & $118.6$ & $138.8$ & $107.1$ & $108.3$ & \cellcolor{Dark}$86.8$ & $105.1$ & $178.1$ & $112.0$ & $112.3$
    \\\midrule
    KPConv~\cite{thomas2019kpconv} & \underline{$99.5$} & $103.2$ & \underline{$91.9$} & \underline{$98.1$} & $110.7$ & $97.6$ & \cellcolor{red!9}$111.9$ & $97.3$ & \cellcolor{Dark}$85.4$
    \\
    PIDS$_{1.2\times}$~\cite{zhang2023pids} & $104.1$ & \cellcolor{red!9}$118.1$ & $98.9$ & $109.5$ & $114.8$ & $103.2$ & $103.9$ & $97.0$ & \cellcolor{Dark}$87.6$
    \\
    PIDS$_{2.0\times}$~\cite{zhang2023pids} & $101.2$ & $110.6$ & $95.7$ & $104.6$ & \cellcolor{red!9}$115.6$ & $98.6$ & $102.2$ & $97.5$ & \cellcolor{Dark}$84.8$
    \\
    Waffle~\cite{puy23waffleiron} & $109.5$ & $123.5$ & $90.1$ & $108.5$ & $99.9$ & \underline{$93.2$} & \cellcolor{red!9}$186.1$ & $\mathbf{91.0}$ & \cellcolor{Dark}\underline{$84.1$}
    \\\midrule
    MinkU$_{34}$~\cite{2019Minkowski} & $100.6$ & $105.3$ & $99.4$ & \cellcolor{red!9}$106.7$ & $98.7$ & \cellcolor{Dark}$97.6$ & \underline{$99.9$} & $99.0$ & $98.3$
    \\
    Cy3D$_{\text{SPC}}$~\cite{zhu2021cylindrical} & $103.3$ & \cellcolor{red!9}$142.5$ & $92.5$ & $113.6$ & \cellcolor{Dark}$70.9$ & $97.0$ & $105.7$ & $104.2$ & $99.7$
    \\
    Cy3D$_{\text{TSC}}$~\cite{zhu2021cylindrical} & $103.1$ & \cellcolor{red!9}$142.5$ & $101.3$ & $116.9$ & \cellcolor{Dark}\underline{$61.7$} & $98.9$ & $111.4$ & $99.0$ & $93.4$
    \\\midrule
    SPV$_{18}$~\cite{tang2020searching} & $100.3$ & \cellcolor{red!9}\underline{$101.3$} & $100.0$ & $104.0$ & \cellcolor{Dark}$97.6$ & $99.2$ & $100.6$ & $99.6$ & $100.2$
    \\
    SPV$_{34}$~\cite{tang2020searching} & $\mathbf{99.2}$ & $\mathbf{98.5}$ & $100.7$ & \cellcolor{red!9}$102.0$ & \cellcolor{Dark}$97.8$ & $99.0$ & $\mathbf{98.4}$ & $98.8$ & $98.1$
    \\
    RPVNet~\cite{xu2021rpvnet} & $111.7$ & \cellcolor{red!9}$118.7$ & $101.0$ & $104.6$ & \cellcolor{Dark}$78.6$ & $106.4$ & $185.7$ & $99.2$ & $99.8$
    \\
    CPGNet~\cite{li2022cpgnet} & $107.3$ & $141.0$ & $92.6$ & $104.3$ & \cellcolor{Dark}$\mathbf{61.1}$ & $\mathbf{90.9}$ & \cellcolor{red!9}$195.6$ & \underline{$95.0$} & $\mathbf{78.2}$
    \\
    2DPASS~\cite{2022_2DPASS} & $106.1$ & $134.9$ & $\mathbf{85.5}$ & $110.2$ & \cellcolor{Dark}$62.9$ & $94.4$ & \cellcolor{red!9}$171.7$ & $96.9$ & $92.7$
    \\
    GFNet~\cite{qiu2022GFNet} & $108.7$ & $131.3$ & $94.4$ & $\mathbf{92.7}$ & \cellcolor{Dark}$61.7$ & $98.6$ & \cellcolor{red!9}$198.9$ & $98.2$ & $93.6$
    \\\bottomrule
    \end{tabular}
}
\end{table}

%% file: tables/nuscenes_seg_ce.tex
\begin{table}[t]
\caption{The \textbf{Corruption Error (CE)} of \textit{12 segmentors} on \textit{nuScenes-C (Seg3D)}. \textbf{Bold}: Best in col. \underline{Underline}: Second best in col. {\small\colorbox{Dark}{Dark}}: Best in row. {\small\colorbox{red!8}{Red}}: Worst in row.}
\vspace{-0.2cm}
\label{tab:nuscenes_c_seg_ce}
\centering\scalebox{0.626}{
\begin{tabular}{r|p{26pt}<{\centering}|p{21pt}<{\centering}p{21pt}<{\centering}p{21pt}<{\centering}p{21pt}<{\centering}p{21pt}<{\centering}p{21pt}<{\centering}p{21pt}<{\centering}p{24.5pt}<{\centering}}
    \toprule
    \textbf{Method} & \textbf{mCE~$\downarrow$} & \textbf{Fog} & \textbf{Wet} & \textbf{Snow} & \textbf{Move} & \textbf{Beam} & \textbf{Cross} & \textbf{Echo} & \textbf{Sensor}
    \\\midrule\midrule
    MinkU$_{18}$~\cite{2019Minkowski} & $100.0$ & $100.0$ & $100.0$ &$100.0$ & $100.0$ & $100.0$ & $100.0$ & $100.0$ & $100.0$
    \\\midrule
    FIDNet~\cite{zhao2021fidnet} & $122.4$ & $75.9$ & $122.6$ & $68.8$ & \cellcolor{red!9}$192.0$ & $164.8$ & \cellcolor{Dark}$58.0$ & $141.7$ & $155.6$
    \\
    CENet~\cite{cheng2022cenet} & $112.8$ & \underline{$71.2$} & $115.5$ & $64.3$ & \cellcolor{red!9}$156.7$ & $159.0$ & \cellcolor{Dark}\underline{$53.3$} & $129.1$ & $153.4$
    \\\midrule
    PolarNet~\cite{zhang2020polarnet} & $115.1$ & $90.1$ & $115.3$ & \cellcolor{Dark}\underline{$59.0$} & \cellcolor{red!9}$208.2$ & $121.1$ & $80.7$ & $128.2$ & $118.2$
    \\\midrule
    Waffle~\cite{puy23waffleiron} & $106.7$ & $94.7$ & $99.9$ &  \cellcolor{Dark}$84.5$ & \cellcolor{red!9}$152.4$ & $110.7$ & $91.1$ & $106.4$ & $114.2$
    \\\midrule
    MinkU$_{34}$~\cite{2019Minkowski} & \underline{$96.4$} & \cellcolor{Dark}$93.0$ & $96.1$ & \cellcolor{red!9}$104.8$ & $\mathbf{93.1}$ & $\mathbf{95.0}$ & $96.3$ & $\mathbf{96.9}$ & $\mathbf{95.9}$
    \\
    Cy3D$_{\text{SPC}}$~\cite{zhu2021cylindrical} & $111.8$ & $86.6$ & $104.7$ & \cellcolor{Dark}$70.3$ & \cellcolor{red!9}$217.5$ & $113.0$ & $75.7$ & $109.2$ & $117.8$
    \\
    Cy3D$_{\text{TSC}}$~\cite{zhu2021cylindrical} & $105.6$ & $83.2$ & $111.1$ & \cellcolor{Dark}$69.7$ & \cellcolor{red!9}$165.3$ & $114.0$ & $74.4$ & $110.7$ & $116.2$
    \\\midrule
    SPV$_{18}$~\cite{tang2020searching} & $106.7$ & $88.4$ & $105.6$ & $98.8$ & \cellcolor{red!9}$156.5$ & $110.1$ & \cellcolor{Dark}$86.0$ & $104.3$ & $103.6$
    \\
    SPV$_{34}$~\cite{tang2020searching} & $97.5$ & \cellcolor{Dark}$95.2$ & \cellcolor{red!9}$99.5$ & $97.3$ & \underline{$95.3$} & \underline{$98.7$} & $97.9$ & $\mathbf{96.9}$ & \underline{$98.7$}
    \\
    2DPASS~\cite{2022_2DPASS} & $98.6$ & $76.6$ & $\mathbf{89.1}$ & \cellcolor{Dark}$76.4$ & \cellcolor{red!9}$142.7$ & $102.2$ & $89.4$ & \underline{$101.8$} & $110.4$
    \\
    GFNet~\cite{qiu2022GFNet} & $\mathbf{92.6}$ & $\mathbf{65.6}$ & \underline{$93.8$} & $\mathbf{47.2}$ & \cellcolor{red!9}$152.5$ & $112.9$ & \cellcolor{Dark}$\mathbf{45.3}$ & $105.5$ & $117.6$
    \\\bottomrule
    \end{tabular}
}
\end{table}

%% file: tables/wod_seg_ce.tex
\begin{table}[t]
\caption{The \textbf{Corruption Error (CE)} of \textit{5 segmentors} on \textit{WOD-C (Seg3D)}. \textbf{Bold}: Best in col. \underline{Underline}: Second best in col. {\small\colorbox{Dark}{Dark}}: Best in row. {\small\colorbox{red!8}{Red}}: Worst in row.}
\vspace{-0.2cm}
\label{tab:wod_c_seg3d_ce}
\centering\scalebox{0.625}{
\begin{tabular}{r|p{26pt}<{\centering}|p{21pt}<{\centering}p{21pt}<{\centering}p{21pt}<{\centering}p{21pt}<{\centering}p{21pt}<{\centering}p{21pt}<{\centering}p{21pt}<{\centering}p{24.5pt}<{\centering}}
    \toprule
    \textbf{Method} & \textbf{mCE~$\downarrow$} & \textbf{Fog} & \textbf{Wet} & \textbf{Snow} & \textbf{Move} & \textbf{Beam} & \textbf{Cross} & \textbf{Echo} & \textbf{Sensor}
    \\\midrule\midrule
    MinkU$_{18}$~\cite{2019Minkowski} & $100.0$ & $100.0$ & $100.0$ & $100.0$ & $100.0$ & $100.0$ & $100.0$ & $100.0$ & $100.0$
    \\\midrule
    MinkU$_{34}$~\cite{2019Minkowski} & $\mathbf{96.2}$ & $\mathbf{96.0}$ & \underline{$94.9$} & \cellcolor{red!9}$99.5$ & $\mathbf{96.2}$ & $\mathbf{95.4}$ & $\mathbf{96.8}$ & $\mathbf{96.8}$ & \cellcolor{Dark}$\mathbf{94.1}$
    \\
    Cy3D$_{\text{TSC}}$~\cite{zhu2021cylindrical} & $106.0$ & \cellcolor{red!9}$111.8$ & $104.1$ & \cellcolor{Dark}$\mathbf{98.4}$ & $110.3$ & $105.8$ & $106.9$ & $108.2$ & $102.6$
    \\\midrule
    SPV$_{18}$~\cite{tang2020searching} & $103.6$ & \cellcolor{red!9}$105.6$ & $104.8$ & \cellcolor{Dark}\underline{$99.2$} & $105.4$ & $104.8$ & \underline{$99.7$} & $104.3$ & $104.9$
    \\
    SPV$_{34}$~\cite{tang2020searching} & \underline{$98.7$} & \underline{$99.7$} & $\mathbf{96.4}$ & $100.4$ & \underline{$100.0$} & \underline{$98.5$} & \cellcolor{red!9}$101.9$ & \underline{$97.9$} & \cellcolor{Dark}\underline{$95.0$}
    \\\bottomrule
    \end{tabular}
}
\end{table}

%% file: tables/kitti_ce.tex
\begin{table}[t]
\caption{The \textbf{Corruption Error (CE)} of \textit{7 detectors} on \textit{KITTI-C}. \textbf{Bold}: Best in col. \underline{Underline}: Second best in col. {\small\colorbox{Dark}{Dark}}: Best in row. {\small\colorbox{red!8}{Red}}: Worst in row.}
\vspace{-0.2cm}
\label{tab:kitti_c_ce}
\centering\scalebox{0.626}{
\begin{tabular}{r|p{26pt}<{\centering}|p{21pt}<{\centering}p{21pt}<{\centering}p{21pt}<{\centering}p{21pt}<{\centering}p{21pt}<{\centering}p{21pt}<{\centering}p{21pt}<{\centering}p{24.5pt}<{\centering}}
    \toprule
    \textbf{Method} & \textbf{mCE~$\downarrow$} & \textbf{Fog} & \textbf{Wet} & \textbf{Snow} & \textbf{Move} & \textbf{Beam} & \textbf{Cross} & \textbf{Echo} & \textbf{Sensor}
    \\\midrule\midrule
    CenterPP~\cite{centerpoint} & $100.0$ & $100.0$ & $100.0$ & $100.0$ & $100.0$ & $100.0$ & $100.0$ & $100.0$ & $100.0$
    \\\midrule
    SECOND~\cite{second} & $95.9$ & $99.7$ & \cellcolor{red!9}$100.6$ & \cellcolor{Dark}\underline{$87.6$} & $97.6$ & $91.5$ & $96.5$ & $99.2$ & $94.8$
    \\
    P-Pillars~\cite{lang2019pointpillars} & $110.7$ & $115.8$ & $106.4$ & \cellcolor{red!9}$124.9$ & $101.6$ & \cellcolor{Dark}$95.3$ & $117.6$ & $109.9$ & $113.9$
    \\
    P-RCNN~\cite{pointrcnn} & $91.9$ & $93.2$ & $90.1$ & $96.8$ & $93.1$ & $86.1$ & \cellcolor{red!9}$100.9$ & $92.4$ & \cellcolor{Dark}$\mathbf{82.5}$
    \\
    PartA$^2$-F~\cite{parta2} & $\mathbf{82.2}$ & \cellcolor{red!9}$\mathbf{89.4}$ & $\mathbf{75.8}$ & $\mathbf{81.3}$ & $\mathbf{86.2}$ & $\mathbf{80.9}$ & \cellcolor{Dark}$\mathbf{71.8}$ & $\mathbf{83.6}$ & \underline{$88.9$}
    \\
    PartA$^2$-A~\cite{parta2} & \underline{$88.6$} & \underline{$92.6$} & \cellcolor{Dark}\underline{$83.2$} & \cellcolor{red!9}$94.6$ & \underline{$86.4$} & $87.0$ & \cellcolor{Dark}\underline{$83.2$} & \underline{$89.3$} & $92.7$
    \\
    PVRCNN~\cite{pvrcnn} & $90.0$ & \cellcolor{red!9}$95.2$ & $86.6$ & $93.1$ & $87.5$ & \cellcolor{Dark}\underline{$86.0$} & $87.1$ & $90.0$ & $94.7$
    \\\bottomrule
    \end{tabular}
}
\end{table}

%% file: tables/nuscenes_det_ce.tex
\begin{table}[t]
\caption{The \textbf{Corruption Error (CE)} of \textit{5 detectors} on \textit{nuScenes-C (Det3D)}. \textbf{Bold}: Best in col. \underline{Underline}: Second best in col. {\small\colorbox{Dark}{Dark}}: Best in row. {\small\colorbox{red!8}{Red}}: Worst in row.}
\vspace{-0.2cm}
\label{tab:nuscenes_c_det_ce}
\centering\scalebox{0.626}{
\begin{tabular}{r|p{26pt}<{\centering}|p{21pt}<{\centering}p{21pt}<{\centering}p{21pt}<{\centering}p{21pt}<{\centering}p{21pt}<{\centering}p{21pt}<{\centering}p{21pt}<{\centering}p{24.5pt}<{\centering}}
    \toprule
    \textbf{Method} & \textbf{mCE~$\downarrow$} & \textbf{Fog} & \textbf{Wet} & \textbf{Snow} & \textbf{Move} & \textbf{Beam} & \textbf{Cross} & \textbf{Echo} & \textbf{Sensor}
    \\\midrule\midrule
    CenterPP~\cite{centerpoint} & $100.0$ & $100.0$ & $100.0$ &$100.0$ & $100.0$ & $100.0$ & $100.0$ & $100.0$ & $100.0$
    \\\midrule
    SECOND~\cite{second} & \underline{$97.5$} & \underline{$95.4$} & \underline{$96.0$} & \underline{$96.1$} & \underline{$100.8$} & \underline{$99.3$} & \cellcolor{Dark}\underline{$92.2$} & $97.6$ & \cellcolor{red!9}\underline{$102.6$}
    \\
    P-Pillars~\cite{lang2019pointpillars} & $102.9$ & $102.9$ & $104.6$ & $102.5$ & \cellcolor{red!9}$106.4$ & $102.4$ & \cellcolor{Dark}$100.9$ & $102.4$ & $\mathbf{101.1}$
    \\
    CenterLR~\cite{centerpoint} & $98.7$ & $97.9$ & $96.5$ & $97.7$ & $102.2$ & $101.1$ & \cellcolor{Dark}$95.5$ & $\mathbf{95.6}$ & \cellcolor{red!9}$103.5$
    \\
    CenterHR~\cite{centerpoint} & $\mathbf{95.8}$ & $\mathbf{93.0}$ & $\mathbf{92.0}$ & $\mathbf{94.9}$ & $\mathbf{97.6}$ & $\mathbf{98.4}$ & \cellcolor{Dark}$\mathbf{91.1}$ & \underline{$96.2$} & \cellcolor{red!9}$103.2$
    \\\bottomrule
    \end{tabular}
}
\end{table}

%% file: tables/wod_det_ce.tex
\begin{table}[t]
\caption{The \textbf{Corruption Error (CE)} of \textit{5 detectors} on \textit{WOD-C (Det3D)}. \textbf{Bold}: Best in col. \underline{Underline}: Second best in col. {\small\colorbox{Dark}{Dark}}: Best in row. {\small\colorbox{red!8}{Red}}: Worst in row.}
\vspace{-0.2cm}
\label{tab:wod_c_det3d_ce}
\centering\scalebox{0.626}{
\begin{tabular}{r|p{26pt}<{\centering}|p{21pt}<{\centering}p{21pt}<{\centering}p{21pt}<{\centering}p{21pt}<{\centering}p{21pt}<{\centering}p{21pt}<{\centering}p{21pt}<{\centering}p{24.5pt}<{\centering}}
    \toprule
    \textbf{Method} & \textbf{mCE~$\downarrow$} & \textbf{Fog} & \textbf{Wet} & \textbf{Snow} & \textbf{Move} & \textbf{Beam} & \textbf{Cross} & \textbf{Echo} & \textbf{Sensor}
    \\\midrule\midrule
    CenterPP~\cite{centerpoint} & $100.0$ & $100.0$ & $100.0$ & $100.0$ & $100.0$ & $100.0$ & $100.0$ & $100.0$ & $100.0$
    \\\midrule
    SECOND~\cite{second} & $121.4$ & $117.9$ & $126.5$ & $127.5$ & \cellcolor{Dark}$113.4$ & $121.3$ & \cellcolor{red!9}$127.8$ & $123.7$ & $113.5$
    \\
    P-Pillars~\cite{lang2019pointpillars} & $127.5$ & $120.8$ & $135.2$ & $129.7$ & $115.2$ & $123.0$ & \cellcolor{red!9}$151.7$ & $131.6$ & \cellcolor{Dark}$113.1$
    \\
    PVRCNN~\cite{pvrcnn} & \underline{$104.9$} & \underline{$110.1$} & \underline{$104.2$} & \cellcolor{Dark}\underline{$95.7$} & \underline{$101.3$} & \cellcolor{red!9}\underline{$110.7$} & \underline{$101.8$} & \underline{$106.0$} & \underline{$109.4$}
    \\
    PV++~\cite{pvrcnn++} & $\mathbf{91.6}$ & \cellcolor{red!9}$\mathbf{95.7}$ & \cellcolor{Dark}$\mathbf{88.3}$ & $\mathbf{90.1}$ & $\mathbf{93.2}$ & $\mathbf{92.5}$ & $\mathbf{88.9}$ & $\mathbf{90.8}$ & $\mathbf{93.2}$
    \\\bottomrule
    \end{tabular}
}
\end{table}

%% file: tables/ablation_optim.tex
\begin{table}[t]
\caption{Ablation study on: \textbf{[left]} in-context (ICA) and out-of-context (OCA) augmentations; \textbf{[middle]} voxelization strategies; and \textbf{[right]} density-insensitive training losses.}
\vspace{-0.2cm}
\label{tab:dv_aug_mask}
\centering\scalebox{0.647}{
\begin{tabular}{c|p{19.8pt}<{\centering}p{19.8pt}<{\centering}|c|cc|p{26pt}<{\centering}p{26pt}<{\centering}p{26pt}<{\centering}}
    \toprule
    \textbf{Method} & \textbf{ICA} & \textbf{OCA} & \textbf{Size} & $\mathcal{L}_{\text{part2full}}$ & $\mathcal{L}_{\text{full2part}}$ & \textbf{mCE}~$\downarrow$ & \textbf{mRR}~$\uparrow$ & \textbf{Clean}
    \\\midrule\midrule
    \cellcolor{red!9}Base~\cite{2019Minkowski} & \cellcolor{red!9}\checkmark & \cellcolor{red!9} & \cellcolor{red!9}Fixed & \cellcolor{red!9} & \cellcolor{red!9} & \cellcolor{red!9}$100.0$ & \cellcolor{red!9}$81.9$ & \cellcolor{red!9}\underline{$62.8$}
    \\\midrule
    Ours - (1) & \checkmark & & Flexible & & & $97.4$ & $84.2$ & $\mathbf{62.9}$
    \\
    Ours - (2) & \checkmark & & Flexible & \checkmark & & \underline{$96.4$} & \underline{$85.1$} & $62.7$
    \\
    Ours - (3) & \checkmark & & Flexible & \checkmark & \checkmark & $\mathbf{96.1}$ & $\mathbf{85.6}$ & $62.7$ 
    \\\midrule
    \cellcolor{red!9}Base~\cite{2019Minkowski} & \cellcolor{red!9} & \cellcolor{red!9}\checkmark & \cellcolor{red!9}Fixed & \cellcolor{red!9} & \cellcolor{red!9} & \cellcolor{red!9}$86.0$ & \cellcolor{red!9}$84.7$ & \cellcolor{red!9}$\mathbf{69.2}$
    \\\midrule
    Ours - (4) & & \checkmark & Flexible & & & $84.5$ & $86.8$ & \underline{$68.2$}
    \\
    Ours - (5) & & \checkmark & Flexible & \checkmark & & \underline{$83.8$} & \underline{$88.1$} & $67.9$
    \\
    Ours - (6) & & \checkmark & Flexible & \checkmark & \checkmark & $\mathbf{83.2}$ & $\mathbf{89.7}$ & $68.1$
    \\\bottomrule
    \end{tabular}
}
\vspace{-0.0cm}
\end{table}

%% file: tables_supp/semkitti_ce.tex
\begin{table*}[t]
\caption{[Complete Results] The \textbf{Corruption Error (CE)} of each method on \textit{SemanticKITTI-C}. \textbf{Bold}: Best in column. \underline{Underline}: Second best in column. All scores are given in percentage ($\%$). \colorbox{Dark}{Dark}: Best in row. \colorbox{red!9}{Red}: Worst in row. Symbol $^{\dagger}$ denotes the baseline model adopted in calculating the CE scores.}
\vspace{0.cm}
\label{tab:semkitti_c_ce_supp}
\centering\scalebox{0.837}{
\begin{tabular}{r|c|cccccccc|c}
    \toprule
    \textbf{Method} & \textbf{mCE~$\downarrow$} & \textbf{Fog} & \textbf{Wet} & \textbf{Snow} & \textbf{Motion} & \textbf{Beam} & \textbf{Cross} & \textbf{Echo} & \textbf{Sensor} & \textbf{mIoU~$\uparrow$}
    \\\midrule\midrule
    MinkUNet$_{18}$$^{\dagger}$~\cite{2019Minkowski} & $100.00$ & $100.00$ & $100.00$ &$100.00$ & $100.00$ & $100.00$ & $100.00$ & $100.00$ & $100.00$ & $62.76$
    \\\midrule
    SqueezeSeg~\cite{wu2018squeezeseg} & $164.87$ & $183.89$ & $158.01$ & $165.45$ & \cellcolor{Dark}$122.35$ & $171.68$ & \cellcolor{red!9}$188.07$ & $158.74$ & $170.81$ & $31.61$
    \\
    SqueezeSegV2~\cite{wu2019squeezesegv2} & $152.45$ & $168.50$ & $141.23$ & $154.64$ & \cellcolor{Dark}$115.16$ & $155.24$ & \cellcolor{red!9}$176.00$ & $145.27$ & $163.52$ & $41.28$
    \\
    RangeNet$_{21}$~\cite{milioto2019rangenet++} & $136.33$ & \cellcolor{red!9}$156.27$ & $128.49$ & $133.93$ & \cellcolor{Dark}$102.62$ & $141.58$ & $148.87$ & $128.29$ & $150.58$ & $47.15$
    \\
    RangeNet$_{53}$~\cite{milioto2019rangenet++} & $130.66$ & $144.28$ & $123.73$ & $128.38$ & \cellcolor{Dark}$104.20$ & $135.53$ & $129.43$ & $125.81$ & \cellcolor{red!9}$153.88$ & $50.29$
    \\
    SalsaNext~\cite{cortinhal2020salsanext} & $116.14$ & \cellcolor{red!9}$147.54$ & $112.06$ & $116.55$ & \cellcolor{Dark}$77.62$ & $115.32$ & $143.52$ & $114.04$ & $102.47$ & $55.80$
    \\
    FIDNet~\cite{zhao2021fidnet} & $113.81$ & $127.67$ & $105.13$ & $107.71$ &  \cellcolor{Dark}$88.88$ & $116.03$ & $121.32$ & $113.74$ & \cellcolor{red!9}$130.03$ & $58.80$
    \\
\bottomrule
    \end{tabular}
}
\vspace{0.2cm}
\end{table*}

%% file: tables_supp/semkitti_rr.tex
\begin{table*}[t]
\caption{[Complete Results] The \textbf{Resilience Rate (RR)} of each method on \textit{SemanticKITTI-C}. \textbf{Bold}: Best in column. \underline{Underline}: Second best in column. All scores are given in percentage ($\%$). \colorbox{Dark}{Dark}: Best in row. \colorbox{red!9}{Red}: Worst in row.}
\vspace{0.cm}
\label{tab:semkitti_c_rr}
\centering\scalebox{0.837}{
\begin{tabular}{r|c|cccccccc|c}
    \toprule
    \textbf{Method} & \textbf{mRR~$\uparrow$} & \textbf{Fog} & \textbf{Wet} & \textbf{Snow} & \textbf{Motion} & \textbf{Beam} & \textbf{Cross} & \textbf{Echo} & \textbf{Sensor} & \textbf{mIoU~$\uparrow$}
    \\\midrule\midrule
    SqueezeSeg~\cite{wu2018squeezeseg} & $66.81$ & $59.63$ & $86.37$ & $71.81$ & $56.72$ & $79.12$ & $68.49$ & \cellcolor{Dark}$87.50$ & \cellcolor{red!9}$24.83$ & $31.61$
    \\
    SqueezeSegV2~\cite{wu2019squeezesegv2} & $65.29$ & $62.11$ & \cellcolor{Dark}$84.84$ & $67.22$ & $55.11$ & $77.98$ & $64.63$ & $81.88$ & \cellcolor{red!9}$28.54$ & $41.28$
    \\
    RangeNet$_{21}$~\cite{milioto2019rangenet++} & $73.42$ & $65.83$ & $86.70$ & $79.38$ & $66.09$ & $80.93$ & $80.55$ & \cellcolor{Dark}$88.10$ & \cellcolor{red!9}$39.79$ & $47.15$
    \\
    RangeNet$_{53}$~\cite{milioto2019rangenet++} & $73.59$ & $72.24$ & $85.64$ & $79.58$ & $59.85$ & $81.13$ & \cellcolor{Dark}$91.63$ & $84.85$ & \cellcolor{red!9}$33.76$ & $50.29$
    \\
    SalsaNext~\cite{cortinhal2020salsanext} & $80.51$ & \cellcolor{red!9}$62.53$ & $86.81$ & $81.63$ & $85.90$ & \cellcolor{Dark}$88.94$ & $72.06$ & $86.08$ & $80.14$ & $55.80$
    \\
    FIDNet~\cite{zhao2021fidnet} & $76.99$ & $74.25$ & \cellcolor{Dark}$87.81$ & $84.49$ & $68.67$ & $83.88$ & $84.12$ & $81.92$ & \cellcolor{red!9}$50.77$ & $58.80$
    \\
    CENet~\cite{cheng2022cenet} & $81.29$ & \cellcolor{red!9}$68.27$ & \cellcolor{Dark}$91.67$ & $85.76$ & $84.27$ & $89.18$ & $72.53$ & $85.37$ & $73.29$ & $62.55$
    \\
\bottomrule
    \end{tabular}
}
\vspace{0.2cm}
\end{table*}

%% file: tables_supp/semkitti_iou.tex
\begin{table*}[t]
\caption{[Complete Results] The \textbf{Intersection-over-Union (IoU)} of each method on \textit{SemanticKITTI-C}. \textbf{Bold}: Best in column. \underline{Underline}: Second best in column. All scores are given in percentage ($\%$). \colorbox{Dark}{Dark}: Best in row. \colorbox{red!9}{Red}: Worst in row.}
\vspace{0.cm}
\label{tab:semkitti_c_iou}
\centering\scalebox{0.837}{
\begin{tabular}{r|c|c|c|cccccccc}
    \toprule
    \textbf{Method} & \textbf{mCE~$\downarrow$} & \textbf{mRR~$\uparrow$} & \textbf{mIoU~$\uparrow$} & \textbf{Fog} & \textbf{Wet} & \textbf{Snow} & \textbf{Motion} & \textbf{Beam} & \textbf{Cross} & \textbf{Echo} & \textbf{Sensor}
    \\\midrule\midrule
    SqueezeSeg~\cite{wu2018squeezeseg} & $164.87$ & $66.81$ & $31.61$ & $18.85$ & $27.30$ & $22.70$ & $17.93$ & $25.01$ & $21.65$ & \cellcolor{Dark}$27.66$ & \cellcolor{red!9}$7.85$
    \\
    SqueezeSegV2~\cite{wu2019squeezesegv2} & $152.45$ & $65.29$ & $41.28$ & $25.64$ & \cellcolor{Dark}$35.02$ & $27.75$ & $22.75$ & $32.19$ & $26.68$ & $33.80$ & \cellcolor{red!9}$11.78$
    \\
    RangeNet$_{21}$~\cite{milioto2019rangenet++} & $136.33$ & $73.42$ & $47.15$ & $31.04$ & $40.88$ & $37.43$ & $31.16$ & $38.16$ & $37.98$ & \cellcolor{Dark}$41.54$ & \cellcolor{red!9}$18.76$
    \\
    RangeNet$_{53}$~\cite{milioto2019rangenet++}  & $130.66$ & $73.59$ & $50.29$ & $36.33$ & $43.07$ & $40.02$ & $30.10$ & $40.80$ & \cellcolor{Dark}$46.08$ & $42.67$ & \cellcolor{red!9}$16.98$
    \\
    SalsaNext~\cite{cortinhal2020salsanext} & $116.14$ & $80.51$ & $55.80$ & \cellcolor{red!9}$34.89$ & $48.44$ & $45.55$ & $47.93$ & \cellcolor{Dark}$49.63$ & $40.21$ & $48.03$ & $44.72$
    \\
    FIDNet~\cite{zhao2021fidnet} & $113.81$ & $76.99$ & $58.80$ & $43.66$ & \cellcolor{Dark}$51.63$ & $49.68$ & $40.38$ & $49.32$ & $49.46$ & $48.17$ & \cellcolor{red!9}$29.85$
    \\
    CENet~\cite{cheng2022cenet} & $103.41$ & $81.29$ & $62.55$ & \cellcolor{red!9}$42.70$ & \cellcolor{Dark}$57.34$ & $53.64$ & $52.71$ & $55.78$ & $45.37$ & $53.40$ & $45.84$
    \\
\bottomrule
    \end{tabular}
}
\vspace{0.2cm}
\end{table*}

%% file: tables_supp/kitti_ce.tex
\begin{table*}[t]
\caption{[Complete Results] The \textbf{Corruption Error (CE)} of each method on \textit{KITTI-C}. \textbf{Bold}: Best in column. \underline{Underline}: Second best in column. All scores are given in percentage ($\%$). \colorbox{Dark}{Dark}: Best in row. \colorbox{red!9}{Red}: Worst in row. Symbol $^{\dagger}$ denotes the baseline model adopted in calculating the CE scores.}
\vspace{0.cm}
\label{tab:kitti_c_ce_supp}
\centering\scalebox{0.837}{
\begin{tabular}{r|c|cccccccc|c}
    \toprule
    \textbf{Method} & \textbf{mCE~$\downarrow$} & \textbf{Fog} & \textbf{Wet} & \textbf{Snow} & \textbf{Motion} & \textbf{Beam} & \textbf{Cross} & \textbf{Echo} & \textbf{Sensor} & \textbf{mAP~$\uparrow$}
    \\\midrule\midrule
    CenterPoint$^{\dagger}$~\cite{centerpoint} & $100.00$ & $100.00$ & $100.00$ & $100.00$ & $100.00$ & $100.00$ & $100.00$ & $100.00$ & $100.00$ & $68.70$
    \\\midrule
    SECOND~\cite{second} & $95.93$ & $99.70$ & \cellcolor{red!9}$100.64$ & \cellcolor{Dark}\underline{$87.64$} & $97.60$ & $91.50$ & $96.50$ & $99.15$ & $94.75$ & $68.49$
    \\
    PointPillars~\cite{lang2019pointpillars} & $110.67$ & $115.78$ & $106.39$ & \cellcolor{red!9}$124.86$ & $101.63$ & \cellcolor{Dark}$95.29$ & $117.62$ & $109.88$ & $113.88$ & $66.70$
    \\
    PointRCNN~\cite{pointrcnn} & $91.88$ & $93.16$ & $90.06$ & $96.81$ & $93.12$ & $86.11$ & \cellcolor{red!9}$100.88$ & $92.41$ & \cellcolor{Dark}$\mathbf{82.49}$ & $70.26$
    \\
    Part-A2$_\text{Free}$~\cite{parta2} & $\mathbf{82.22}$ & \cellcolor{red!9}$\mathbf{89.42}$ & $\mathbf{75.78}$ & $\mathbf{81.32}$ & $\mathbf{86.15}$ & $\mathbf{80.89}$ & \cellcolor{Dark}$\mathbf{71.79}$ & $\mathbf{83.55}$ & \underline{$88.88$} & $\mathbf{76.28}$
    \\
    Part-A2$_\text{Anchor}$~\cite{parta2} & \underline{$88.62$} & \underline{$92.56$} & \underline{$83.19$} & \cellcolor{red!9}$94.63$ & \underline{$86.36$} & $87.03$ & \cellcolor{Dark}\underline{$83.18$} & \underline{$89.32$} & $92.66$ & \underline{$73.98$}
    \\
    PV-RCNN~\cite{pvrcnn} & $90.04$ & $95.18$ & $86.64$ & \cellcolor{red!9}$93.08$ & $87.51$ & \cellcolor{Dark}\underline{$86.03$} & $87.09$ & $90.02$ & $94.73$ & $72.36$
    \\\bottomrule
    \end{tabular}
}
\vspace{0.2cm}
\end{table*}

%% file: tables_supp/kitti_rr.tex
\begin{table*}[t]
\caption{[Complete Results] The \textbf{Resilience Rate (RR)} of each method on \textit{KITTI-C}. \textbf{Bold}: Best in column. \underline{Underline}: Second best in column. All scores are given in percentage ($\%$). \colorbox{Dark}{Dark}: Best in row. \colorbox{red!9}{Red}: Worst in row.}
\vspace{0.cm}
\label{tab:kitti_c_rr}
\centering\scalebox{0.837}{
\begin{tabular}{r|c|cccccccc|c}
    \toprule
    \textbf{Method} & \textbf{mRR~$\uparrow$} & \textbf{Fog} & \textbf{Wet} & \textbf{Snow} & \textbf{Motion} & \textbf{Beam} & \textbf{Cross} & \textbf{Echo} & \textbf{Sensor} & \textbf{mAP~$\uparrow$}
    \\\midrule\midrule
    PointPillars~\cite{lang2019pointpillars} & $74.94$ & $68.52$ & \cellcolor{Dark}$100.01$ & \cellcolor{red!9}$53.63$ & $70.60$ & $78.32$ & $89.97$ & $82.22$ & $56.22$ & $66.70$
    \\
    SECOND~\cite{second} & \underline{$82.94$} & \underline{$77.73$} & \cellcolor{Dark}$100.03$ & $\mathbf{80.19}$ & $71.82$ & \underline{$79.05$} & \underline{$98.10$} & $86.51$ & \cellcolor{red!9}\underline{$70.08$} & $68.49$
    \\
    PointRCNN~\cite{pointrcnn} & $\mathbf{83.46}$ & $\mathbf{80.15}$ & \cellcolor{Dark}$\mathbf{102.22}$ & \cellcolor{red!9}$71.45$ & $73.33$ & $\mathbf{80.90}$ & $93.51$ & $\mathbf{88.27}$ & $\mathbf{77.90}$ & $70.26$
    \\
    Part-A2$_\text{Free}$~\cite{parta2} & $81.87$ & $76.11$ & \cellcolor{Dark}$100.01$ & \underline{$76.26$} & $72.30$ & $77.95$ & $\mathbf{99.10}$ & $86.08$ & \cellcolor{red!9}$67.15$ & $\mathbf{76.28}$
    \\
    Part-A2$_\text{Anchor}$~\cite{parta2} & $80.67$ & $76.49$ & \cellcolor{Dark}$99.99$ & $69.37$ & \underline{$74.40$} & $76.21$ & $96.95$ & $85.55$ & \cellcolor{red!9}$66.44$ & \underline{$73.98$}
    \\
    PV-RCNN~\cite{pvrcnn} & $81.73$ & $76.51$ & \cellcolor{Dark}\underline{$100.73$} & $72.03$ & $\mathbf{75.23}$ & $78.61$ & $97.28$ & \underline{$87.06$} & \cellcolor{red!9}$66.35$ & $72.36$
    \\
    CenterPoint~\cite{centerpoint} & $79.73$ & $77.29$ & \cellcolor{Dark}$100.01$ & $70.68$ & $69.78$ & $72.61$ & $96.07$ & $85.74$ & \cellcolor{red!9}$65.68$ & $68.70$
    \\\bottomrule
    \end{tabular}
}
\vspace{0.2cm}
\end{table*}

%% file: tables_supp/kitti_ap.tex
\begin{table*}[t]
\caption{[Complete Results] The \textbf{Average Precision (AP)} of each method on \textit{KITTI-C}. \textbf{Bold}: Best in column. \underline{Underline}: Second best in column. All scores are given in percentage ($\%$). \colorbox{Dark}{Dark}: Best in row. \colorbox{red!9}{Red}: Worst in row.}
\vspace{0.cm}
\label{tab:kitti_c_ap}
\centering\scalebox{0.837}{
\begin{tabular}{r|c|c|c|cccccccc}
    \toprule
    \textbf{Method} & \textbf{mCE~$\downarrow$} & \textbf{mRR~$\uparrow$} & \textbf{mAP~$\uparrow$} & \textbf{Fog} & \textbf{Wet} & \textbf{Snow} & \textbf{Motion} & \textbf{Beam} & \textbf{Cross} & \textbf{Echo} & \textbf{Sensor}
    \\\midrule\midrule
    PointPillars~\cite{lang2019pointpillars} & $110.67$ & $74.94$ & $66.70$ & $45.70$ & \cellcolor{Dark}$66.71$ & \cellcolor{red!9}$35.77$ & $47.09$ & $52.24$ & $60.01$ & $54.84$ & $37.50$
    \\
    SECOND~\cite{second} & $95.93$ & \underline{$82.94$} & $68.49$ & $53.24$ & \cellcolor{Dark}$68.51$ & \underline{$54.92$} & $49.19$ & $54.14$ & $67.19$ & $59.25$ & \cellcolor{red!9}$48.00$
    \\
    PointRCNN~\cite{pointrcnn} & $91.88$ & $\mathbf{83.46}$ & $70.26$ & $56.31$ & \cellcolor{Dark}$71.82$ & \cellcolor{red!9}$50.20$ & $51.52$ & $56.84$ & $65.70$ & $62.02$ & $\mathbf{54.73}$
    \\
    Part-A2$_\text{Free}$~\cite{parta2} & $\mathbf{82.22}$ & $81.87$ & $\mathbf{76.28}$ & $\mathbf{58.06}$ & \cellcolor{Dark}$\mathbf{76.29}$ & $\mathbf{58.17}$ & $\mathbf{55.15}$ & $\mathbf{59.46}$ & $\mathbf{75.59}$ & $\mathbf{65.66}$ & \cellcolor{red!9}\underline{$51.22$}
    \\
    Part-A2$_\text{Anchor}$~\cite{parta2} & \underline{$88.62$} & $80.67$ & \underline{$73.98$} & \underline{$56.59$} & \cellcolor{Dark}\underline{$73.97$} & $51.32$ & \underline{$55.04$} & $56.38$ & \underline{$71.72$} & \underline{$63.29$} & \cellcolor{red!9}$49.15$
    \\
    PV-RCNN~\cite{pvrcnn} & $90.04$ & $81.73$ & $72.36$ & $55.36$ & \cellcolor{Dark}$72.89$ & $52.12$ & $54.44$ & \underline{$56.88$} & $70.39$ & $63.00$ & \cellcolor{red!9}$48.01$
    \\
    CenterPoint~\cite{centerpoint} & $100.00$ & $79.73$ & $68.70$ & $53.10$ & \cellcolor{Dark}$68.71$ & $48.56$ & $47.94$ & $49.88$ & $66.00$ & $58.90$ & \cellcolor{red!9}$45.12$
    \\\bottomrule
    \end{tabular}
}
\vspace{0.2cm}
\end{table*}

%% file: tables_supp/nuscenes_seg_ce.tex
\begin{table*}[t]
\caption{[Complete Results] The \textbf{Corruption Error (CE)} of each method on \textit{nuScenes-C (Seg3D)}. \textbf{Bold}: Best in column. \underline{Underline}: Second best in column. All scores are given in percentage ($\%$). \colorbox{Dark}{Dark}: Best in row. \colorbox{red!9}{Red}: Worst in row.}
\vspace{0.cm}
\label{tab:nuscenes_c_seg_ce_supp}
\centering\scalebox{0.837}{
\begin{tabular}{r|c|cccccccc|c}
    \toprule
    \textbf{Method} & \textbf{mCE~$\downarrow$} & \textbf{Fog} & \textbf{Wet} & \textbf{Snow} & \textbf{Motion} & \textbf{Beam} & \textbf{Cross} & \textbf{Echo} & \textbf{Sensor} & \textbf{mIoU~$\uparrow$}
    \\\midrule\midrule
    MinkUNet$_{18}$$^{\dagger}$~\cite{2019Minkowski} & $100.00$ & $100.00$ & $100.00$ &$100.00$ & $100.00$ & $100.00$ & $100.00$ & $100.00$ & $100.00$ & $75.76$
    \\\midrule
    FIDNet~\cite{zhao2021fidnet} & $122.42$ & $75.93$ & $122.58$ & $68.78$ & \cellcolor{red!9}$192.03$ & $164.84$ & \cellcolor{Dark}$57.95$ & $141.66$ & $155.56$ & $71.38$
    \\
    CENet~\cite{cheng2022cenet} & $112.79$ & \underline{$71.16$} & $115.48$ & $64.31$ & \cellcolor{red!9}$156.67$ & $159.03$ & \cellcolor{Dark}\underline{$53.27$} & $129.08$ & $153.35$ & $73.28$
    \\\midrule
    PolarNet~\cite{zhang2020polarnet} & $115.09$ & $90.10$ & $115.33$ & \cellcolor{Dark}\underline{$58.98$} & \cellcolor{red!9}$208.19$ & $121.07$ & $80.67$ & $128.17$ & $118.23$ & $71.37$
    \\\midrule
    WaffleIron~\cite{puy23waffleiron} & $106.73$ & $94.76$ & $99.92$ &  \cellcolor{Dark}$84.51$ & \cellcolor{red!9}$152.35$ & $110.65$ & $91.09$ & $106.41$ & $114.15$ & $76.07$
    \\\midrule
    MinkUNet$_{34}$~\cite{2019Minkowski} & \underline{$96.37$} & \cellcolor{Dark}$92.95$ & $96.09$ & \cellcolor{red!9}$104.78$ & $\mathbf{93.05}$ & $\mathbf{95.04}$ & $96.27$ & $\mathbf{96.88}$ & $\mathbf{95.90}$ & \underline{$76.90$}
    \\
    Cylinder3D$_{\text{SPC}}$~\cite{zhu2021cylindrical} & $111.84$ & $86.60$ & $104.68$ & \cellcolor{Dark}$70.29$ & \cellcolor{red!9}$217.47$ & $113.00$ & $75.67$ & $109.21$ & $117.78$ & $76.15$
    \\
\bottomrule
    \end{tabular}
}
\vspace{0.2cm}
\end{table*}

%% file: tables_supp/nuscenes_seg_rr.tex
\begin{table*}[t]
\caption{[Complete Results] The \textbf{Resilience Rate (RR)} of each method on \textit{nuScenes-C (Seg3D)}. \textbf{Bold}: Best in column. \underline{Underline}: Second best in column. All scores are given in percentage ($\%$). \colorbox{Dark}{Dark}: Best in row. \colorbox{red!9}{Red}: Worst in row.}
\vspace{0.cm}
\label{tab:nuscenes_c_seg_rr}
\centering\scalebox{0.837}{
\begin{tabular}{r|c|cccccccc|c}
    \toprule
    \textbf{Method} & \textbf{mRR~$\uparrow$} & \textbf{Fog} & \textbf{Wet} & \textbf{Snow} & \textbf{Motion} & \textbf{Beam} & \textbf{Cross} & \textbf{Echo} & \textbf{Sensor} & \textbf{mIoU~$\uparrow$}
    \\\midrule\midrule
    FIDNet~\cite{zhao2021fidnet} & $73.33$ & \underline{$90.78$} & \cellcolor{Dark}$95.29$ & $82.61$ & $68.51$ & $67.44$ & $80.48$ & $68.31$ & \cellcolor{red!9}$33.20$ & $71.38$
    \\
    CENet~\cite{cheng2022cenet} & $76.04$ & $\mathbf{91.44}$ & \cellcolor{Dark}$95.35$ & $84.12$ & $79.57$ & $68.19$ & \underline{$83.09$} & $72.75$ & \cellcolor{red!9}$33.82$ & $73.28$
    \\\midrule
    PolarNet~\cite{zhang2020polarnet} & $76.34$ & $81.59$ & \cellcolor{Dark}$97.95$ & \underline{$90.82$} & $62.49$ & $86.75$ & \cellcolor{red!9}$57.12$ & $75.16$ & $58.86$ & $71.37$
    \\\midrule
    WaffleIron~\cite{puy23waffleiron} & $72.78$ & $73.71$ & \cellcolor{Dark}$97.19$ &  $65.19$ & $78.16$ & $85.70$ & \cellcolor{red!9}$43.54$ & $80.86$ & $57.85$ & $76.07$
    \\\midrule
    MinkUNet$_{18}$~\cite{2019Minkowski} & $74.44$ & $70.80$ & \cellcolor{Dark}$97.56$ & $53.26$ & $96.87$ & $90.47$ & \cellcolor{red!9}$35.08$ & $84.25$ & $67.25$ & $75.76$
    \\
    MinkUNet$_{34}$~\cite{2019Minkowski} & $75.08$ & $74.01$ & $97.44$ & $48.76$ & \cellcolor{Dark}$\mathbf{97.84}$ & $\mathbf{91.16}$ & \cellcolor{red!9}$38.13$ & \underline{$84.47$} & $\mathbf{68.87}$ & \underline{$76.90$}
    \\
    Cylinder3D$_{\text{SPC}}$~\cite{zhu2021cylindrical} & $72.94$ & $78.59$ & \cellcolor{Dark}$95.46$ & $76.26$ & \cellcolor{red!9}$55.33$ & $84.64$ & $58.36$ & $79.45$ & $55.46$ & $76.15$
    \\
\bottomrule
    \end{tabular}
}
\vspace{0.2cm}
\end{table*}

%% file: tables_supp/nuscenes_seg_iou.tex
\begin{table*}[t]
\caption{[Complete Results] The \textbf{Intersection-over-Union (IoU)} of each method on \textit{nuScenes-C (Seg3D)}. \textbf{Bold}: Best in column. \underline{Underline}: Second best in column. All scores are given in percentage ($\%$). \colorbox{Dark}{Dark}: Best in row. \colorbox{red!9}{Red}: Worst in row.}
\vspace{0.cm}
\label{tab:nuscenes_c_seg_iou}
\centering\scalebox{0.837}{
\begin{tabular}{r|c|c|c|cccccccc}
    \toprule
    \textbf{Method} & \textbf{mCE~$\downarrow$} & \textbf{mRR~$\uparrow$} & \textbf{mIoU~$\uparrow$} & \textbf{Fog} & \textbf{Wet} & \textbf{Snow} & \textbf{Motion} & \textbf{Beam} & \textbf{Cross} & \textbf{Echo} & \textbf{Sensor}
    \\\midrule\midrule
    FIDNet~\cite{zhao2021fidnet} & $122.42$ & $73.33$ & $71.38$ & $64.80$ & \cellcolor{Dark}$68.02$ & $58.97$ & $48.90$ & $48.14$ & $57.45$ & $48.76$ & \cellcolor{red!9}$23.70$
    \\
    CENet~\cite{cheng2022cenet} & $112.79$ & $76.04$ & $73.28$ & \underline{$67.01$} & \cellcolor{Dark}$69.87$ & $61.64$ & $58.31$ & $49.97$ & \underline{$60.89$} & $53.31$ & \cellcolor{red!9}$24.78$
    \\\midrule
    PolarNet~\cite{zhang2020polarnet} & $115.09$ & $76.34$ & $71.37$ & $58.23$ & \cellcolor{Dark}$69.91$ & \underline{$64.82$} & $44.60$ & $61.91$ & \cellcolor{red!9}$40.77$ & $53.64$ & $42.01$
    \\\midrule
    WaffleIron~\cite{puy23waffleiron} & $106.73$ & $72.78$ & $76.07$ & $56.07$ & \cellcolor{Dark}$73.93$ & $49.59$ & $59.46$ & $65.19$ & \cellcolor{red!9}$33.12$ & $61.51$ & $44.01$
    \\\midrule
    MinkUNet$_{18}$~\cite{2019Minkowski} & $100.00$ & $74.44$ & $75.76$ & $53.64$ & \cellcolor{Dark}$73.91$ & $40.35$ & $73.39$ & $68.54$ & \cellcolor{red!9}$26.58$ & \underline{$63.83$} & $50.95$
    \\
    MinkUNet$_{34}$~\cite{2019Minkowski} & \underline{$96.37$} & $75.08$ & \underline{$76.90$} & $56.91$ & $74.93$ & $37.50$ & \cellcolor{Dark}$\mathbf{75.24}$ & $\mathbf{70.10}$ & \cellcolor{red!9}$29.32$ & $\mathbf{64.96}$ & $\mathbf{52.96}$
    \\
    Cylinder3D$_{\text{SPC}}$~\cite{zhu2021cylindrical} & $111.84$ & $72.94$ & $76.15$ & $59.85$ & \cellcolor{Dark}$72.69$ & $58.07$ & \cellcolor{red!9}$42.13$ & $64.45$ & $44.44$ & $60.50$ & $42.23$
    \\
\bottomrule
    \end{tabular}
}
\vspace{0.2cm}
\end{table*}

%% file: tables_supp/nuscenes_det_ce.tex
\begin{table*}[t]
\caption{[Complete Results] The \textbf{Corruption Error (CE)} of each method on \textit{nuScenes-C (Det3D)}. \textbf{Bold}: Best in column. \underline{Underline}: Second best in column. All scores are given in percentage ($\%$). \colorbox{Dark}{Dark}: Best in row. \colorbox{red!12}{Red}: Worst in row. Symbol $^{\dagger}$ denotes the baseline model adopted in calculating the CE scores.}
\vspace{0.cm}
\label{tab:nuscenes_c_det_ce_supp}
\centering\scalebox{0.837}{
\begin{tabular}{r|c|cccccccc|c}
    \toprule
    \textbf{Method} & \textbf{mCE~$\downarrow$} & \textbf{Fog} & \textbf{Wet} & \textbf{Snow} & \textbf{Motion} & \textbf{Beam} & \textbf{Cross} & \textbf{Echo} & \textbf{Sensor} & \textbf{NDS~$\uparrow$}
    \\\midrule\midrule
    CenterPoint-PP$^{\dagger}$~\cite{centerpoint} & $100.00$ & $100.00$ & $100.00$ & $100.00$ & $100.00$ & $100.00$ & $100.00$ & $100.00$ & $100.00$ & $45.99$
    \\\midrule
    SECOND-MH~\cite{second} & \underline{$97.50$} & \underline{$95.40$} & \underline{$96.01$} & \underline{$96.09$} & \underline{$100.81$} & \underline{$99.26$} & \cellcolor{Dark}\underline{$92.16$} & $97.64$ & \cellcolor{red!12}\underline{$102.64$} & $47.87$
    \\
    PointPillars-MH~\cite{lang2019pointpillars} & $102.90$ & $102.85$ & $104.56$ & $102.53$ & \cellcolor{red!12}$106.44$ & $102.39$ & \cellcolor{Dark}$100.94$ & $102.42$ & $\mathbf{101.05}$ & $43.33$
    \\
    CenterPoint-LR~\cite{centerpoint} & $98.74$ & $97.88$ & $96.46$ & $97.70$ & $102.15$ & $101.06$ & \cellcolor{Dark}$95.54$ & $\mathbf{95.60}$ & \cellcolor{red!12}$103.53$ & \underline{$49.72$}
    \\
    CenterPoint-HR~\cite{centerpoint} & $\mathbf{95.80}$ & $\mathbf{93.01}$ & $\mathbf{92.01}$ & $\mathbf{94.91}$ & $\mathbf{97.56}$ & $\mathbf{98.38}$ & \cellcolor{Dark}$\mathbf{91.11}$ & \underline{$96.21$} & \cellcolor{red!12}$103.23$ & $\mathbf{50.31}$
    \\\bottomrule
    \end{tabular}
}
\vspace{0.2cm}
\end{table*}

%% file: tables_supp/nuscenes_det_rr.tex
\begin{table*}[t]
\caption{[Complete Results] The \textbf{Resilience Rate (RR)} of each method on \textit{nuScenes-C (Det3D)}. \textbf{Bold}: Best in column. \underline{Underline}: Second best in column. All scores are given in percentage ($\%$). \colorbox{Dark}{Dark}: Best in row. \colorbox{red!9}{Red}: Worst in row.}
\vspace{0.cm}
\label{tab:nuscenes_c_det_rr}
\centering\scalebox{0.837}{
\begin{tabular}{r|c|cccccccc|c}
    \toprule
    \textbf{Method} & \textbf{mRR~$\uparrow$} & \textbf{Fog} & \textbf{Wet} & \textbf{Snow} & \textbf{Motion} & \textbf{Beam} & \textbf{Cross} & \textbf{Echo} & \textbf{Sensor} & \textbf{NDS~$\uparrow$}
    \\\midrule\midrule
    PointPillars-MH~\cite{lang2019pointpillars} & $\mathbf{77.24}$ & $76.53$ & \cellcolor{Dark}\underline{$99.05$} & $68.06$ & \underline{$87.79$} & $\mathbf{77.57}$ & $79.88$ & \underline{$71.31$} & \cellcolor{red!9}$\mathbf{57.70}$ & $43.33$
    \\
    SECOND-MH~\cite{second} & \underline{$76.96$} & $\mathbf{79.38}$ & \cellcolor{Dark}$\mathbf{99.42}$ & $\mathbf{70.86}$ & $86.32$ & $74.45$ & $\mathbf{84.19}$ & $71.28$ & \cellcolor{red!9}$49.76$ & $47.87$
    \\
    CenterPoint-PP~\cite{centerpoint} & $76.68$ & $76.13$ & \cellcolor{Dark}$98.74$ & $67.91$ & $\mathbf{90.87}$ & \underline{$76.45$} & $76.58$ & $70.73$ & \cellcolor{red!9}\underline{$56.06$} & $45.99$
    \\
    CenterPoint-LR~\cite{centerpoint} & $72.49$ & $73.19$ & \cellcolor{Dark}$95.21$ & $65.99$ & $81.54$ & $69.33$ & $76.65$ & $\mathbf{71.40}$ & \cellcolor{red!9}$46.58$ & \underline{$49.72$}
    \\
    CenterPoint-HR~\cite{centerpoint} & $75.26$ & \underline{$78.61$} & \cellcolor{Dark}$98.93$ & \underline{$69.03$} & $85.89$ & $71.97$ & \underline{$81.45$} & $69.75$ & \cellcolor{red!9}$46.47$ & $\mathbf{50.31}$
    \\\bottomrule
    \end{tabular}
}
\vspace{0.2cm}
\end{table*}

%% file: tables_supp/nuscenes_det_nds.tex
\begin{table*}[t]
\caption{[Complete Results] The \textbf{nuScenes Detection Score (NDS)} of each method on \textit{nuScenes-C (Det3D)}. \textbf{Bold}: Best in column. \underline{Underline}: Second best in column. All scores are given in percentage ($\%$). \colorbox{Dark}{Dark}: Best in row. \colorbox{red!9}{Red}: Worst in row.}
\vspace{0.cm}
\label{tab:nuscenes_c_det_nds}
\centering\scalebox{0.837}{
\begin{tabular}{r|c|c|c|cccccccc}
    \toprule
    \textbf{Method} & \textbf{mCE~$\downarrow$} & \textbf{mRR~$\uparrow$} & \textbf{NDS~$\uparrow$} & \textbf{Fog} & \textbf{Wet} & \textbf{Snow} & \textbf{Motion} & \textbf{Beam} & \textbf{Cross} & \textbf{Echo} & \textbf{Sensor}
    \\\midrule\midrule
    PointPillars-MH~\cite{lang2019pointpillars} & $102.90$ & $\mathbf{77.24}$ & $43.33$ & $33.16$ & \cellcolor{Dark}$42.92$ & $29.49$ & $38.04$ & $33.61$ & $34.61$ & $30.90$ & \cellcolor{red!9}\underline{$25.00$}
    \\
    SECOND-MH~\cite{second} & \underline{$97.50$} & \underline{$76.96$} & $47.87$ & \underline{$38.00$} & \cellcolor{Dark}\underline{$47.59$} & \underline{$33.92$} & $41.32$ & \underline{$35.64$} & \underline{$40.30$} & $34.12$ & \cellcolor{red!9}$23.82$
    \\
    CenterPoint-PP~\cite{centerpoint} & $100.00$ & $76.68$ & $45.99$ & $35.01$ & \cellcolor{Dark}$45.41$ & $31.23$ & \underline{$41.79$} & $35.16$ & $35.22$ & $32.53$ & \cellcolor{red!9}$\mathbf{25.78}$
    \\
    CenterPoint-LR~\cite{centerpoint} & $98.74$ & $72.49$ & \underline{$49.72$} & $36.39$ & \cellcolor{Dark}$47.34$ & $32.81$ & $40.54$ & $34.47$ & $38.11$ & $\mathbf{35.50}$ & \cellcolor{red!9}$23.16$
    \\
    CenterPoint-HR~\cite{centerpoint} & $\mathbf{95.80}$ & $75.26$ & $\mathbf{50.31}$ & $\mathbf{39.55}$ & \cellcolor{Dark}$\mathbf{49.77}$ & $\mathbf{34.73}$ & $\mathbf{43.21}$ & $\mathbf{36.21}$ & $\mathbf{40.98}$ & \underline{$35.09$} & \cellcolor{red!9}$23.38$
    \\\bottomrule
    \end{tabular}
}
\vspace{0.2cm}
\end{table*}

%% file: tables_supp/wod_seg_ce.tex
\begin{table*}[t]
\caption{[Complete Results] The \textbf{Corruption Error (CE)} of each method on \textit{WOD-C (Seg3D)}. \textbf{Bold}: Best in column. \underline{Underline}: Second best in column. All scores are given in percentage ($\%$).  {\small\colorbox{Dark}{Dark}}: Best in row. {\small\colorbox{red!8}{Red}}: Worst in row.}
\vspace{0.cm}
\label{tab:wod_c_seg3d_ce_supp}
\centering\scalebox{0.837}{
\begin{tabular}{r|c|cccccccc|c}
    \toprule
    \textbf{Method} & \textbf{mCE~$\downarrow$} & \textbf{Fog} & \textbf{Wet} & \textbf{Snow} & \textbf{Motion} & \textbf{Beam} & \textbf{Cross} & \textbf{Echo} & \textbf{Sensor} & \textbf{mIoU~$\uparrow$}
    \\\midrule\midrule
    MinkUNet$_{18}$$^{\dagger}$~\cite{2019Minkowski} & $100.00$ & $100.00$ & $100.00$ & $100.00$ & $100.00$ & $100.00$ & $100.00$ & $100.00$ & $100.00$ & \underline{$69.06$}
    \\\midrule
    MinkUNet$_{34}$~\cite{2019Minkowski} & $\mathbf{96.21}$ & $\mathbf{96.00}$ & \underline{$94.90$} & \cellcolor{red!9}$99.53$ & $\mathbf{96.20}$ & $\mathbf{95.43}$ & $\mathbf{96.79}$ & $\mathbf{96.75}$ & \cellcolor{Dark}$\mathbf{94.08}$ & $\mathbf{70.15}$
    \\
    Cylinder3D$_{\text{TSC}}$~\cite{zhu2021cylindrical} & $106.02$ & \cellcolor{red!9}$111.81$ & $104.08$ & \cellcolor{Dark}$\mathbf{98.39}$ & $110.30$ & $105.77$ & $106.87$ & $108.24$ & $102.69$ & $65.93$
    \\\midrule
    SPVCNN$_{18}$~\cite{tang2020searching} & $103.60$ & \cellcolor{red!9}$105.63$ & $104.79$ & \cellcolor{Dark}\underline{$99.17$} & $105.41$ & $104.85$ & \underline{$99.74$} & $104.28$ & $104.91$ & $67.35$
    \\
    SPVCNN$_{34}$~\cite{tang2020searching} & \underline{$98.72$} & \underline{$99.67$} & $\mathbf{96.36}$ & $100.43$ & \underline{$100.00$} & \underline{$98.55$} & \cellcolor{red!9}$101.93$ & \underline{$97.87$} & \cellcolor{Dark}\underline{$94.97$} & $69.01$
    \\\bottomrule
    \end{tabular}
}
\vspace{0.2cm}
\end{table*}

%% file: tables_supp/wod_seg_rr.tex
\begin{table*}[t]
\caption{[Complete Results] The \textbf{Resilience Rate (RR)} of each method on \textit{WOC-C (Seg3D)}. \textbf{Bold}: Best in column. \underline{Underline}: Second best in column. All scores are given in percentage ($\%$). {\small\colorbox{Dark}{Dark}}: Best in row. {\small\colorbox{red!8}{Red}}: Worst in row.}
\vspace{0.cm}
\label{tab:wod_c_seg3d_rr}
\centering\scalebox{0.837}{
\begin{tabular}{r|c|cccccccc|c}
    \toprule
    \textbf{Method} & \textbf{mRR~$\uparrow$} & \textbf{Fog} & \textbf{Wet} & \textbf{Snow} & \textbf{Motion} & \textbf{Beam} & \textbf{Cross} & \textbf{Echo} & \textbf{Sensor} & \textbf{mIoU~$\uparrow$}
    \\\midrule\midrule
    MinkUNet$_{18}$~\cite{2019Minkowski} & $91.22$ & $97.00$ & $88.31$ & $83.62$ & \cellcolor{Dark}$99.80$ & $92.89$ & $94.66$ & $91.75$ & \cellcolor{red!9}$81.73$ & \underline{$69.06$}
    \\
    MinkUNet$_{34}$~\cite{2019Minkowski} & $91.80$ & $\mathbf{97.38}$ & $89.78$ & \cellcolor{red!9}$82.61$ & \cellcolor{Dark}$\mathbf{99.93}$ & \underline{$93.78$} & $94.77$ & \underline{$92.02$} & \underline{$84.13$} & $\mathbf{70.15}$
    \\
    Cylinder3D$_{\text{TSC}}$~\cite{zhu2021cylindrical} & $\mathbf{92.39}$ & $95.69$ & \underline{$90.10$} & $\mathbf{88.62}$ & \cellcolor{Dark}$99.68$ & $\mathbf{94.16}$ & \underline{$95.54$} & $91.52$ & \cellcolor{red!9}$83.83$ & $65.93$
    \\\midrule
    SPVCNN$_{18}$~\cite{tang2020searching} & $91.60$ & $96.70$ & $87.78$ & \underline{$86.27$} & \cellcolor{Dark}$99.84$ & $92.67$ & $\mathbf{97.19}$ & $91.74$ & \cellcolor{red!9}$80.62$ & $67.35$
    \\
    SPVCNN$_{34}$~\cite{tang2020searching} & \underline{$92.04$} & \underline{$97.23$} & $\mathbf{90.44}$ & \cellcolor{red!9}$83.42$ & \cellcolor{Dark}\underline{$99.87$} & $93.71$ & $93.75$ & $\mathbf{92.94}$ & $\mathbf{84.96}$ & $69.01$
    \\\bottomrule
    \end{tabular}
}
\vspace{0.2cm}
\end{table*}

%% file: tables_supp/wod_seg_iou.tex
\begin{table*}[t]
\caption{[Complete Results] The \textbf{Intersection-over-Union (IoU)} of each method on \textit{WOD-C (Seg3D)}. \textbf{Bold}: Best in column. \underline{Underline}: Second best in column. All scores are given in percentage ($\%$).  {\small\colorbox{Dark}{Dark}}: Best in row. {\small\colorbox{red!8}{Red}}: Worst in row.}
\vspace{0.cm}
\label{tab:wod_c_seg3d_iou}
\centering\scalebox{0.837}{
\begin{tabular}{r|c|c|c|cccccccc}
    \toprule
    \textbf{Method} & \textbf{mCE~$\downarrow$} & \textbf{mRR~$\uparrow$} & \textbf{mIoU~$\uparrow$} & \textbf{Fog} & \textbf{Wet} & \textbf{Snow} & \textbf{Motion} & \textbf{Beam} & \textbf{Cross} & \textbf{Echo} & \textbf{Sensor}
    \\\midrule\midrule
    MinkUNet$_{18}$~\cite{2019Minkowski} & $100.00$ & $91.22$ & \underline{$69.06$} & $66.99$ & $60.99$ & \underline{$57.75$} & \cellcolor{Dark}\underline{$68.92$} & $64.15$ & $65.37$ & $63.36$ & \cellcolor{red!9}$56.44$
    \\
    MinkUNet$_{34}$~\cite{2019Minkowski} & $\mathbf{96.21}$ & $91.80$ & $\mathbf{70.15}$ & $\mathbf{68.31}$ & $62.98$ & \cellcolor{red!9}$57.95$ & \cellcolor{Dark}$\mathbf{70.10}$ & $\mathbf{65.79}$ & $\mathbf{66.48}$ & $\mathbf{64.55}$ & $\mathbf{59.02}$
    \\
    Cylinder3D$_{\text{TSC}}$~\cite{zhu2021cylindrical} & $106.02$ & $\mathbf{92.39}$ & $65.93$ & $63.09$ & $59.40$ & $58.43$ & \cellcolor{Dark}$65.72$ & $62.08$ & $62.99$ & $60.34$ & \cellcolor{red!9}$55.27$
    \\\midrule
    SPVCNN$_{18}$~\cite{tang2020searching} & $103.60$ & $91.60$ & $67.35$ & $65.13$ & $59.12$ & $58.10$ & \cellcolor{Dark}$67.24$ & $62.41$ & \underline{$65.46$} & $61.79$ & \cellcolor{red!9}$54.30$
    \\
    SPVCNN$_{34}$~\cite{tang2020searching} & \underline{$98.72$} & \underline{$92.04$} & $69.01$ & \underline{$67.10$} & $\mathbf{62.41}$ & \cellcolor{red!9}$\mathbf{57.57}$ & \cellcolor{Dark}\underline{$68.92$} & \underline{$64.67$} & $64.70$ & \underline{$64.14$} & \underline{$58.63$}
    \\\bottomrule
    \end{tabular}
}
\vspace{0.2cm}
\end{table*}

%% file: tables_supp/wod_det_ce.tex
\begin{table*}[t]
\caption{[Complete Results] The \textbf{Corruption Error (CE)} of each method on \textit{WOD-C (Det3D)}. \textbf{Bold}: Best in column. \underline{Underline}: Second best in column. All scores are given in percentage ($\%$). {\small\colorbox{Dark}{Dark}}: Best in row. {\small\colorbox{red!8}{Red}}: Worst in row.}
\vspace{0.cm}
\label{tab:wod_c_det3d_ce_supp}
\centering\scalebox{0.837}{
\begin{tabular}{r|c|cccccccc|c}
    \toprule
    \textbf{Method} & \textbf{mCE~$\downarrow$} & \textbf{Fog} & \textbf{Wet} & \textbf{Snow} & \textbf{Motion} & \textbf{Beam} & \textbf{Cross} & \textbf{Echo} & \textbf{Sensor} & \textbf{mAPH~$\uparrow$}
    \\\midrule\midrule
    CenterPoint$^{\dagger}$~\cite{centerpoint} & $100.00$ & $100.00$ & $100.00$ & $100.00$ & $100.00$ & $100.00$ & $100.00$ & $100.00$ & $100.00$ & \underline{$63.59$}
    \\\midrule
    SECOND~\cite{second} & $121.43$ & $117.86$ & $126.51$ & $127.51$ & \cellcolor{Dark}$113.37$ & $121.25$ & \cellcolor{red!9}$127.82$ & $123.66$ & $113.48$ & $53.37$
    \\
    PointPillars~\cite{lang2019pointpillars} & $127.53$ & $120.76$ & $135.23$ & $129.65$ & $115.23$ & $122.99$ & \cellcolor{red!9}$151.71$ & $131.64$ & \cellcolor{Dark}$113.05$ & $50.17$
    \\
    PV-RCNN~\cite{pvrcnn} & \underline{$104.90$} & \underline{$110.08$} & \underline{$104.22$} & \cellcolor{Dark}\underline{$95.68$} & \underline{$101.33$} & \cellcolor{red!9}\underline{$110.70$} & \underline{$101.84$} & \underline{$106.00$} & \underline{$109.37$} & $61.27$
    \\
    PV-RCNN++~\cite{pvrcnn++} & $\mathbf{91.60}$ & $\mathbf{95.71}$ & \cellcolor{Dark}$\mathbf{88.32}$ & \cellcolor{red!9}$\mathbf{90.05}$ & $\mathbf{93.24}$ & $\mathbf{92.50}$ & $\mathbf{88.94}$ & $\mathbf{90.81}$ & $\mathbf{93.23}$ & $\mathbf{67.45}$
    \\\bottomrule
    \end{tabular}
}
\vspace{0.2cm}
\end{table*}

%% file: tables_supp/wod_det_rr.tex
\begin{table*}[t]
\caption{[Complete Results] The \textbf{Resilience Rate (RR)} of each method on \textit{WOD-C (Det3D)}. \textbf{Bold}: Best in column. \underline{Underline}: Second best in column. All scores are given in percentage ($\%$). {\small\colorbox{Dark}{Dark}}: Best in row. {\small\colorbox{red!8}{Red}}: Worst in row.}
\vspace{0.cm}
\label{tab:wod_c_det3d_rr}
\centering\scalebox{0.837}{
\begin{tabular}{r|c|cccccccc|c}
    \toprule
    \textbf{Method} & \textbf{mRR~$\uparrow$} & \textbf{Fog} & \textbf{Wet} & \textbf{Snow} & \textbf{Motion} & \textbf{Beam} & \textbf{Cross} & \textbf{Echo} & \textbf{Sensor} & \textbf{mAPH~$\uparrow$}
    \\\midrule\midrule
    PointPillars~\cite{lang2019pointpillars} & $81.23$ & \cellcolor{red!9}$62.27$ & \cellcolor{Dark}$99.16$ & $92.31$ & $69.62$ & $\mathbf{87.56}$ & $79.33$ & $86.53$ & $\mathbf{73.09}$ & $50.17$
    \\
    SECOND~\cite{second} & $81.12$ & \cellcolor{red!9}$61.63$ & \cellcolor{Dark}$99.29$ & $88.44$ & $67.42$ & $83.79$ & $92.34$ & $87.76$ & $68.26$ & $53.37$
    \\
    PV-RCNN~\cite{pvrcnn} & $82.43$ & \cellcolor{red!9}$60.91$ & \cellcolor{Dark}$\mathbf{100.00}$ & $\mathbf{98.55}$ & \underline{$69.82$} & $80.84$ & $\mathbf{97.26}$ & $88.84$ & $63.21$ & $61.27$
    \\
    CenterPoint~\cite{centerpoint} & \underline{$83.30$} & \cellcolor{red!9}$\mathbf{67.72}$ & \cellcolor{Dark}$98.82$ & $92.14$ & $68.45$ & $85.56$ & $94.86$ & \underline{$89.65$} & $69.16$ & \underline{$63.59$}
    \\
    PV-RCNN++~\cite{pvrcnn++} & $\mathbf{84.14}$ & \cellcolor{red!9}\underline{$67.46$} & \cellcolor{Dark}\underline{$99.60$} & \underline{$92.97$} & $\mathbf{70.20}$ & \underline{$85.74$} & \underline{$95.94$} & $\mathbf{90.38}$ & \underline{$70.82$} & $\mathbf{67.45}$
    \\\bottomrule
    \end{tabular}
}
\vspace{0.2cm}
\end{table*}

%% file: tables_supp/wod_det_aph.tex
\begin{table*}[t]
\caption{[Complete Results] The \textbf{Average Precision (APH)} of each method on \textit{WOD-C (Det3D)}. \textbf{Bold}: Best in column. \underline{Underline}: Second best in column. All scores are given in percentage ($\%$). {\small\colorbox{Dark}{Dark}}: Best in row. {\small\colorbox{red!8}{Red}}: Worst in row.}
\vspace{0.cm}
\label{tab:wod_c_det3d_aph}
\centering\scalebox{0.837}{
\begin{tabular}{r|c|c|c|cccccccc}
    \toprule
    \textbf{Method} & \textbf{mCE~$\downarrow$} & \textbf{mRR~$\uparrow$} & \textbf{mAPH~$\uparrow$} & \textbf{Fog} & \textbf{Wet} & \textbf{Snow} & \textbf{Motion} & \textbf{Beam} & \textbf{Cross} & \textbf{Echo} & \textbf{Sensor}
    \\\midrule\midrule
    PointPillars~\cite{lang2019pointpillars} & $127.53$ & $81.23$ & $50.17$ & \cellcolor{red!9}$31.24$ & \cellcolor{Dark}$49.75$ & $46.07$ & $34.93$ & $43.93$ & $39.80$ & $43.41$ & $36.67$
    \\
    SECOND~\cite{second} & $121.43$ & $81.12$ & $53.37$ & \cellcolor{red!9}$32.89$ & \cellcolor{Dark}$52.99$ & $47.20$ & $35.98$ & $44.72$ & $49.28$ & $46.84$ & $36.43$
    \\
    PV-RCNN~\cite{pvrcnn} & $104.90$ & $82.43$ & $61.27$ & \cellcolor{red!9}$37.32$ & \cellcolor{Dark}$61.27$ & \underline{$60.38$} & $42.78$ & $49.53$ & $59.59$ & $54.43$ & $38.73$
    \\
    CenterPoint~\cite{centerpoint} & \underline{$100.00$} & \underline{$83.30$} & \underline{$63.59$} & \cellcolor{red!9}\underline{$43.06$} & \cellcolor{Dark}\underline{$62.84$} & $58.59$ & \underline{$43.53$} & \underline{$54.41$} & \underline{$60.32$} & \underline{$57.01$} & \underline{$43.98$}
    \\
    PV-RCNN++~\cite{pvrcnn++} & $\mathbf{91.60}$ & $\mathbf{84.14}$ & $\mathbf{67.45}$ & \cellcolor{red!9}$\mathbf{45.50}$ & \cellcolor{Dark}$\mathbf{67.18}$ & $\mathbf{62.71}$ & $\mathbf{47.35}$ & $\mathbf{57.83}$ & $\mathbf{64.71}$ & $\mathbf{60.96}$ & $\mathbf{47.77}$
    \\\bottomrule
    \end{tabular}
}
\vspace{0.2cm}
\end{table*}